\documentclass[10pt]{article} 

\usepackage[preprint]{rlj} 

\usepackage{amssymb}            
\usepackage{mathtools}          
\usepackage{mathrsfs}           
\usepackage{graphicx}           
\usepackage{subcaption}         
\usepackage[space]{grffile}     
\usepackage{url}                
\usepackage{lipsum}             
\usepackage{float}
\usepackage{booktabs}
\usepackage{multirow}

\usepackage{array}
\usepackage{makecell}
\usepackage[table]{xcolor}
\usepackage{wrapfig}
\usepackage{subcaption} 
\usepackage{rotating}

\renewcommand{\arraystretch}{1.2}

\newcolumntype{C}[1]{>{\centering\arraybackslash}m{#1}}

\title{Towards a Practical Understanding of Lagrangian Methods in Safe Reinforcement Learning}
\setrunningtitle{Towards a Practical Understanding of Lagrangian Methods in Safe Reinforcement Learning}

\author{Lindsay Spoor\textsuperscript{1,$\dagger$}, Álvaro Serra-Gómez\textsuperscript{1}, Aske Plaat\textsuperscript{1}, Thomas Moerland\textsuperscript{1}}

\emails{l.j.spoor@liacs.leidenuniv.nl}

\affiliations{
$^{1}$\textbf{Leiden Institute of Advanced Computer Science, Leiden University, Leiden, The Netherlands}\\

}


\begin{document}

\maketitle  

\begin{abstract}
    Safe reinforcement learning addresses constrained optimization problems where maximizing performance must be balanced against safety constraints, and Lagrangian methods are a widely used approach for this purpose. However, the effectiveness of Lagrangian methods depends crucially on the choice of the Lagrange multiplier $\lambda$, which governs the multi-objective trade-off between return and cost. A common practice is to update the multiplier automatically during training. Although this approach is standard in practice, there remains limited empirical evidence on the optimally achievable trade-off between return and cost as a function of $\lambda$, and there is currently no systematic benchmark comparing automated update mechanisms to this empirical optimum.
    Therefore, we study (i) the \textit{constraint geometry} for eight widely used safety tasks and (ii) the previously overlooked \textit{constraint-regime sensitivity} of different Lagrange multiplier update mechanisms in safe reinforcement learning. Through the lens of multi-objective analysis, we present empirical Pareto frontiers that offer a complete visualization of the trade-off between return and cost in the underlying optimization problem. Our results reveal the highly sensitive nature of $\lambda$ and further show that the restrictiveness of the constraint cost can vary across different cost limits within the same task. This highlights the importance of careful cost limit selection across different regions of cost restrictiveness when evaluating safe reinforcement learning methods. We provide a recommended set of cost limits for each evaluated task and offer an open-source code base: \url{https://github.com/lindsayspoor/Lagrangian_SafeRL}.
\end{abstract}

\section{Introduction}
\label{sec:Introduction}

Reinforcement learning (RL) addresses sequential decision-making problems by enabling agents to learn from feedback in the form of rewards, with the goal of maximizing their long-term cumulative reward \citep{sutton_reinforcement_nodate}. Despite their success in achieving high performance on tasks without critical safety concerns, agents deployed in safety-critical domains must often deal with restrictive constraints. For example, a robot learning locomotion must satisfy safety constraints such as torque limits or avoiding collisions ~\citep{he_autocost_2023, huang_safedreamer_2024} and falling when walking in the real world ~\citep{ha_learning_2020}, often requiring a detour from the unconstrained optimal policy. Safety also plays a crucial role in operational domains such as power systems, where agents tasked with scheduling or real-time control must ensure stability and reliability while simultaneously optimizing performance \citep{yan_real-time_2020, chen_deep_2025, su_review_2024}.

Safe reinforcement learning, also referred to as constrained reinforcement learning, provides a framework in which the learning objectives are extended to explicitly incorporate constraints imposed on the agent. In this framework, the optimization problem has multiple conflicting objectives: the agent must learn a policy that maximizes expected return while simultaneously keeping constraint costs below a specified limit. Safety in RL has been extensively studied over the last decade, leading to a variety of approaches that tackle the constrained optimization problem. 

Classical Lagrangian methods have emerged as a popular choice in practical applications in which constraints are softly enforced during training \citep{garcia_comprehensive_2015}, showing performance close to the optima while respecting constraints in safety-critical tasks~\citep{ray_benchmarking_nodate}. 
By augmenting the objective with a penalty term weighted by a Lagrange multiplier $\lambda$, these methods reformulate the constrained problem into an unconstrained one, allowing us to apply any standard RL algorithm while implicitly accounting for safety.

The Lagrange multiplier controls the trade-off between reward and constraint cost: setting it too low encourages safety-compromising policies that violate constraints, while setting it too high can lead to overly conservative policies with poor returns. Theoretically, if the optimal multiplier $\lambda^*$ is known, the solution to the relaxed unconstrained problem is equivalent to the constrained multi-objective problem \citep{borkar_actor-critic_2005}. In practice, however, identifying $\lambda^*$ \textit{"can be as hard as solving the RL problem itself"} \citep{paternain_constrained_2019} due to the lack of a direct relation between the multiplier value and the resulting policy performance. Tuning $\lambda$ is computationally expensive and sensitive to the task, since the optimal trade-off often lies on a Pareto frontier between return and cost \citep{roijers_survey_2013, moaert_multi-objective_nodate}.

Although it is common practice to automatically update $\lambda$ during training \citep{tessler_reward_2018, stooke_responsive_2020}, these methods introduce additional complexity without any guarantees with respect to the constraint cost limit. There remains limited empirical evidence on the achievable optimal trade-off between return and cost as a function of $\lambda$, and there is currently no systematic benchmark study comparing automated update mechanisms to this empirical optimum. Moreover, in practice, one would oftentimes set the cost limit to an arbitrary value without considering the underlying constraint restrictiveness of the problem. As a result, there is limited understanding of how the choice of cost limit influences the practically achievable performance, both within individual tasks and across different tasks.

This paper aims to fill the gap on the practical performance of Lagrangian methods in safe RL, by presenting a systematic empirical analysis focusing on the role of the Lagrange multiplier, along with recommendations to the safe RL benchmarking community and a code base that allows for an empirical analysis of safe RL benchmark tasks. Our contributions are the following:
\begin{enumerate}
    \item \textbf{Constraint geometry} \hspace{1em} We systematically analyze how the choice of the Lagrange multiplier influences the resulting model performance. We visualize the trade-off between the return and constraint cost as a function of $\lambda$, and find that the optimal value for $\lambda$ is highly task-dependent. Additionally, we provide empirical Pareto frontiers that show the underlying constraint cost restrictiveness for eight widely-used Safety-Gymnasium tasks \citep{ji_safety-gymnasium_2024}, and, we find that this measure of restrictiveness can differ for varying cost limits.
    \item \textbf{Constraint-regime sensitivity} \hspace{1em } We compare two Lagrange multiplier update mechanisms in terms of their performance across a range of cost limits. Our results show that, due to variations in the slope of the empirical Pareto frontiers both across tasks and across cost limits within the same task, there is no consistently best-performing update mechanism for Lagrangian-based safe RL. Instead, the relative performance of each method depends strongly on the chosen cost limit, suggesting the importance of careful cost-limit selection and systematic evaluation across multiple constraint settings in safe RL benchmarks.
\end{enumerate}

\section{Related work}
\label{sec:Related Work}

\textbf{Lagrangian methods in safe RL} \hspace{1em} \cite{borkar_actor-critic_2005} introduced the dual gradient descent framework for actor–critic methods and showed that updating the Lagrange multiplier via gradient ascent guarantees convergence to the optimal value $\lambda^*$. In this framework, they provided that the policy and value function updates must occur on faster, converged timescales, compared to a slower timescale on which the Lagrange multiplier is updated. Subsequent theoretical work by \cite{paternain_constrained_2019} showed that constrained RL problems exhibit zero duality gap, providing the theoretical guarantee that the constrained MDP can, in principle, be solved exactly in the dual domain. They further introduced primal–dual approaches for probabilistic constraints \citep{paternain_safe_2022}, demonstrating that safe policies can be obtained under realistic uncertainty models.

\cite{tessler_reward_2018} introduced RCPO, a Lagrangian-based algorithm that updates the multiplier through gradient ascent. Several works have since explored extensions of Lagrangian methods. \cite{ding_provably_2023} extended the Lagrangian framework to multi-agent RL, and they furthermore proposed a regularized Lagrangian framework to guarantee safety beyond the asymptotic convergence \citep{ding_last-iterate_2024}. \cite{jayant_model-based_2022} introduced a model-based Lagrangian method and showed that integrating model dynamics can accelerate convergence while maintaining safety guarantees. \cite{stooke_responsive_2020} revisited the Lagrange multiplier update mechanism and introduced an automated update method that relies on proportional-integral-derivative control, and show that this method stabilizes training compared to pure gradient ascent. 

\textbf{Multi-objective RL} \hspace{1em} Constrained RL is closely related to multi-objective RL, as both involve balancing multiple objectives. Noted by \cite{ray_benchmarking_nodate}, safety requirements typically exhibit a saturation effect: once the safety threshold is satisfied, further improvements no longer make the system any safer. This property corresponds to the constraint threshold in the constrained formulation. Although this has no direct equivalent in multi-objective optimization, a multi-objective analysis on the return and cost of constrained optimization problems in RL could add value to our understanding of the underlying constraint geometry of the problem, as the optimal trade-off theoretically lies on a return-cost Pareto frontier \citep{roijers_survey_2013, moaert_multi-objective_nodate}. 

\textbf{Empirical studies of safe RL} \hspace{1em} From an empirical standpoint, \cite{ray_benchmarking_nodate} introduced the Safety Gym benchmark suite, providing standardized environments to assess safe RL algorithms. Their study highlighted that simple Lagrangian-based methods perform competitively among safe RL algorithms, but did not explicitly analyze the role of the Lagrange multiplier itself. Focusing on the \textit{update} of the Lagrange multiplier, \cite{stooke_responsive_2020} provided the first systematic empirical insights into PID-controlled multiplier updates, examining the individual influence of its three tunable hyperparameters.

Despite the well-known sensitivity and practical difficulty of tuning $\lambda$, empirical investigations that provide practical intuition about its behavior remain limited, even though prior work has advanced the theoretical and algorithmic foundations of Lagrangian methods. To the best of our knowledge, no prior work has systematically characterized the empirical behavior of $\lambda$ in Lagrangian methods, nor provided a systematic multi-objective analysis of the return–cost trade-off. 

\section{Constrained Markov Decision Process}
\label{sec:Constrained Markov Decision Process}

The reinforcement learning problem is commonly formalized as a Markov Decision Process (MDP) \citep{sutton_reinforcement_nodate}. When the problem additionally contains a set of constraints, we speak of constrained reinforcement learning, for which we use the formal framework of a Constrained Markov Decision Process (CMDP) \citep{altman_constrained_1999}. A CMDP is described by the tuple $\mathcal{M}=(\mathcal{S,A}, p, r , p_0, \gamma, \mathcal{C})$, where $\mathcal{S}$ and $\mathcal{A}$ are the set of all possible states and actions, respectively, $p$ is the transition dynamics distribution $p:\mathcal{S}\times\mathcal{A}\rightarrow \Delta(\mathcal{S})$, $r$ is the reward function $r: \mathcal{S \times A \times S}\rightarrow \mathbb{R}$, $p_0\in\Delta(\mathcal{S})$ is the initial state distribution and $\gamma \in [0,1)$ is the discount factor that governs the importance of future rewards. A set of cost functions $\mathcal{C}\coloneq \{C_1, ..., C_m\}$ with cost limits $d_1,...,d_m$ maps transition tuples to costs, $\mathcal{C:S\times A\times S}\rightarrow \mathbb{R}^m$. Actions are selected from a policy $\pi_{\theta}$, where $\pi$ defines a mapping $\pi: \mathcal{S}\rightarrow\Delta(\mathcal{A}), s\mapsto \pi(\cdot|s)$, and $\theta$ is the set of parameters, for a stationary parametrized policy $\pi$ in the set of all policies $\Pi$. We denote  $R(\tau) = \sum_{k=0}^{\infty}\gamma^k r(s_{t+k}, a_{t+k}, s_{t+k+1})$ as the return of a trajectory $\tau=(s_t, a_t, s_{t+1}, ...) \sim p_{\pi_{\theta}}(s'|s,a)$. 

The state value function for the return is defined as $V^R_{\pi_{\theta}}(s_0) \overset{\cdot}{=} \mathbb{E}_{\tau\sim\pi_{\theta}}\big[ R(\tau)|s_t=s\big]$ and yields the return objective of the CMDP, which is to maximize the expected cumulative discounted reward $J^R(\pi_{\theta})$, which is equivalent to $V^R_{\pi_{\theta}}(s_0)$. The expected cumulative discounted cost of the policy $\pi_{\theta}$ is defined as $J^{C_i}(\pi_{\theta})$, which is equivalent to the state value function for the cost, $V^{C_i}_{\pi_{\theta}}(s_0) \overset{\cdot}{=} \mathbb{E}_{\tau\sim\pi_{\theta}}\big[ C_i(\tau)|s_t=s\big]$, with $C_i(\tau) = \sum_{k=0}^{\infty}\gamma^k C_i(s_{t+k}, a_{t+k}, s_{t+k+1})$. The feasible set of stationary parametrized policies is $\Pi_{\mathcal{C}} \overset{\cdot}{=} \{\pi_{\theta}\in\Pi:\forall i, J^{C_i}(\pi_{\theta})\leq d_i\}$. The optimization problem of a CMDP can be expressed as:
\begin{eqnarray} \label{eq: optimization problem CMDP}
\begin{aligned}
    &\max_{\pi_{\theta} \in \Pi_{\theta}} J^R(\pi_{\theta}) \\
    &\text{s.t.} \quad J^{C_i}(\pi_{\theta}) \leq d_i, i=1,...,m,
\end{aligned}
\end{eqnarray}
where $\Pi_{\theta}\subseteq \Pi$ denotes the set of parametrized policies with parameters $\theta$. Compared to traditional MDPs \citep{bellman_markovian_1957}, local policy search for CMDPs involves the additional requirement that each policy iteration remains feasible with respect to the CMDP constraints. Therefore, instead of optimizing over $\Pi_{\theta}$, the optimization algorithm should optimize over $\Pi_{\theta} \cap \Pi_{\mathcal{C}}$. The optimal policy $\pi^{*}$ of a CMDP is then found by $\pi^* = \arg\max_{\pi_{\theta} \in \Pi_{\mathcal{C}}} J^R(\pi_{\theta})$.

\section{Methodology}
\label{sec:Methodology}

\subsection{Lagrangian methods}
\label{sec:Lagrangian Methods}

The optimization problem in Eq. \ref{eq: optimization problem CMDP} is in a primal form, implying that the constraints must be strictly satisfied at every step and thus each policy update has to remain feasible. We can soften the constraints during training by moving to the dual problem following the Lagrangian method. This method converts a CMDP into an unconstrained one with Lagrange relaxation \citep{bertsekas_nonlinear_1997, boyd_convex_2023}:
\begin{eqnarray} \label{eq: unconstrained CMDP problem}
    \min\limits_{\lambda\geq0}\max\limits_{\theta}\mathcal{L}(\lambda,\theta) = \min\limits_{\lambda\geq0}\max\limits_{\theta}\Big[J^R(\pi_{\theta})- \Big(\sum_{i=1}^{m}\lambda_i(J^{C_i}(\pi_{\theta})-d_i)\Big)\Big],
\end{eqnarray}
where $\mathcal{L}$ is the Lagrangian and $\lambda_i$ is the Lagrange multiplier for the $i$-th constraint. For convenience in notation we drop the index $i$ and use $\lambda$, $J^{\mathcal{C}}$ and $d$ to encompass the entire set of constraints collectively in the rest of this paper \footnote{Dropping index $i$ is done only to avoid clutter in notation. We still implicitly refer to the entire set of constraints, in which each individual constraint corresponds with its own cost limit. For clarification: this means that it is still possible to assume multiple constraints in the CMDP.}. The inequality constraints of the optimization problem are now relaxed to a penalty loss term $\xi = J^{\mathcal{C}}(\pi_{\theta})-d$. This allows us to find the optimal solution of a CMDP using any standard RL algorithm with a modified optimization objective. Intuitively, $\lambda$ is a penalty parameter in the optimization objective, which can be viewed as a parameter that defines the trade-off between the return and cost. 

\textbf{Fixed Lagrange multiplier} \hspace{1em}
The Lagrange multiplier $\lambda$ can be manually set to a constant value and kept fixed throughout training. The resulting unconstrained problem in Eq. \ref{eq: unconstrained CMDP problem} can then be solved by maximizing over the policy parameters $\theta$. When $\lambda$ is chosen to be the optimal value $\lambda^*$, this formulation is equivalent to solving the original constrained problem from Eq. \ref{eq: optimization problem CMDP} and one would be able to find the optimal solution at the saddle point $\big(\theta(\lambda^*),\lambda^*\big)$ \citep{borkar_actor-critic_2005}.

\subsection{Automated multiplier updates}
\label{sec:Automated Multiplier Updates}

Finding the optimal value $\lambda^*$ is often computationally- and time-intensive in practice, which motivates the search for automated alternatives. Eq. \ref{eq: unconstrained CMDP problem} is in dual form and convex, which allows us to efficiently solve it using gradient descent. Then, the \textit{dual gradient descent algorithm} alternates between the optimization of the policy parameters $\theta$ and the Lagrange multiplier $\lambda$ \citep{boyd_convex_2023}. On a fast time-scale the unconstrained problem in Eq. \ref{eq: unconstrained CMDP problem} is solved to update $\theta$. Then, on a slower time-scale, $\lambda$ is updated by minimizing the penalty loss following a preferred update rule. If the agent violates fewer constraints, $\lambda$ will gradually be decreased, and vice versa, until all cost functions satisfy their respective cost limits.

\textbf{Gradient ascent (GA) update} \hspace{1em}
Performing gradient descent by taking $\nabla_{\lambda}\mathcal{L} = -\xi$ and substituting this in $\lambda_{k+1}=\lambda_k-\eta\cdot\nabla_{\lambda}\mathcal{L}$, with $\eta$ a step-size parameter, yields a gradient \textit{ascent} (GA)-update on the Lagrange multiplier as in Eq. \ref{eq: lag_multiplier_update}.
\begin{eqnarray} \label{eq: lag_multiplier_update}
\lambda_{k+1}=\big(\lambda_k+\eta\cdot\xi\big)_+,
\end{eqnarray}
where $(\cdot)_{+}$ denotes the projection onto $\mathbb{R}_{+}$ and comes from the KKT conditions for inequality-constrained optimization \citep{boyd_convex_2023, borkar_actor-critic_2005, tessler_reward_2018}.

\textbf{PID-controlled update} \hspace{1em}
The gradient ascent update from Eq. \ref{eq: lag_multiplier_update} integrates the constraint, but its inherent learning dynamics can lead to oscillations when modeled with second-order dynamics \citep{platt_constrained_1988}. This is because the outputs are adjusted proportional to the \textit{accumulated} constraint violations over time. This results in frequent constraint violations during intermediate iterates. \cite{stooke_responsive_2020} proposed an update method that utilizes the derivatives of the penalty term, introducing an additional proportional and derivative control term to the Lagrange multiplier update as shown in Eq. \ref{eq: PID controlled update}.
\begin{eqnarray} \label{eq: PID controlled update}
    \lambda_{k+1}=\big(K_P\xi_k+K_I I_k+K_D\partial_k\big)_+,
\end{eqnarray}
where $\xi_k = J^{\mathcal{C}}(\pi_{\theta_k})-d_k$ is the penalty loss as a function of update iteration $k$, $I_k=(I_{k-1}+\xi_k)_+$ is the integral of the penalty loss, $\partial_k=\big(J^{\mathcal{C}}(\pi_{\theta_k})-J^{\mathcal{C}}(\pi_{\theta_{k-1}})\big)_+$ is the derivative of the constraint, and $K_P, K_I$ and $K_D$ are tunable step-size parameters corresponding to the proportional, integral and derivative coefficients, respectively \citep{aastrom2006pid}. 


\subsection{Pareto frontiers and constraint restrictiveness} \label{sec: pareto frontiers and optimal lambda}

Geometrically, the optimal solution to the optimization problem in Eq.~\ref{eq: optimization problem CMDP} lies on the upper boundary of the achievable return for a given cost limit $d$, which forms the Pareto frontier (PF) of the return as a function of the cost: $R(C)$. Each point on this frontier corresponds to a policy for which no other policy achieves both higher return and lower cost. The constrained optimization problem therefore amounts to selecting the point on the Pareto frontier corresponding to the desired cost limit $d$.

The local slope of the frontier, $\frac{dR}{dC}$, captures the marginal trade-off between return and cost. Intuitively, it indicates how much additional return can be gained per unit increase in allowed cost. A steep slope indicates that a small reduction in cost would lead to a large decrease in return; in other words, the constraint is highly restrictive in this region. Vice versa, flatter regions of the curve correspond to regimes where tightening the cost constraint has only a limited effect on return.


\section{Experimental setup}
\label{sec:Experimental Setup}
Experiments are conducted on eight benchmark tasks from the Safety Gymnasium task suite \citep{ji_safety-gymnasium_2024}, spanning different levels of task and constraint complexity. These tasks consist of four safe navigation tasks and four safe velocity tasks. 

To construct empirical PFs of the return-cost plane for each task, we train a set of 25 distinct policies with a manually fixed Lagrange multiplier, by sweeping $\lambda$ over the set $\lambda \in \{10^{\ell_i}\}_{i=1}^{25}$, where $\ell_i$ are evenly spaced in $(-1,1)$. For each value of $\lambda$, we compute the average return and cost over the final 5\% of training. The collection of points induced by sweeping over $\lambda$ forms an empirical PF in the return-cost plane. Each fixed $\lambda$ yields a point on this frontier, representing the approximate optimal return-cost trade-off under that particular $\lambda$.

Furthermore, we compare the performance of the fixed multiplier $\lambda^*$ at the cost limit, the GA-updated $\lambda$ and PID-updated $\lambda$ with each other. We train 10 seeds of models for $3.5\cdot 10^7$ timesteps each. To ensure that multiple levels of cost restrictiveness are taken into account, we analyze models trained at a cost limit of 10.0 and 25.0 for navigation tasks, and at 25.0 and 400.0 for velocity tasks.


We train the Lagrangian version of PPO \citep{ray_benchmarking_nodate, schulman_proximal_2017} using the Omnisafe benchmark suite \citep{ji_omnisafe_nodate}. PPO-Lag is employed for the GA-update of the Lagrange multiplier, and CPPO-PID for the PID-controlled update \citep{stooke_responsive_2020}. In total, we performed $2480$ individual runs for our experiments The full set of experiments is mostly CPU-heavy and required approximately $57\cdot10^3$ CPU core-hours of compute.
All additional details regarding the experimental setup can be found in the Supplementary Materials Section \ref{sec: supp: details on experimental setup}. Code: \url{https://github.com/lindsayspoor/Lagrangian_SafeRL}.

\section{Results}
\label{sec:Results}

\begin{figure}[H]
  \centering

  \begin{subfigure}{0.33\textwidth}
    \centering
    \includegraphics[width=1\linewidth]{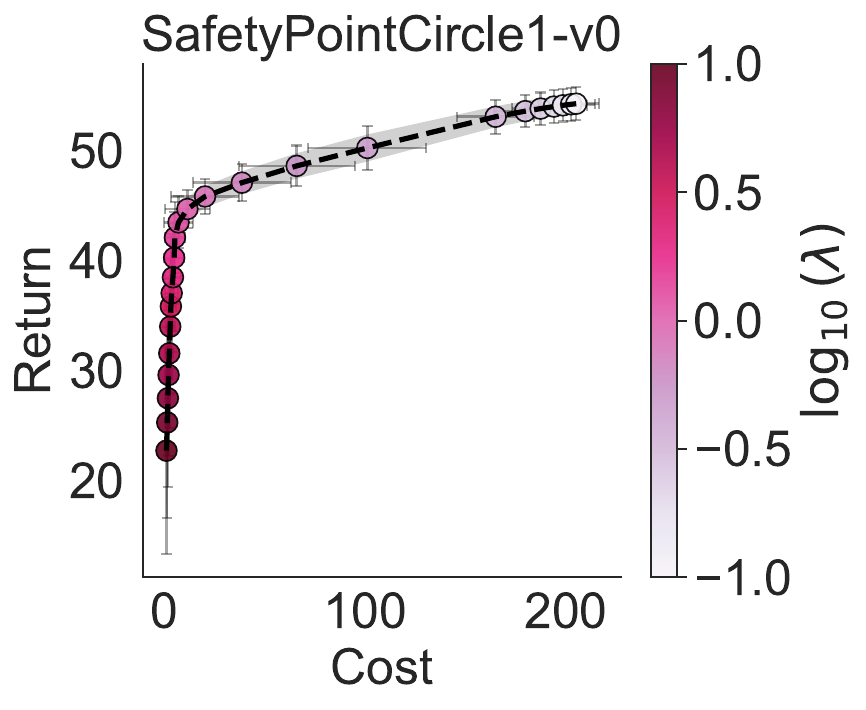}
    \caption{}
    \label{fig:pareto_rsi_circle}
  \end{subfigure}\hfill
  \begin{subfigure}{0.33\textwidth}
    \centering
    \includegraphics[width=1\linewidth]{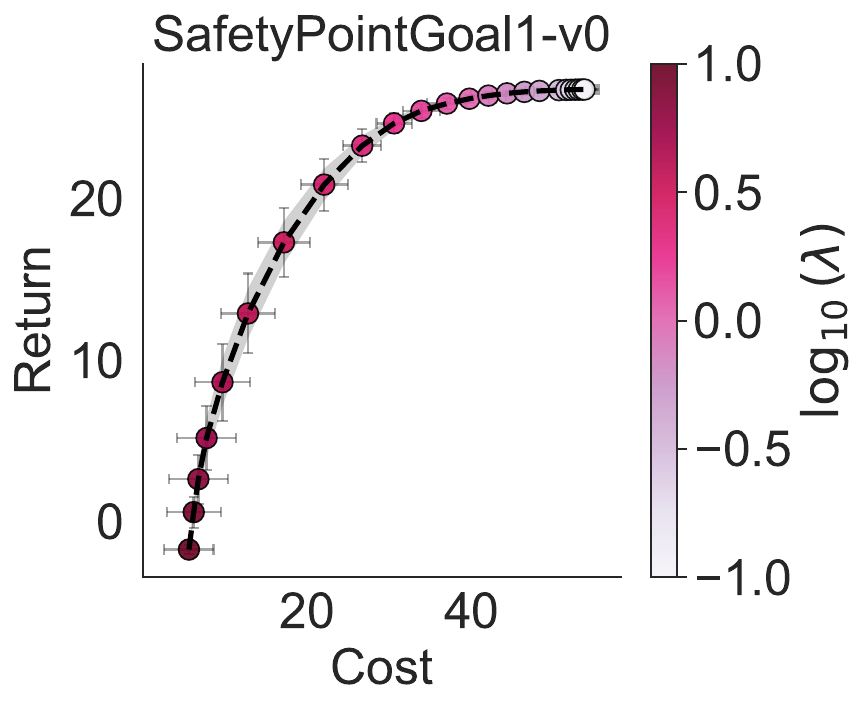}
    \caption{}
    \label{fig:pareto_rsi_goal}
  \end{subfigure}\hfill
  \begin{subfigure}{0.33\textwidth}
    \centering
    \includegraphics[width=1\linewidth]{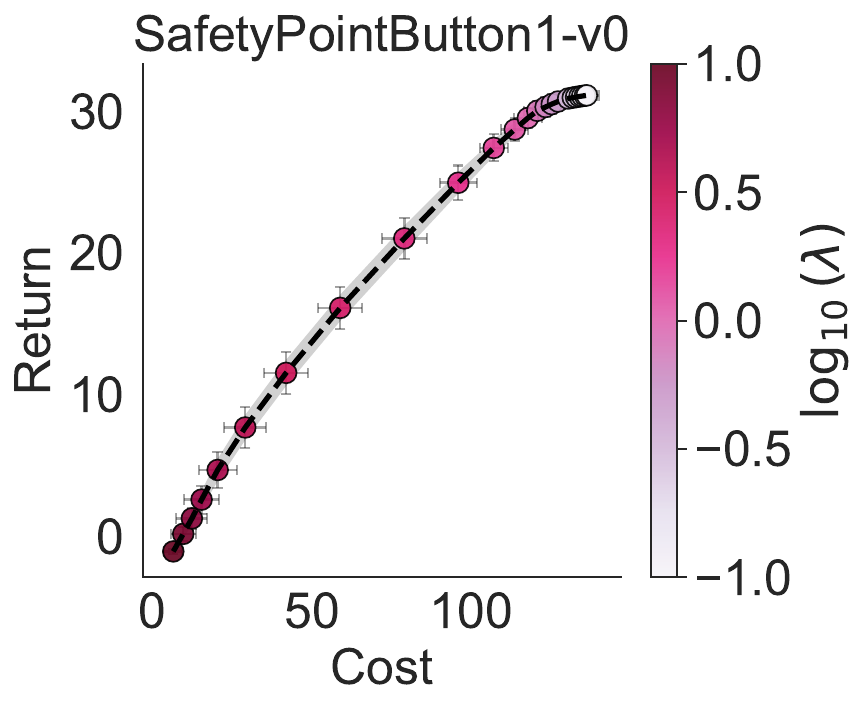}
    \caption{}
    \label{fig:pareto_rsi_button}
  \end{subfigure}
  \vspace{0.5em}
  \begin{subfigure}{0.33\textwidth}
    \centering
    \includegraphics[width=1\linewidth]{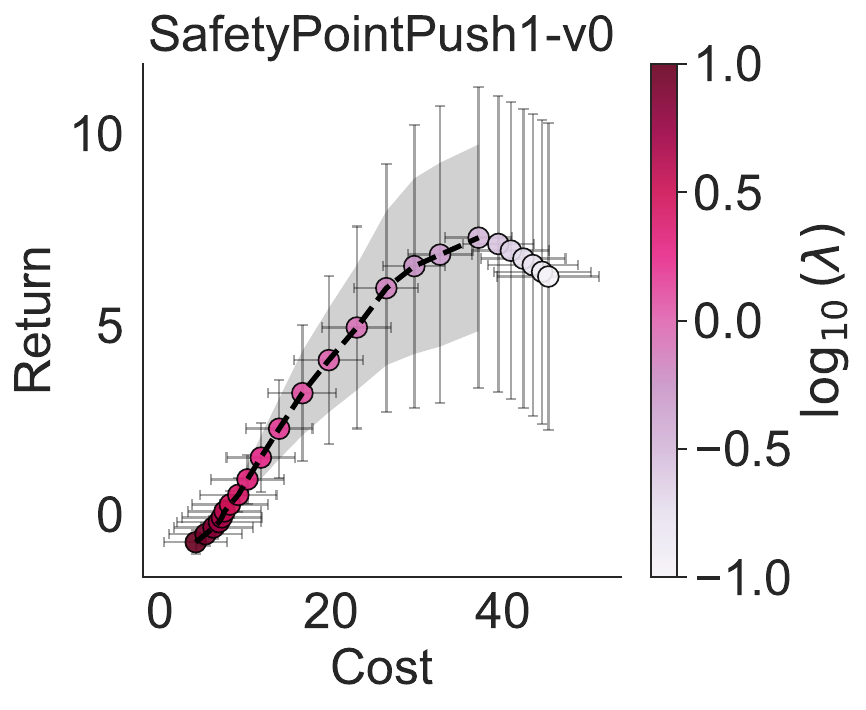}
    \caption{}
    \label{fig:pareto_rsi_push}
  \end{subfigure}\hfill
  \begin{subfigure}{0.33\textwidth}
    \centering
    \includegraphics[width=1\linewidth]{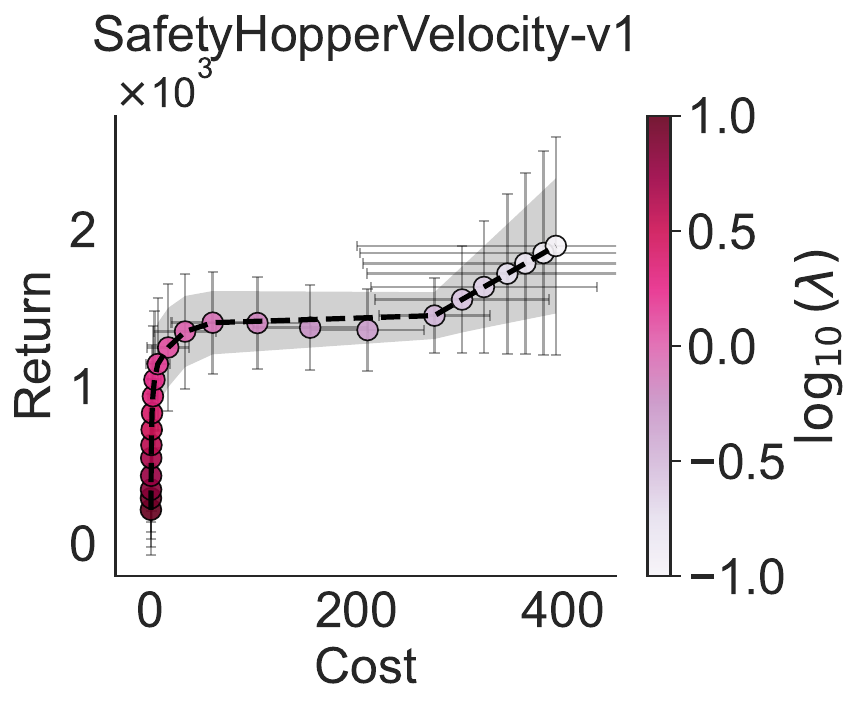}
    \caption{}
    \label{fig:pareto_rsi_hopper}
  \end{subfigure}\hfill
  \begin{subfigure}{0.33\textwidth}
    \centering
    \includegraphics[width=1\linewidth]{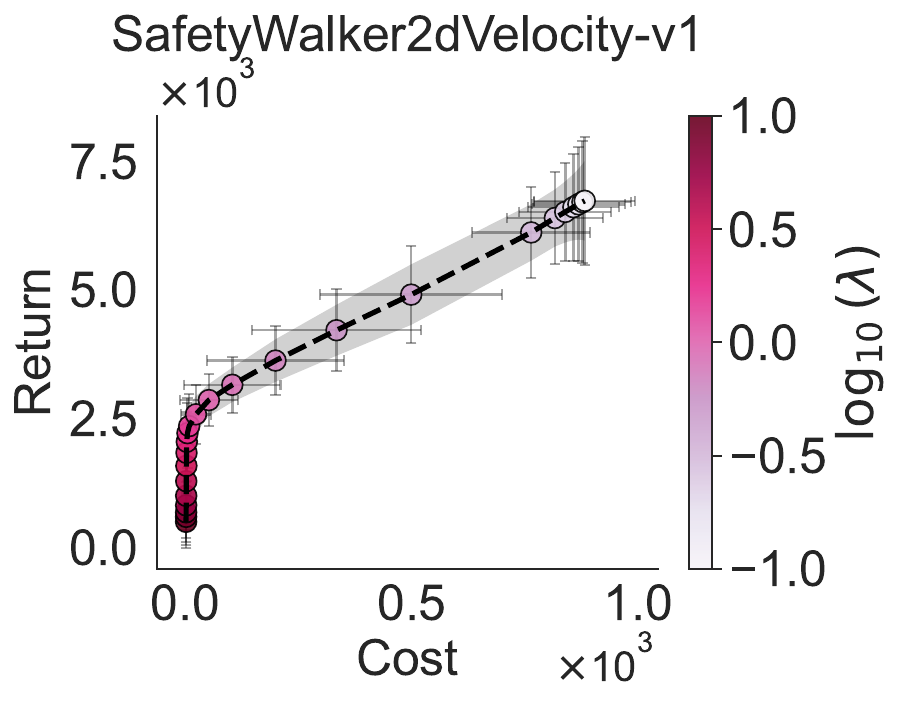}
    \caption{}
    \label{fig:pareto_rsi_walker2d}
  \end{subfigure}
  \vspace{0.5em}
  \begin{subfigure}{0.33\textwidth}
    \centering
    \includegraphics[width=1\linewidth]{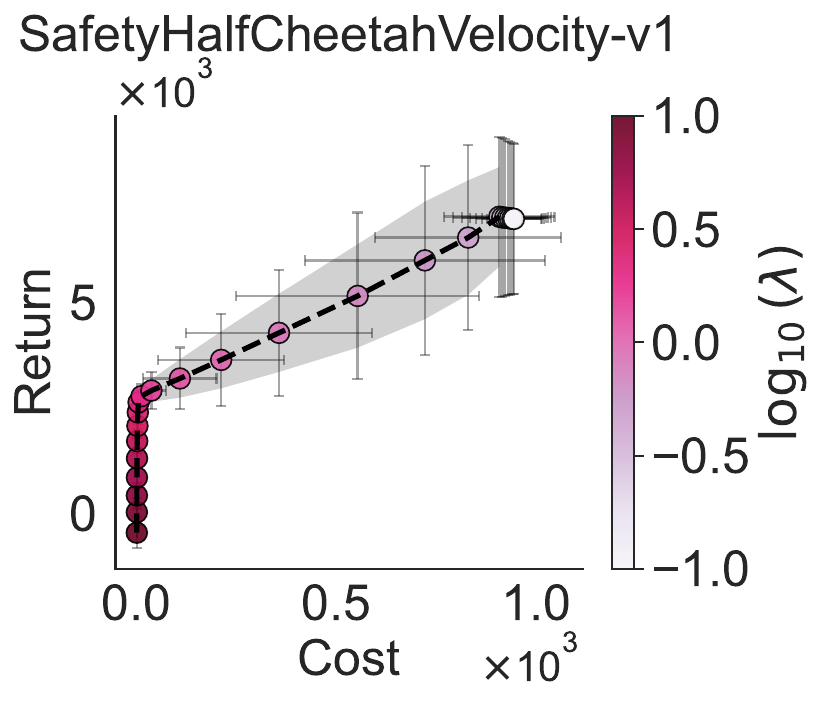}
    \caption{}
    \label{fig:pareto_rsi_halfcheetah}
  \end{subfigure}
  \begin{subfigure}{0.33\textwidth}
    \centering
    \includegraphics[width=1\linewidth]{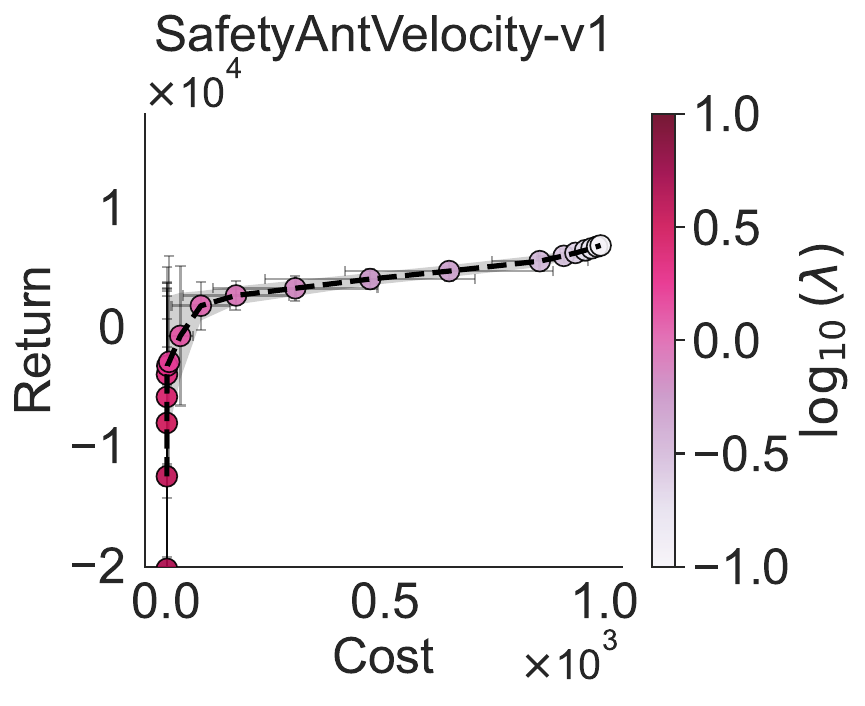}
    \caption{}
    \label{fig:pareto_rsi_ant}
  \end{subfigure}

  \caption{Smoothed empirical PFs of return versus cost as a function of $\lambda$, averaged over 10 seeds. Error bars denote the $1\sigma$ across seeds, and the shaded region shows the 95\% confidence interval of the mean empirical PF. The local slope of each curve captures the sensitivity of the return with respect to changes in the allowed cost. E.g., the geometries of SafetyPointCircle1-v0 (\ref{fig:pareto_rsi_circle}), SafetyPointButton1-v0 (\ref{fig:pareto_rsi_button})  and SafetyHopperVelocity-v1 (\ref{fig:pareto_rsi_hopper}) highlight three distinct patterns. Their shapes differ both across and within the tasks themselves, indicating that the restrictiveness of cost limits changes non-linearly.} 
  \label{fig:pareto_curves_rsi}
\end{figure}

Figure \ref{fig:pareto_curves_rsi} shows the empirical Pareto frontiers obtained by sweeping over $\lambda$, which reveal differences in the constraint geometry across the evaluated tasks. We then study the GA and PID update mechanisms and their sensitivity to different constraint regimes and present the results in Table \ref{tab:comparison_table}.

\textbf{Constraint geometry} \hspace{1em}
\label{sec:Constraint geometry}
Figure \ref{fig:pareto_curves_rsi} visualizes the empirical PFs for all evaluated tasks directly by plotting return as a function of cost. The local slope of the curve, $\frac{dR}{dC}$, captures the sensitivity of return to changes in the allowed cost. We highlight three distinct patterns we observe in the PFs: In Figure \ref{fig:pareto_rsi_circle}, the region with a cost \footnotesize $\lesssim$ \normalsize  $20$ has a steep slope, meaning that the cost is highly restrictive in this region. In other words, decreasing the allowed cost would result in a huge decrease in return in this region. On the other hand, the region with a cost \footnotesize $\gtrsim$ \normalsize $20$ is less restrictive, where a slight increase in the allowed cost yields only marginal improvements in return. In Figure \ref{fig:pareto_rsi_button}, the slope of the curve is relatively consistent across all values of $\lambda$, indicating that the restrictiveness of the cost limit acts relatively linear. Figure \ref{fig:pareto_rsi_hopper} shows an example of non-monotonic trade-off behavior: the region with a cost \footnotesize $\lesssim$ \normalsize $25$, having a steep slope, is highly restrictive, for $25$ \footnotesize $\lesssim$ \normalsize cost \footnotesize $\lesssim$ \normalsize $300$ it is comparatively less restrictive with a flatter curve, and for a cost \footnotesize $\gtrsim$ \normalsize $300$ the constraint becomes restrictive again.

These highlighted cases show that not merely the numerical cost limit, but rather the slope of the PF, determines how restrictive a safety requirement truly is. Comparing the navigation and velocity tasks, we observe that the slope of the PF of the velocity tasks is generally steeper than that of the navigation tasks. This can be explained by differences in the scale of the return and cost across the two task categories. Consequently, the velocity tasks exhibit a more restrictive constraint regime than the navigation tasks. Alternatively visualized, Figure \ref{fig:lambda_profiles_rsi} in Appendix \ref{sec:appendix: optimality} shows the return-cost profiles as a function of $\lambda$ for all evaluated tasks. These profiles differ substantially across tasks, both in scale and shape, indicating that the return-cost dynamics are highly task-specific.

\begin{table}[t!]
\centering
\caption{Performance in average return and constraint cost, computed over the final 5\% of training timesteps. A fixed optimal multiplier $\lambda^*$ (selected per cost limit) is compared against GA- and PID-updated multipliers ($K_P=10^{-4}, K_I=10^{-4}, K_D=0.0$), each trained under the corresponding cost limit. Navigation tasks use cost limits of 10.0 and 25.0, while velocity tasks use 25.0 and 400.0. Results are averaged over 10 seeds with standard deviations. Per task, the best-performing feasible method (satisfying the cost limit) is highlighted in green, while costs violating the limit are shown in red. The results indicate that there is no single method that consistently performs best within a task, but rather depends on the chosen cost limit.}
\label{tab:comparison_table}

\renewcommand{\arraystretch}{1}
\definecolor{lightgreen}{RGB}{220,245,220}
\definecolor{lightred}{RGB}{255,220,220}
\definecolor{darkgreen}{RGB}{160,220,160}
\definecolor{midgreen}{RGB}{190,233,190}

\resizebox{\textwidth}{!}{%
\begin{tabular}{
>{\centering\arraybackslash}m{2cm}
>{\centering\arraybackslash}m{1.1cm}
!{\vrule width 1pt}
C{2.6cm} C{2.4cm} C{2.4cm}
!{\vrule width 1pt}
C{2.5cm} C{2.5cm} C{2.5cm}
}

\multicolumn{8}{c}{\large \textbf{Navigation tasks}} \vspace{0.3em}\\
\Xhline{1.2pt}

&
& \multicolumn{3}{c!{\vrule width 1pt}}{\textbf{Cost limit = 10.0}}
& \multicolumn{3}{c}{\textbf{Cost limit = 25.0}} \\

&
& fixed $\lambda^*$ & GA & PID
& fixed $\lambda^*$ & GA & PID \\
\Xhline{1pt}

\multirow[t]{2}{2.2cm}{\centering SafetyPoint\\Circle1-v0}
& \vspace{0.3em} Return
&  \cellcolor{midgreen} $\mathbf{43.42}$ {\footnotesize $\pm \mathbf{3.35}$}
&  $42.43$ {\footnotesize $\pm 5.25$}
&  $41.30$ {\footnotesize $\pm 5.06$}
&  $45.16$ {\footnotesize $\pm 1.06$}
&  $41.80$ {\footnotesize $\pm 10.25$}
&  \cellcolor{midgreen} $\mathbf{46.10}$ {\footnotesize $\pm \mathbf{0.96}$} \\ 

& \vspace{0.5em} Cost
&  \cellcolor{midgreen}$3.91$ {\footnotesize $\pm 2.60$}
&  $9.61$ {\footnotesize $\pm 2.43$}
&  $4.10$ {\footnotesize $\pm 4.02$}
&  $4.75$ {\footnotesize $\pm 3.93$}
&  $19.94$ {\footnotesize $\pm 7.73$}
& \cellcolor{midgreen} $23.37$ {\footnotesize $\pm 6.63$} \\
\hline

\multirow[t]{2}{2.2cm}{\centering SafetyPoint\\Goal1-v0}
& \vspace{0.3em} Return
& $5.80$ {\footnotesize $\pm 2.77$}
& \cellcolor{midgreen} $\mathbf{6.62}$ {\footnotesize $\pm \mathbf{3.12}$}
& $22.34$ {\footnotesize $\pm 1.64$}
& $23.80$ {\footnotesize $\pm 0.92$}
& $22.83$ {\footnotesize $\pm 1.85$}
& $24.73$ {\footnotesize $\pm 0.70$} \\

& \vspace{0.5em} Cost
&  $8.49$ {\footnotesize $\pm 4.84$}
& \cellcolor{midgreen} $9.06$ {\footnotesize $\pm 2.46$}
& \cellcolor{lightred} $27.47$ {\footnotesize $\pm 2.13$}
& \cellcolor{lightred} $25.48$ {\footnotesize $\pm 1.82$}
& \cellcolor{lightred} $25.52$ {\footnotesize $\pm 0.56$}
& \cellcolor{lightred} $30.84$ {\footnotesize $\pm 1.98$} \\
\hline

\multirow[t]{2}{2.2cm}{\centering SafetyPoint\\Button1-v0}
& \vspace{0.3em} Return
& $0.28$ {\footnotesize $\pm 0.59$}
& \cellcolor{midgreen} $\mathbf{0.47}$ {\footnotesize $\pm \mathbf{0.61}$}
& $3.83$ {\footnotesize $\pm 0.95$}
& $3.79$ {\footnotesize $\pm 1.12$}
& $5.59$ {\footnotesize $\pm 0.93$}
& \cellcolor{midgreen} $\mathbf{5.41}$ {\footnotesize $\pm \mathbf{1.58}$} \\

& \vspace{0.5em} Cost
& \cellcolor{lightred} $11.39$ {\footnotesize $\pm 2.08$}
& \cellcolor{midgreen} $7.78$ {\footnotesize $\pm 1.51$}
& \cellcolor{lightred} $15.56$ {\footnotesize $\pm 4.34$}
& $14.11$ {\footnotesize $\pm 4.93$}
& \cellcolor{lightred} $26.73$ {\footnotesize $\pm 2.30$}
& \cellcolor{midgreen} $24.85$ {\footnotesize $\pm 8.39$} \\
\hline

\multirow[t]{2}{2.2cm}{\centering SafetyPoint\\Push1-v0}
& \vspace{0.3em} Return
& $0.88$ {\footnotesize $\pm 0.63$}
& \cellcolor{midgreen}$\mathbf{0.59}$ {\footnotesize $\pm \mathbf{0.41}$}
& $1.98$ {\footnotesize $\pm 1.56$}
& $4.96$ {\footnotesize $\pm 3.52$}
& $4.38$ {\footnotesize $\pm 2.05$}
& \cellcolor{midgreen}$\mathbf{5.49}$ {\footnotesize $\pm \mathbf{2.84}$} \\

& \vspace{0.5em} Cost
& \cellcolor{lightred} $11.88$ {\footnotesize $\pm 2.42$}
& \cellcolor{midgreen} $8.88$ {\footnotesize $\pm 3.66$}
& \cellcolor{lightred} $10.68$ {\footnotesize $\pm 2.26$}
& \cellcolor{lightred} $28.05$ {\footnotesize $\pm 5.85$}
& \cellcolor{lightred} $25.35$ {\footnotesize $\pm 2.37$}
& \cellcolor{midgreen} $20.36$ {\footnotesize $\pm 3.27$} \\
\Xhline{1.5pt}

\\
\multicolumn{8}{c}{\large \textbf{Velocity tasks}}\vspace{0.3em} \\
\Xhline{1.2pt}

&
& \multicolumn{3}{c!{\vrule width 1pt}}{\textbf{Cost limit = 25.0}}
& \multicolumn{3}{c}{\textbf{Cost limit = 400.0}} \\

&
& fixed $\lambda^*$ & GA & PID
& fixed $\lambda^*$ & GA & PID \\
\Xhline{1pt}


\multirow[t]{2}{2.2cm}{\centering SafetyHopper\\Velocity-v1}
& \vspace{0.3em} Return
& $1292.34$ {\footnotesize $\pm 603.52$}
& \cellcolor{midgreen}$\mathbf{1682.52}$ {\footnotesize $\pm \mathbf{37.62}$}
& $1443.69$ {\footnotesize $\pm 510.46$}
& $1979.64$ {\footnotesize $\pm 623.42$}
& $1682.93$ {\footnotesize $\pm 140.68$}
& \cellcolor{midgreen}$\mathbf{1713.47}$ {\footnotesize $\pm \mathbf{159.91}$} \\

& \vspace{0.5em} Cost
& $6.31$ {\footnotesize $\pm 6.25$}
& \cellcolor{midgreen}$24.54$ {\footnotesize $\pm 6.23$}
& \cellcolor{lightred}$26.56$ {\footnotesize $\pm 15.69$}
& \cellcolor{lightred}$428.91$ {\footnotesize $\pm 198.81$}
& $357.75$ {\footnotesize $\pm 35.00$}
&\cellcolor{midgreen}$347.49$ {\footnotesize $\pm 44.96$} \\
\hline

\multirow[t]{2}{2.2cm}{\centering SafetyWalker2d\\Velocity-v1}
& \vspace{0.3em} Return
& $2668.69 $ {\footnotesize $\pm 60.60$}
& \cellcolor{midgreen}$\mathbf{3127.55}$ {\footnotesize $\pm \mathbf{76.92}$}
& $3132.17$ {\footnotesize $\pm 82.10$}
& $4334.02$ {\footnotesize $\pm 1189.20$} 
& \cellcolor{midgreen}$\mathbf{3399.87}$ {\footnotesize $\pm \mathbf{418.21}$} 
& $4089.29$ {\footnotesize $\pm 610.61$} \\

& \vspace{0.5em} Cost
&  $2.96$ {\footnotesize $\pm 3.82$}
& \cellcolor{midgreen} $22.64$ {\footnotesize $\pm 5.61$}
& \cellcolor{lightred} $29.82$ {\footnotesize $\pm 12.71$}
& \cellcolor{lightred} $410.22$ {\footnotesize $\pm 318.62$} 
& \cellcolor{midgreen} $381.70$ {\footnotesize $\pm 58.99$} 
& \cellcolor{lightred} $417.78$ {\footnotesize $\pm 25.41$} \\
\hline

\multirow[t]{2}{2.2cm}{\centering \small SafetyHalfCheetah\\Velocity-v1}
& \vspace{0.3em} Return
& $2861.85$ {\footnotesize $\pm 66.76$}
& $2745.51$ {\footnotesize $\pm 540.52$}
& \cellcolor{midgreen}$\mathbf{2965.10}$ {\footnotesize $\pm \mathbf{86.61}$}
& \cellcolor{midgreen}$\mathbf{4088.51}$ {\footnotesize $\pm \mathbf{1972.41}$}
& $2940.21$ {\footnotesize $\pm 534.06$} 
& $3365.50$ {\footnotesize $\pm 818.29$} \\

& \vspace{0.5em} Cost
&  $4.82$ {\footnotesize $\pm 2.95$}
&  $22.26$ {\footnotesize $\pm 15.09$}
& \cellcolor{midgreen} $16.76$ {\footnotesize $\pm 28.28$}
& \cellcolor{midgreen} $283.79$ {\footnotesize $\pm 427.59$} 
&  $315.36$ {\footnotesize $\pm 190.69$} 
&  $386.97$ {\footnotesize $\pm 62.40$} \\
\hline

\multirow[t]{2}{2.2cm}{\centering SafetyAnt\\Velocity-v1}
& \vspace{0.3em} Return
& $-2602.45$ {\footnotesize $\pm 6697.29$}
& $3387.12$ {\footnotesize $\pm 22.61$}
& \cellcolor{midgreen}$\mathbf{3332.30}$ {\footnotesize $\pm \mathbf{72.87}$}
& $4023.17$ {\footnotesize $\pm 981.96$} 
& \cellcolor{midgreen}$\mathbf{2727.06}$ {\footnotesize $\pm \mathbf{1054.88}$} 
& $2204.24$ {\footnotesize $\pm 4391.94$} \\

& \vspace{0.5em} Cost
&  $0.52$ {\footnotesize $\pm 0.95$}
& \cellcolor{lightred} $27.75$ {\footnotesize $\pm 4.96$}
& \cellcolor{midgreen}$18.60$ {\footnotesize $\pm 7.60$}
& \cellcolor{lightred} $506.27$ {\footnotesize $\pm 295.64$} 
& \cellcolor{midgreen} $323.48$ {\footnotesize $\pm 162.65$} 
&  $376.74$ {\footnotesize $\pm 132.07$} \\
\Xhline{1.2pt}

\end{tabular}
}
\end{table}

\textbf{Constraint-regime sensitivity} \hspace{1em}
\label{sec:Performance comparison}
Table \ref{tab:comparison_table} shows the performance of all three methods in terms of average return and cost, computed over the last 5\% of training. The best-performing method in each setting is highlighted in green. Since we consider only those methods that satisfy the specified cost limit, we observe that the \textit{best} method does not always refer to the method with the highest return. We remark that there is no best-performing method for SafetyPointGoal1-v0 at a cost limit of 25.0. We observe that the optimal method differs both across tasks and across cost limits within the same task, indicating that the constraint restrictiveness influences which update algorithm performs best.

Costs that violate the specified cost limit are highlighted in red. In some cases, a method overshoots or undershoots the target cost. For example, for the fixed $\lambda^*$ method on the SafetyPointCircle1-v0 task, the cost limit was set to 25.0, while the achieved cost was $4.75 \pm 3.93$. Despite the observed deviations from the specified cost limits, the corresponding performance still approaches the empirical PF. This is illustrated in Figures \ref{fig:pareto_curves_rsi_zoomed_combined} and \ref{fig:pareto_curves_rsi_zoomed_combined_10_400} in Supplementary Materials Section \ref{sec:supp: additional results}, which project the results from Table \ref{tab:comparison_table} onto the empirical PFs. Small deviations of the fixed $\lambda^*$ from the PF arise from smoothing. The projected GA and PID results further show that no single method consistently performs best across or within tasks. Moreover, the substantial variance reported in Table~\ref{tab:comparison_table}, even across 10 seeds, indicates that methods performing well on average may still produce individual runs that violate the constraint, while methods with lower average performance may occasionally satisfy it. This variability raises the question of whether practitioners should impose stricter cost limits during training to increase the likelihood of satisfying the desired constraint, given that Lagrangian methods are not strict constraint enforcers by design.


\section{Discussion \& conclusion}
\label{sec:Discussion & Conclusion}

\textbf{Limitations} \hspace{1em} The empirical PFs in Figure \ref{fig:pareto_curves_rsi} reveal the constraint geometry of the underlying optimization problems and demonstrate that these insights can substantially improve the interpretation and evaluation of safe RL methods by evaluating for multiple cost limits. We point out that these geometric insights do not directly reflect the intrinsic difficulty of learning the tasks; regimes with a steeper slope are not necessarily harder to solve, but rather indicate that the safety requirement is more restrictive.

Moreover, although Lagrangian methods can be combined with any standard RL algorithm, we focused on PPO, which remains state-of-the-art and is the most widely used algorithm for Lagrangian approaches in practice. As such, PPO provides a representative and practically relevant testbed for studying the behavior of the Lagrange multiplier. Though, empirical results may vary across algorithms. We therefore present preliminary results for Lagrangian-based SAC in Supplementary Section \ref{sec: supp: saclagrsi} \citep{haarnoja2018soft}. These indicate that off-policy algorithms produce constraint geometries different from on-policy methods, motivating further evaluation of our approach across a broader set of Lagrangian-based algorithms.

In addition, the PID-controlled update mechanism was evaluated under a single parameter configuration ($K_P = 10^{-4}$, $K_I = 10^{-4}$, $K_D = 0$), corresponding to PI-control. While this reflects the most commonly used practical configuration, broader exploration of PID-parameters would provide deeper insight into the dynamics and robustness of Lagrange multiplier updates.

\textbf{Conclusion} \hspace{1em} 
In this work, we empirically studied the constraint geometry of eight tasks from the Safety Gymnasium suite \citep{ji_safety-gymnasium_2024}, one of the most widely used and standardized benchmarks for safe RL. These tasks were selected to capture diversity in return–cost dynamics, constraint cost restrictiveness, and levels of task complexity, providing a sufficiently broad empirical basis for the results presented in this study. Given the diversity of safe RL domains, we encourage the benchmark community to accompany published benchmarks with analyses of empirical PFs and to extend such analyses to additional domains, such as visual safety tasks \citep{tomilin_hasard_2025} or operational control benchmarks \citep{ramanujam_safeor-gym_2025}. We release an open-source code base that enables researchers to reproduce and extend our experiments. 
Since the quality of an algorithm can not be properly assessed by evaluating a single point on the PF, we recommend evaluating Lagrangian safe RL methods across multiple cost limits to capture different levels of restrictiveness. Accordingly, we provide recommended cost limits for each task in Appendix Section \ref{sec:app: reccomendations on cost limits}, and encourage the community to adopt similar practices when proposing new safe RL benchmark tasks.

\appendix

\renewcommand{\thesection}{A.\arabic{section}}
\renewcommand{\thesubsection}{A.\arabic{section}.\arabic{subsection}}
\renewcommand{\thefigure}{A.\arabic{figure}}
\renewcommand{\thetable}{A.\arabic{table}}

\setcounter{section}{0}
\setcounter{figure}{0}
\setcounter{table}{0}

\newpage

\section{Additional results}

\label{sec:appendix: optimality}
\begin{figure}[H]
  \centering

  \begin{subfigure}{0.33\textwidth}
    \centering
    \includegraphics[width=1\linewidth]{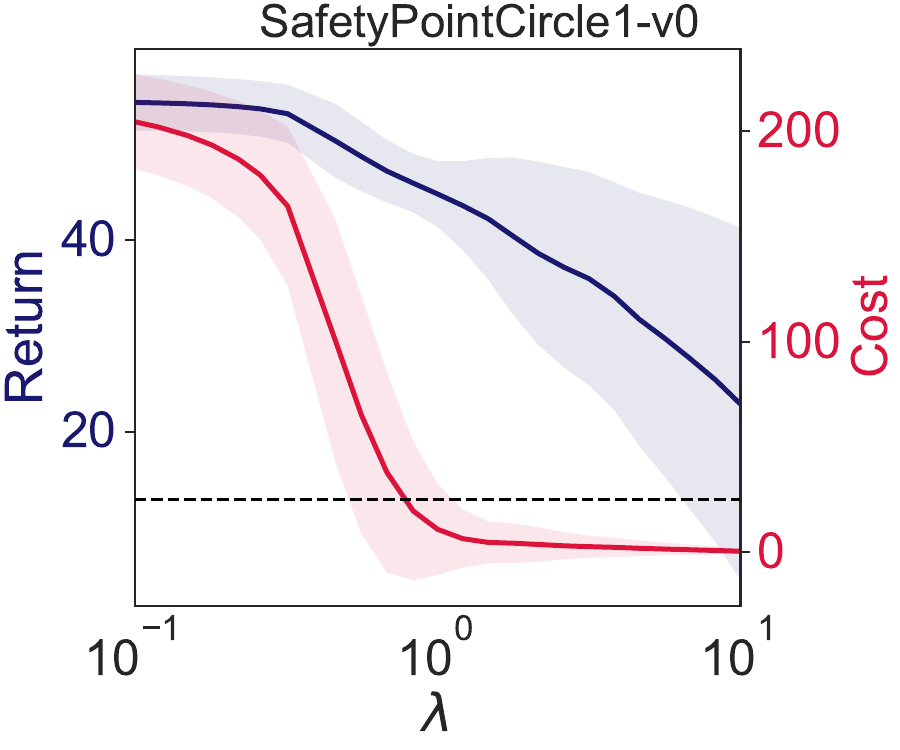}
    \caption{}
    \label{fig:lambda_profile_rsi_circle}
  \end{subfigure}\hfill
  \begin{subfigure}{0.33\textwidth}
    \centering
    \includegraphics[width=1\linewidth]{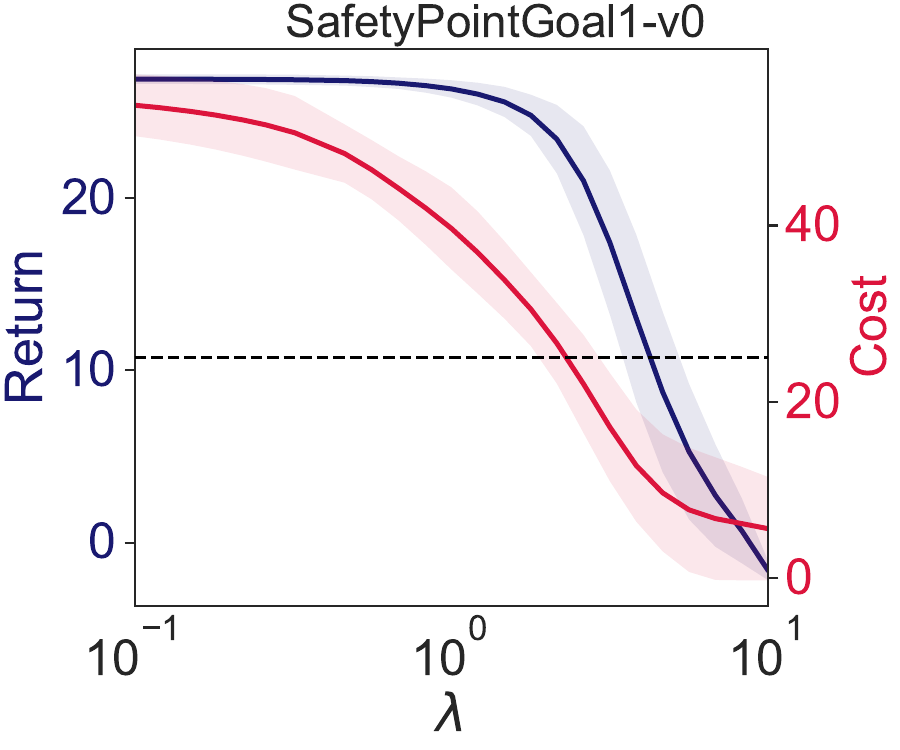}
    \caption{}
    \label{fig:lambda_profile_rsi_goal}
  \end{subfigure}\hfill
  \begin{subfigure}{0.33\textwidth}
    \centering
    \includegraphics[width=1\linewidth]{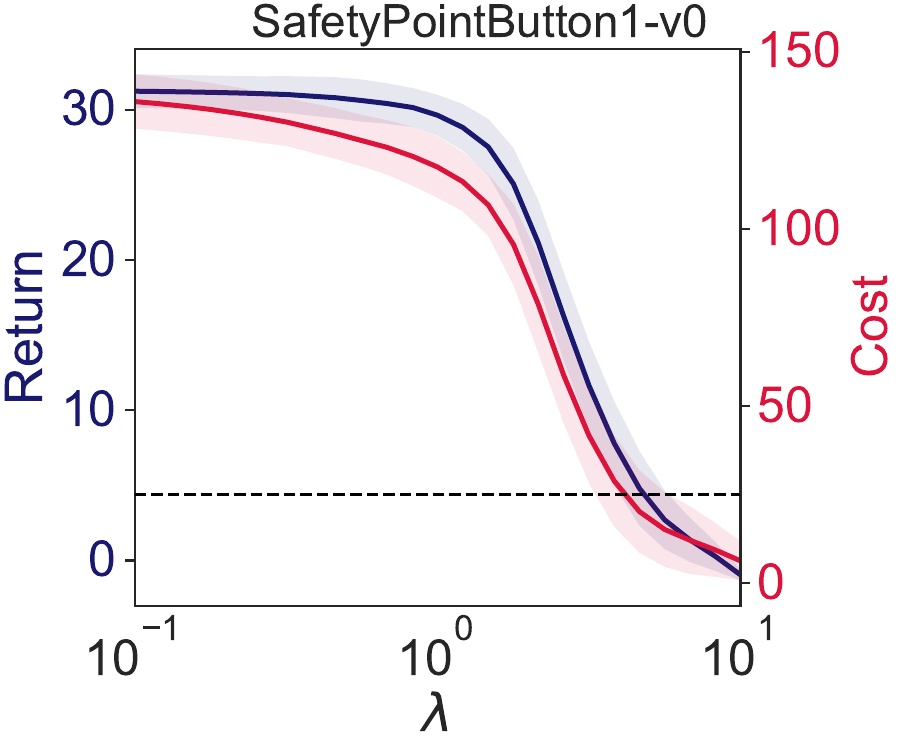}
    \caption{}
    \label{fig:lambda_profile_rsi_button}
  \end{subfigure}
  \vspace{0.5em}
  \begin{subfigure}{0.33\textwidth}
    \centering
    \includegraphics[width=1\linewidth]{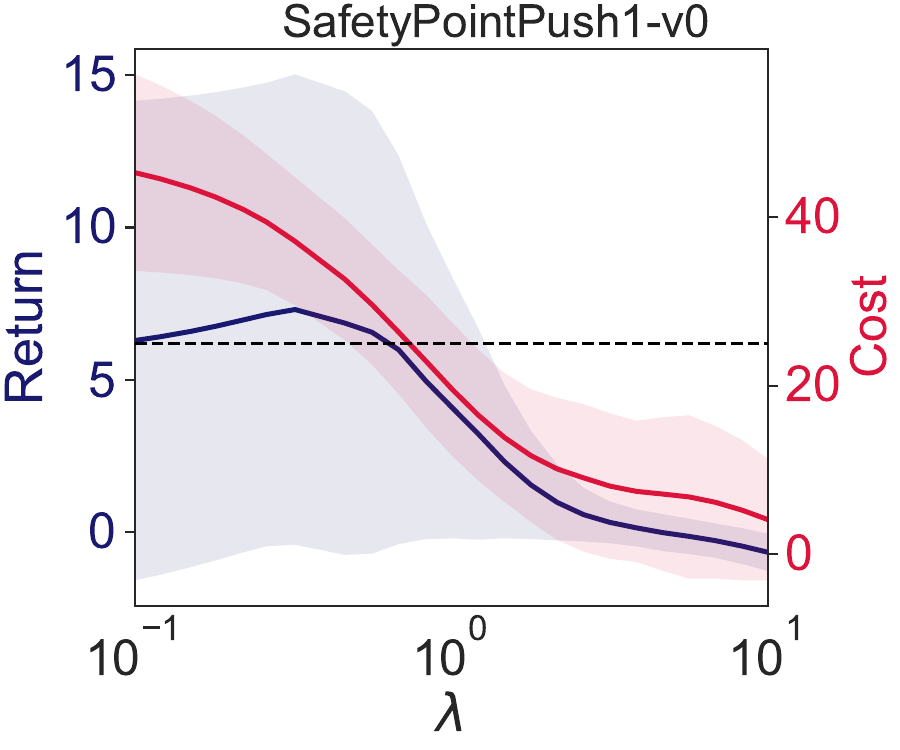}
    \caption{}
    \label{fig:lambda_profile_rsi_push}
  \end{subfigure}
  \begin{subfigure}{0.33\textwidth}
    \centering
    \includegraphics[width=1\linewidth]{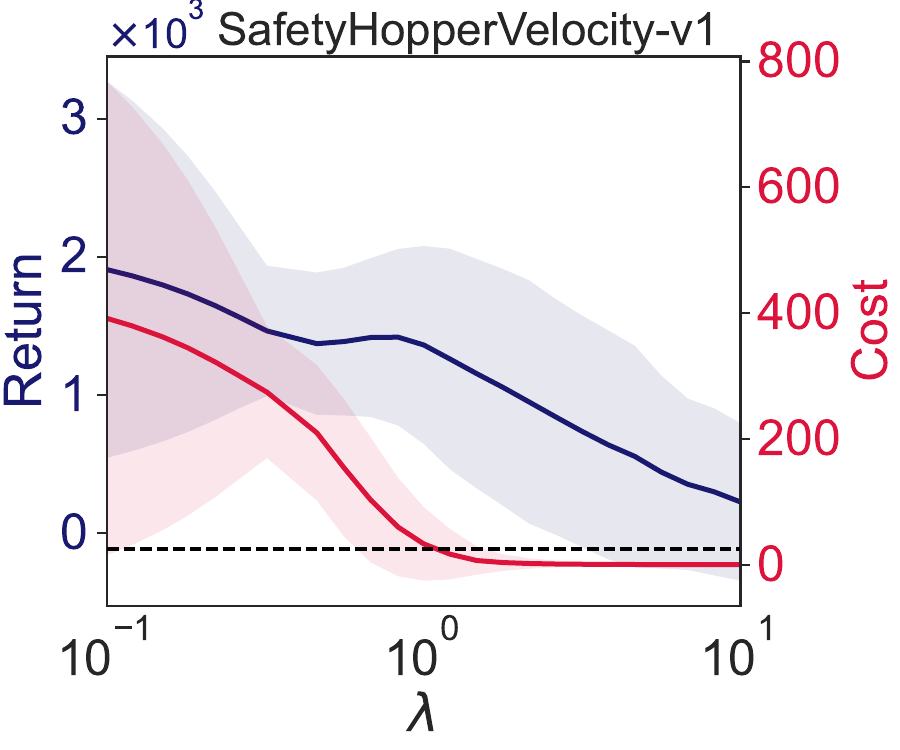}
    \caption{}
    \label{fig:lambda_profile_rsi_hopper}
  \end{subfigure}\hfill
  \begin{subfigure}{0.33\textwidth}
    \centering
    \includegraphics[width=1\linewidth]{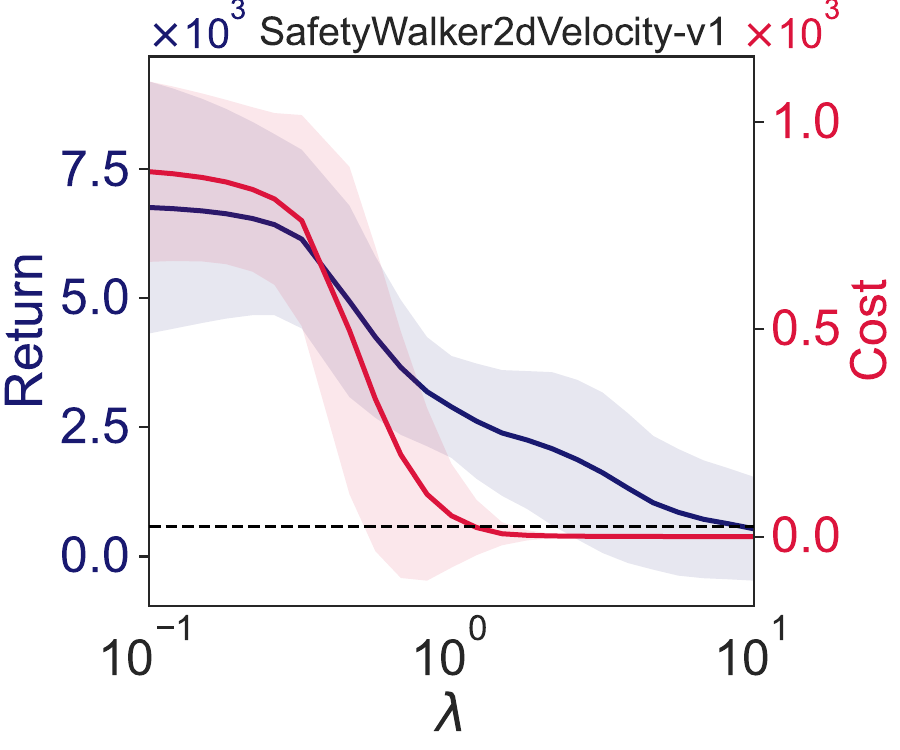}
    \caption{}
    \label{fig:lambda_profile_rsi_walker2d}
  \end{subfigure}
  \vspace{0.5em}
  \begin{subfigure}{0.33\textwidth}
    \centering
    \includegraphics[width=1\linewidth]{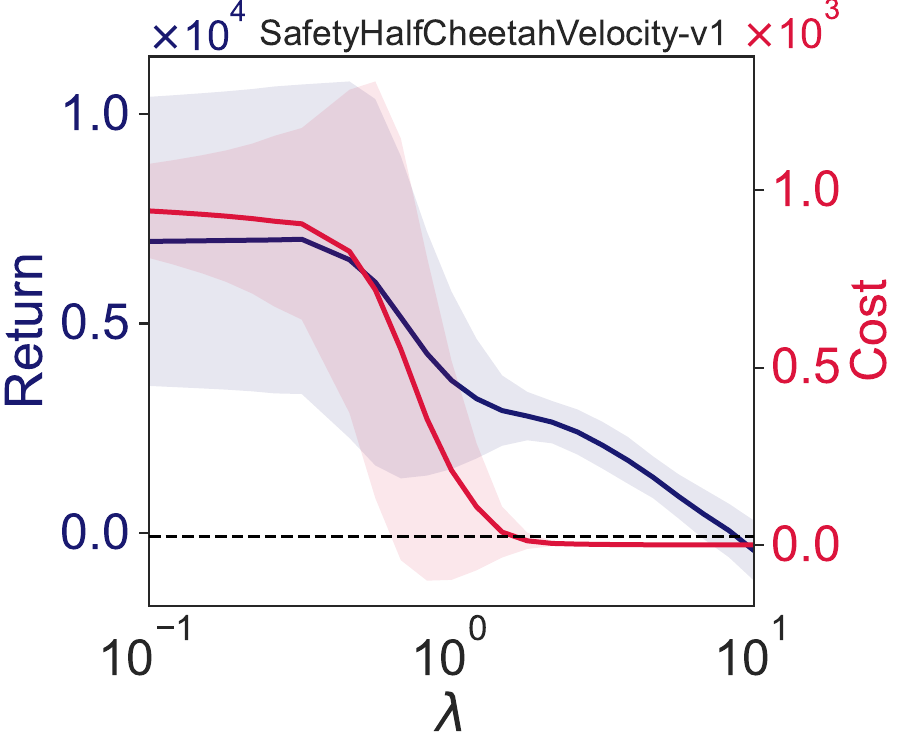}
    \caption{}
    \label{fig:lambda_profile_rsi_halfcheetah}
  \end{subfigure}
  \begin{subfigure}{0.33\textwidth}
    \centering
    \includegraphics[width=1\linewidth]{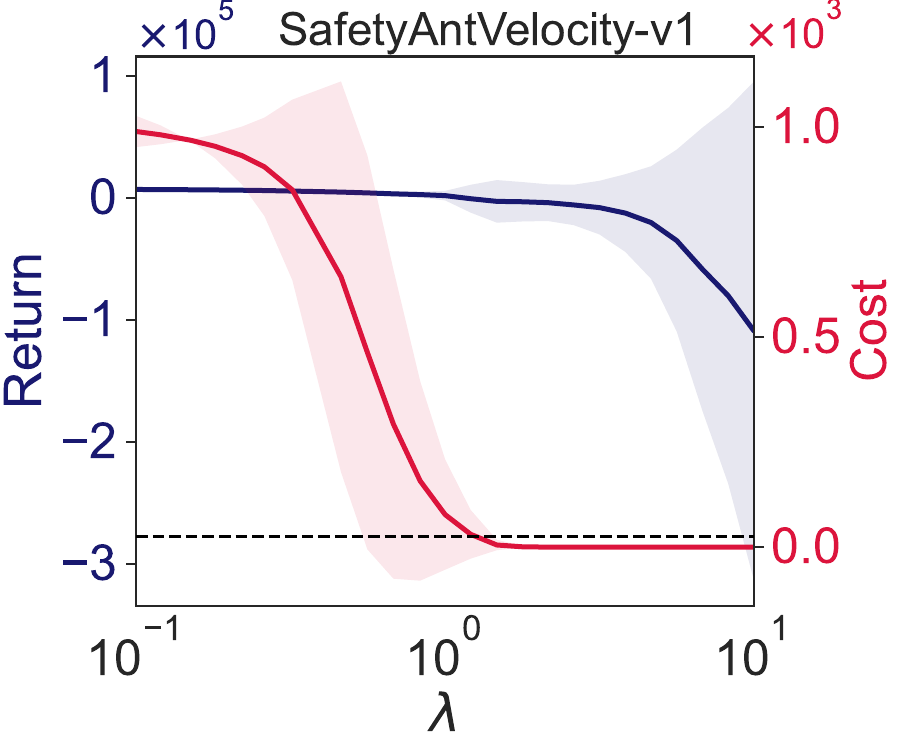}
    \caption{}
    \label{fig:lambda_profile_rsi_ant}
  \end{subfigure}

  \caption{Smoothened $\lambda$-profiles averaged over 10 seeds, with the shaded regions indicating the 95\% confidence intervals of the return and cost, for all evaluated tasks. A cost limit of 25.0 is indicated by the dashed lines. These $\lambda$-profiles offer practical guidance on tuning $\lambda$ with respect to a desired cost limit. The return- and cost-curves are monotonically decreasing with increasing $\lambda$ for all tasks, except for SafetyPointPush1-v0 (\ref{fig:lambda_profile_rsi_push}) and SafetyHopperVelocity-v1 (\ref{fig:lambda_profile_rsi_hopper}), where small regions of $\lambda$ show a slight increase in the return curve. The variance across 10 seeds is high, indicating the highly sensitive nature of $\lambda$. }
  \label{fig:lambda_profiles_rsi}
\end{figure}

We explicitly visualize the trade-off between return and cost for different fixed values of the Lagrange multiplier $\lambda$. These $\lambda$-profiles are constructed by training 10 seeds of 25 models for $3.5 \cdot 10^7$ timesteps each, where every model is trained with a distinct, manually fixed multiplier $\lambda \in \{10^{\ell_i}\}_{i=1}^{25}$, where $\ell_i$ are evenly spaced in $(-1,1)$. Note that $\lambda = 0$ would correspond to standard PPO. For each value of $\lambda$, we compute the average return and cost over the final 5\% of training, of which the results are plotted in Figure \ref{fig:lambda_profiles_rsi}. These profiles differ substantially across tasks, both in scale and shape, indicating that the return-cost dynamics are highly task-specific.

If we draw a horizontal line corresponding to a particular cost limit, e.g., 25.0 in the figure, we observe that relaxing the constraint, i.e. allowing for a higher cost, consistently increases the achievable return, except for SafetyPointPush1-v0 (\ref{fig:lambda_profile_rsi_push}) and SafetyHopperVelocity-v1 (\ref{fig:lambda_profile_rsi_hopper}), where small regions of $\lambda$ show a slight decrease in the return curve for increasing cost/decreasing $\lambda$. These two deviating results may be caused by smoothing effects across seeds with high variance. Due to the difference in shapes and ranges of the cost curves, the optimal value for the Lagrange multiplier, $\lambda^*$, i.e., where the cost curve crosses a specified cost limit, varies across tasks. This indicates that the optimal value $\lambda^*$ is highly task-dependent and dependent on the value of the cost limit. Nevertheless, these $\lambda$-profiles can offer practical guidance on tuning $\lambda$ with respect to a desired cost limit for each of the presented tasks, and offer insights on the return-cost trade-off complementary to the empirical PFs shown in Figure \ref{fig:pareto_curves_rsi} of the main paper.

\section{Recommendations on cost limits} \label{sec:app: reccomendations on cost limits}
Table \ref{tab:cost_limits} presents a list of recommended cost limits for each of the eight evaluated benchmark tasks. These cost limits are selected based on the slopes of the empirical PFs presented in Figure \ref{fig:pareto_curves_rsi} of the main paper. We selected cost limits in a restrictive region, i.e., with a steep slope, and cost limits in a less restrictive region, i.e., where the curve is flatter. Based on these different cost limits, the best-performing update mechanism for the Lagrange multiplier can vary within the same task, as is shown in Table \ref{tab:comparison_table} in the main paper. We therefore encourage the community to use these recommended cost limits when evaluating Lagrangian-based algorithms. Additionally, when proposing new safe RL benchmark tasks, we recommend providing similar lists of cost limits to offer practical guidance to benchmark users.

\begin{table}[H]
\centering
\small
\caption{Recommended cost limits per task.}
\label{tab:cost_limits}
\begin{tabular}{ll}
\toprule
\textbf{Task} & \textbf{Cost limits} \\
\midrule
SafetyPointCircle1-v0        & 20; 150 \\
SafetyPointGoal1-v0          & 15; 40 \\
SafetyPointButton1-v0        & 50; 150 \\
SafetyPointPush1-v0          & 10; 50 \\
SafetyHopperVelocity-v1      & 5; 200; 400 \\
SafetyWalker2dVelocity-v1    & 5; 400 \\
SafetyHalfCheetahVelocity-v1 & 5; 400 \\
SafetyAntVelocity-v1         & 5; 400 \\
\bottomrule
\end{tabular}
\end{table}



\bibliography{main_folder/LagrangianSafeRLnew}
\bibliographystyle{rlj}

\beginSupplementaryMaterials

\renewcommand{\thesection}{SM.\arabic{section}}
\renewcommand{\thesubsection}{SM.\arabic{section}.\arabic{subsection}}
\renewcommand{\thefigure}{SM.\arabic{figure}}
\renewcommand{\thetable}{SM.\arabic{table}}

\setcounter{section}{0}
\setcounter{figure}{0}
\setcounter{table}{0}
\setcounter{equation}{0}

\section{Details on experimental setup} \label{sec: supp: details on experimental setup}

Code: \url{https://github.com/lindsayspoor/Lagrangian_SafeRL}

\paragraph{Training details} \label{sec:supp: Training details}

We train the Lagrangian version of PPO \citep{ray_benchmarking_nodate, schulman_proximal_2017} using the Omnisafe benchmark suite \citep{ji_omnisafe_nodate}. PPO-Lag is employed for the GA-update of the Lagrange multiplier, and CPPO-PID for the PID-controlled update \citep{stooke_responsive_2020}. Each model is trained for $3.5\cdot 10^{7}$ timesteps, and all results are averaged across 10 seeds. In total, we performed $8 \textrm{ tasks } \times (25\textrm{ values of } \lambda + ($3$ \textrm{ update methods } \times 2 \textrm{ cost limits})) \times 10 \textrm{ seeds } = 2480$ individual runs for our experiments. Each run is CPU-heavy, and required approximately 23 CPU core-hours and 13.8GB of RAM. The full set of experiments required approximately $57\cdot10^3$ CPU core-hours of compute, corresponding to 148 days on a 16-core CPU.

\textbf{Reward-scale invariance}
\hspace{1em}
For all of our trained models, we implemented a scaling factor in the policy gradient \citep{stooke_responsive_2020}, according to Eq. \ref{eq:rsi_policy_gradient}. Doing so, we enforce that the total gradient will be equally balanced by $J^R$ and $J^{\mathcal{C}}$ at $\lambda=1.0$. Moreover, this ensures that the dynamics of $\lambda$-updates have a fixed meaning both within and across tasks.

\begin{equation}\label{eq:rsi_policy_gradient}
    \nabla_{\theta}\mathcal{L} = (1-u_k)\nabla_{\theta}J^R(\pi_{\theta_k})-u_k\beta_k\nabla_{\theta}J^{\mathcal{C}}(\pi_{\theta_k}),
\end{equation}

where $u_k = \frac{\lambda_k}{1+\lambda_k}$ and scaling factor $\beta_k$ is shown in Eq. \ref{eq:scaling_factor}.

\begin{equation}\label{eq:scaling_factor}
    \beta_k = \frac{||\nabla_{\theta}J^R(\pi_{\theta_k})||}{||\nabla_{\theta}J^{\mathcal{C}}(\pi_{\theta_k})||}.
\end{equation}

\textbf{Hyperparameters} 
\hspace{1em}
The hyperparameters used for the complete set of experiments in the paper are shown in Table \ref{tab:hparams}.

\begin{table}[H]
\centering
\footnotesize
\caption{Hyperparameter settings for PPO-Lag and CPPO-PID.}
\label{tab:hparams}

\begin{tabular}{@{}ll @{\hspace{1.5cm}} ll@{}}
\toprule
\multicolumn{2}{c}{\textbf{General training settings}} 
& \multicolumn{2}{c}{\textbf{Other settings}} \\
\cmidrule(r){1-2} \cmidrule(l){3-4}

Steps / epoch & 20{,}000 
& \multicolumn{2}{l}{\textbf{Reward-scale invariance}} \\

Update iterations / epoch & 20 
& $\Delta\beta$ EMA $\alpha$ & 0.9 \\

Batch size & 1024 
& \multicolumn{2}{l}{\textbf{Lagrangian update}} \\

Clip ratio & 0.2 
& $\lambda$ init & 1.0 \\

Target KL & 0.02 
& \multicolumn{2}{l}{\textbf{GA update (PPO-Lag)}} \\

Entropy coefficient & 0.0 
& $\eta$ & 0.035 \\

Advantage estimation & GAE 
& \multicolumn{2}{l}{\textbf{PID update (CPPO-PID)}} \\

$\gamma_r, \gamma_c$ & 0.99 
& PID $d$-delay & 10 \\

GAE $\lambda_r, \lambda_c$ & 0.95 
& PID $\Delta p, \Delta d$ EMA $\alpha$ & 0.95 \\

Actor network & [512, 512], ELU 
& Penalty max & 100 \\

Critic network & [512, 512], ELU 
& $K_P$ & $10^{-4}$ \\

Actor \& Critic LR & $3\times 10^{-4}$ 
& $K_I$ & $10^{-4}$ \\

& 
& $K_D$ & $0.0$ \\

\bottomrule
\end{tabular}
\end{table}

\paragraph{Evaluated tasks} \label{sec: supp: evaluated tasks}

All experiments are conducted with the Omnisafe benchmark suite \citep{ji_omnisafe_nodate}, in which we evaluate Lagrangian methods across eight benchmark tasks from the Safety Gymnasium task suite \citep{ji_safety-gymnasium_2024}. We evaluate on four \textit{safe navigation} tasks, level-1 Circle, Goal, Button, and Push. The environments use the Point agent, a robot constrained to a 2D plane with two actuators: a rotational action that controls the angular velocity of the agent around the $z$-axis, and an action that applies force to the agent for forward/backward movement along the agent's facing direction. The objective is to accomplish a task-specific goal that involves interacting with objects and navigating in a certain direction, while avoiding contact with certain areas or objects. 

We furthermore evaluate on four \textit{safe velocity} tasks, which contain locomotion agents Hopper, Walker2d, HalfCheetah and Ant, each of which contains actuators at each of their joints that controls the torque around their rotational axes. The objectives of these tasks involves moving forward as quickly as possible while adhering to a velocity constraint. 

\begin{figure}[H]
  \centering

  \hspace{-3em}
  \begin{subfigure}{0.24\textwidth}
    \centering
    \includegraphics[width=1.2\linewidth]{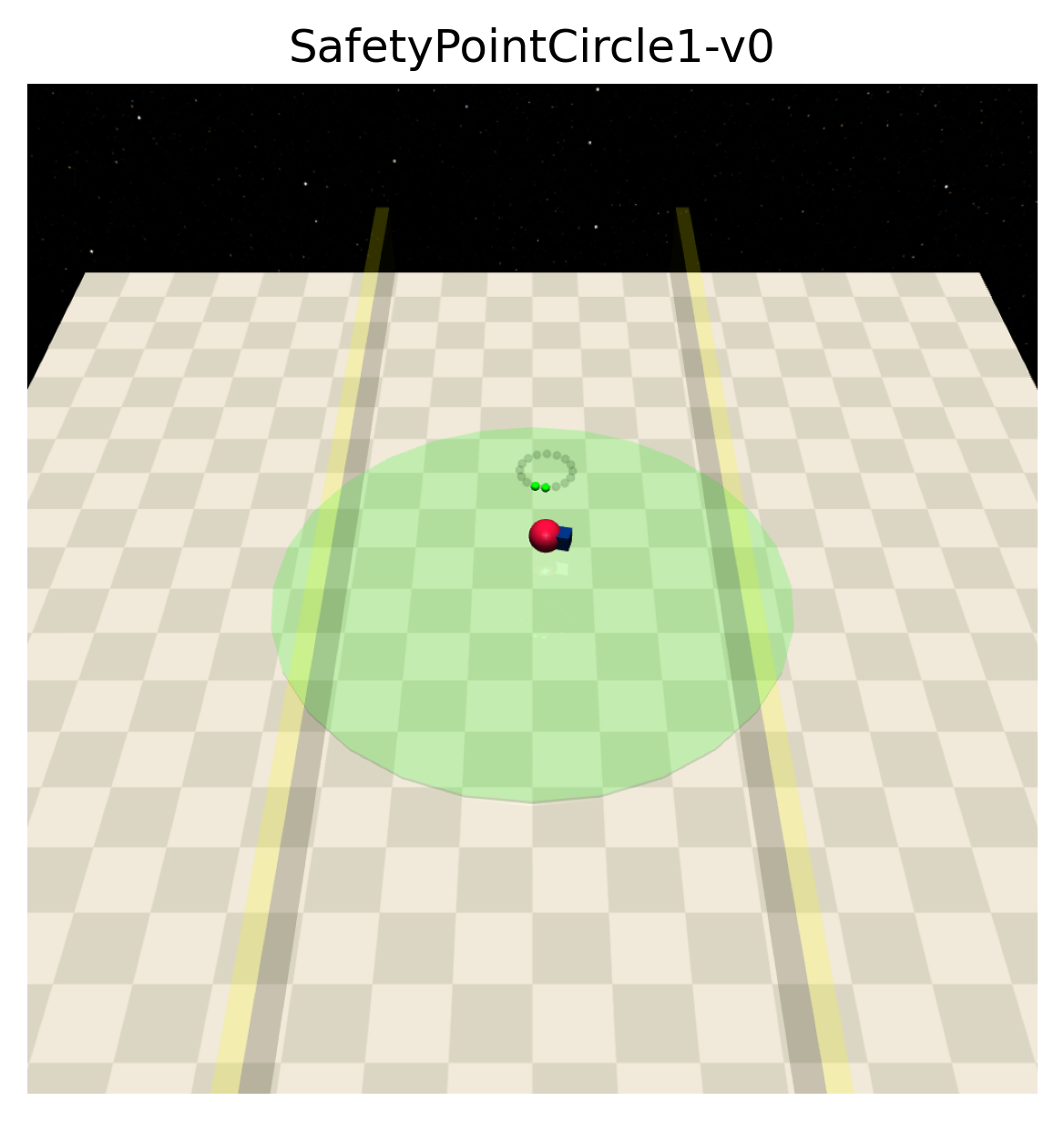}
    \caption{}
    \label{fig:task_circle}
  \end{subfigure}\hfill
  \begin{subfigure}{0.24\textwidth}
    \centering
    \includegraphics[width=1.2\linewidth]{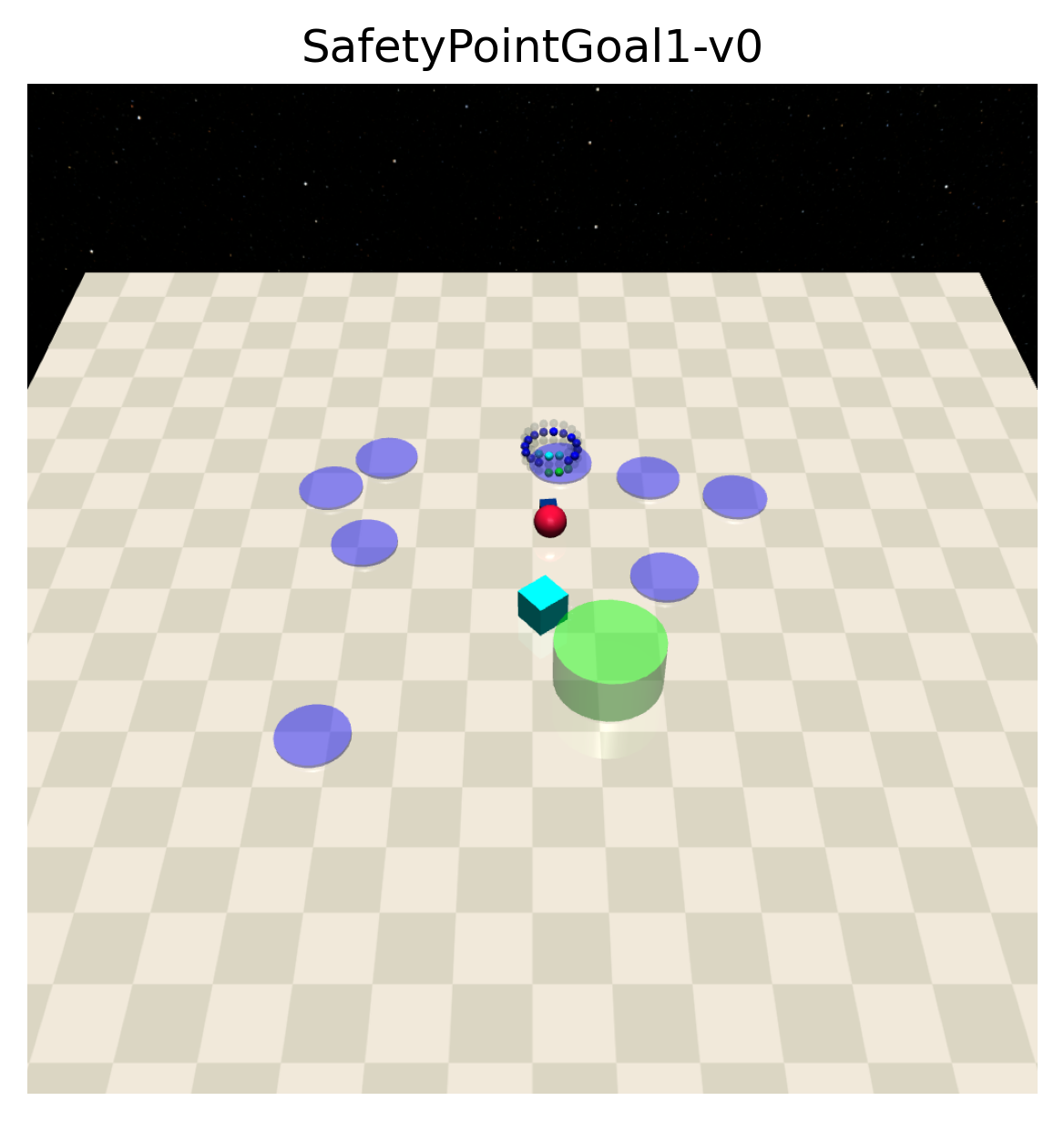}
    \caption{}
    \label{fig:task_goal}
  \end{subfigure}\hfill
  \begin{subfigure}{0.24\textwidth}
    \centering
    \includegraphics[width=1.2\linewidth]{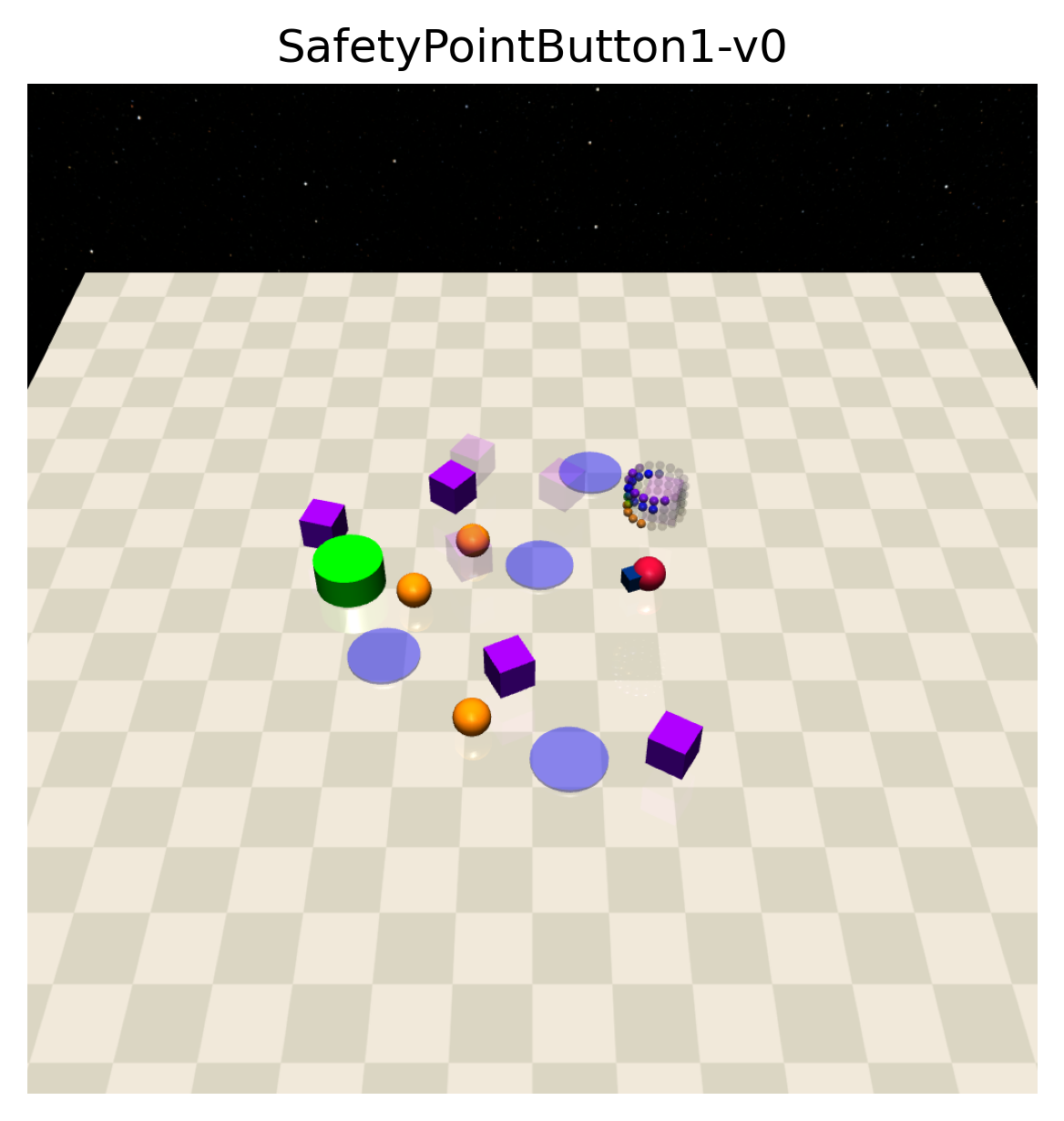}
    \caption{}
    \label{fig:task_button}
  \end{subfigure}\hfill
  \begin{subfigure}{0.24\textwidth}
    \centering
    \includegraphics[width=1.2\linewidth]{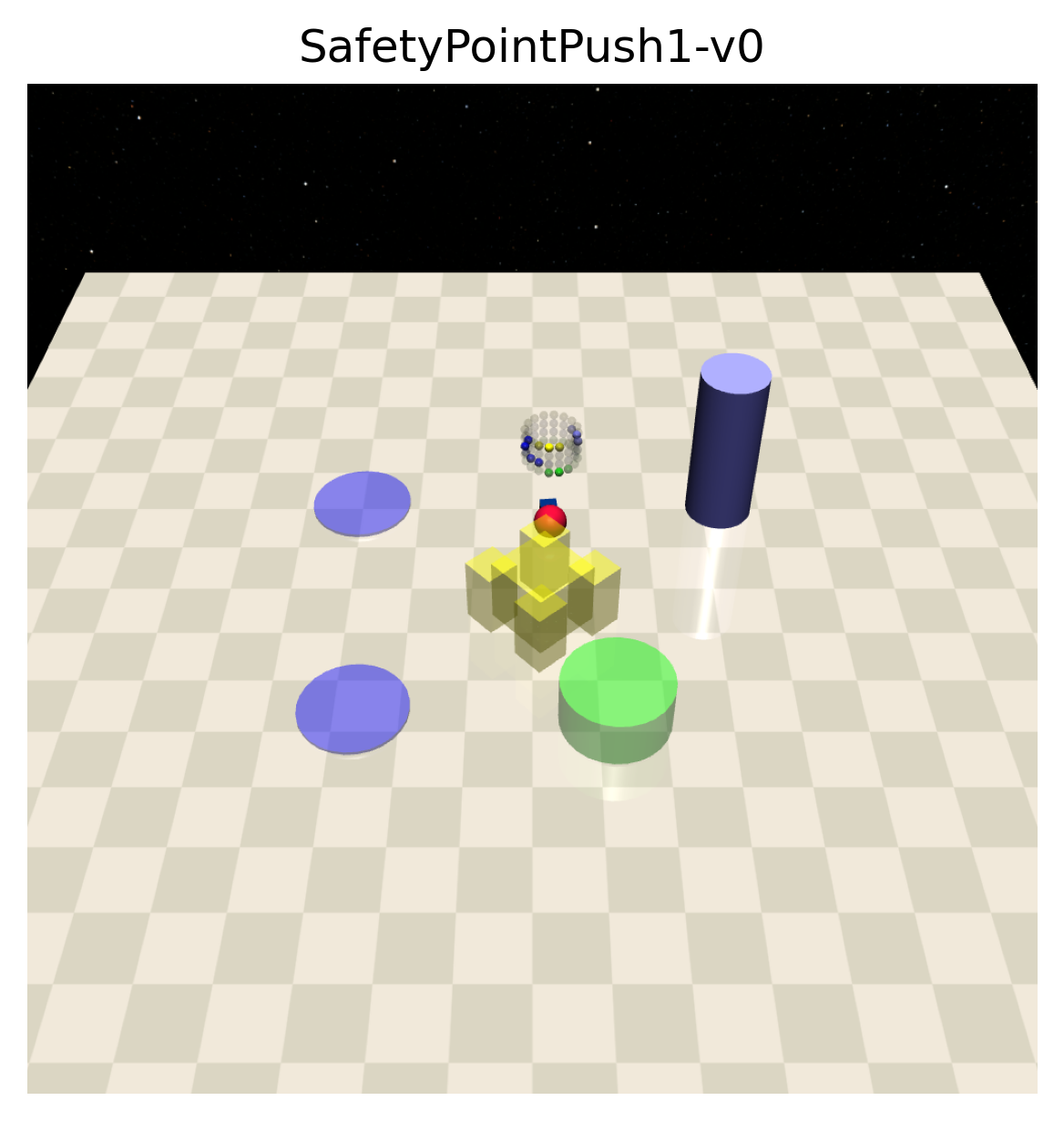}
    \caption{}
    \label{fig:task_push}
  \end{subfigure}
  \vspace{0.5em}
  \hspace{-3em}
  \begin{subfigure}{0.24\textwidth}
    \centering
    \includegraphics[width=1.2\linewidth]{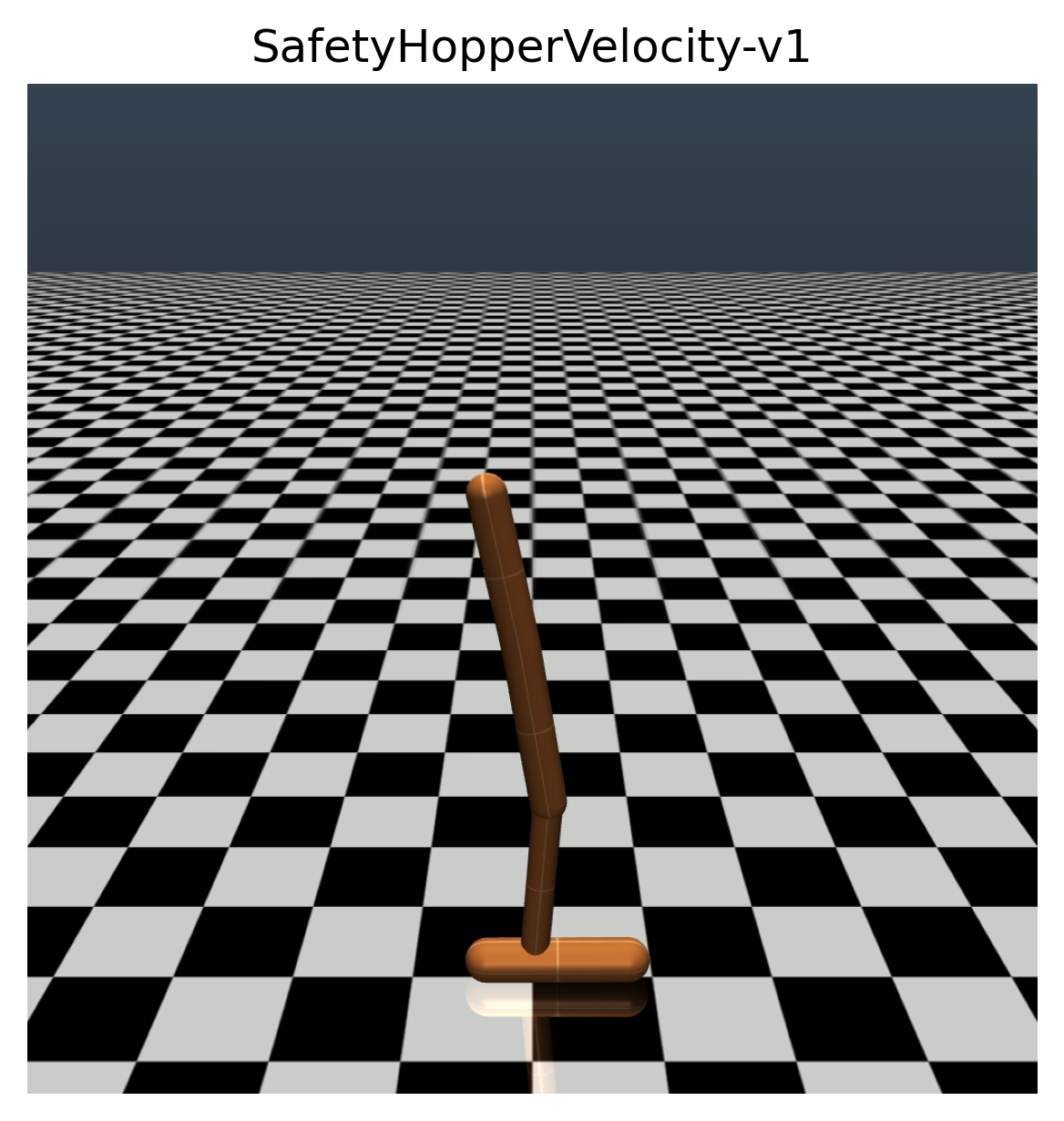}
    \caption{}
    \label{fig:task_hopper}
  \end{subfigure}\hfill
  \begin{subfigure}{0.24\textwidth}
    \centering
    \includegraphics[width=1.2\linewidth]{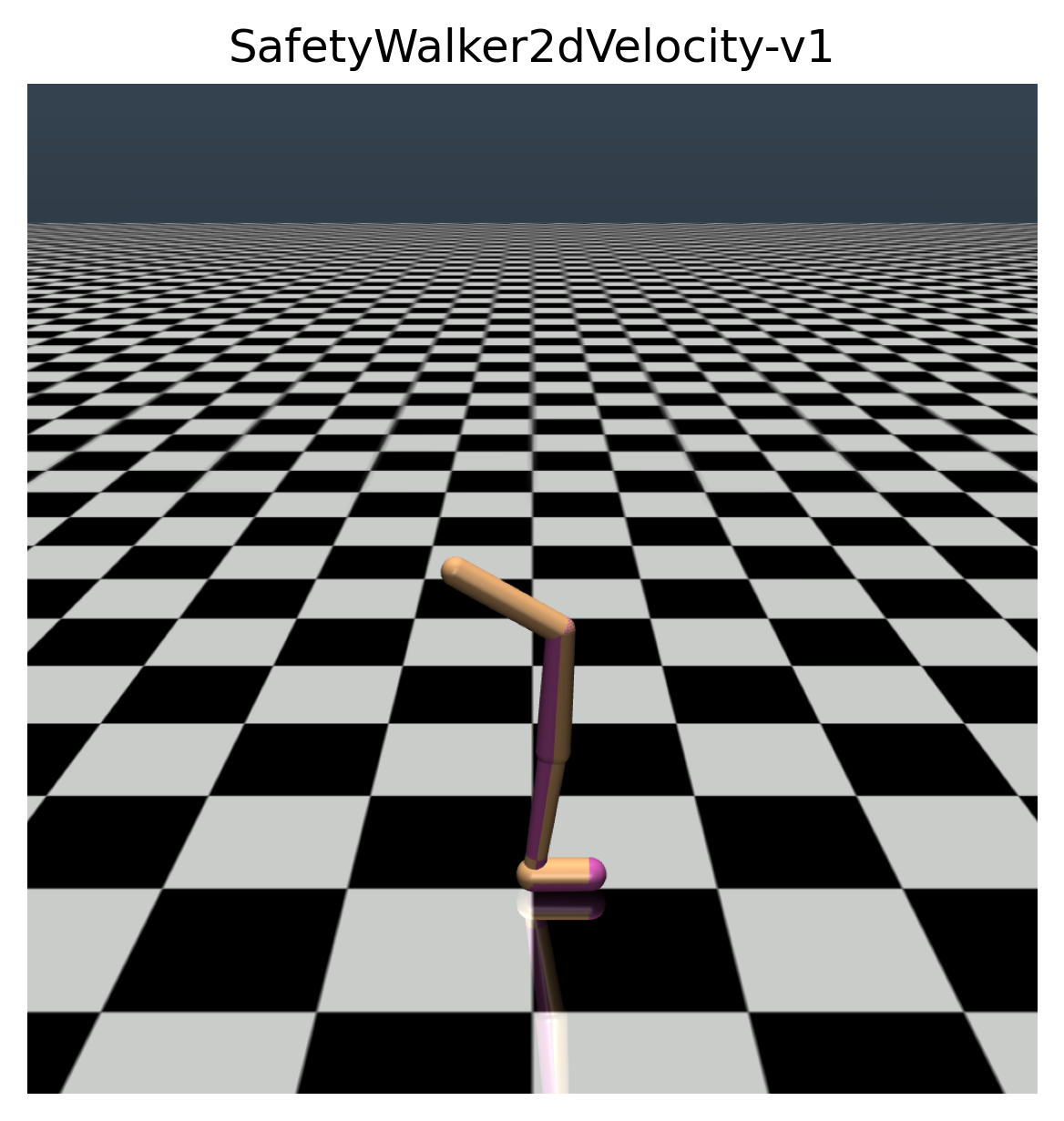}
    \caption{}
    \label{fig:task_walker2d}
  \end{subfigure}\hfill
  \begin{subfigure}{0.24\textwidth}
    \centering
    \includegraphics[width=1.2\linewidth]{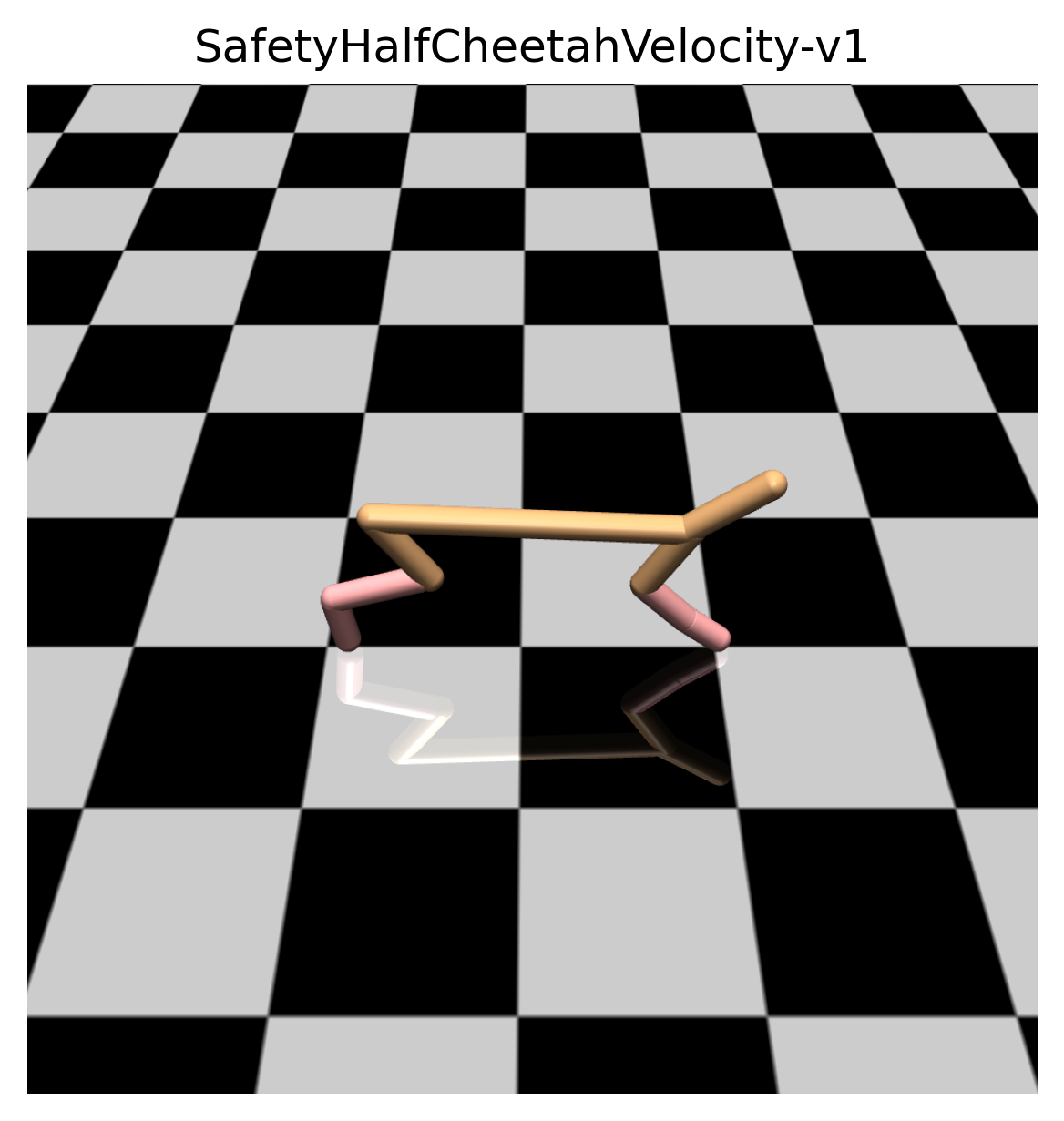}
    \caption{}
    \label{fig:task_halfcheetah}
  \end{subfigure}\hfill
  \begin{subfigure}{0.24\textwidth}
    \centering
    \includegraphics[width=1.2\linewidth]{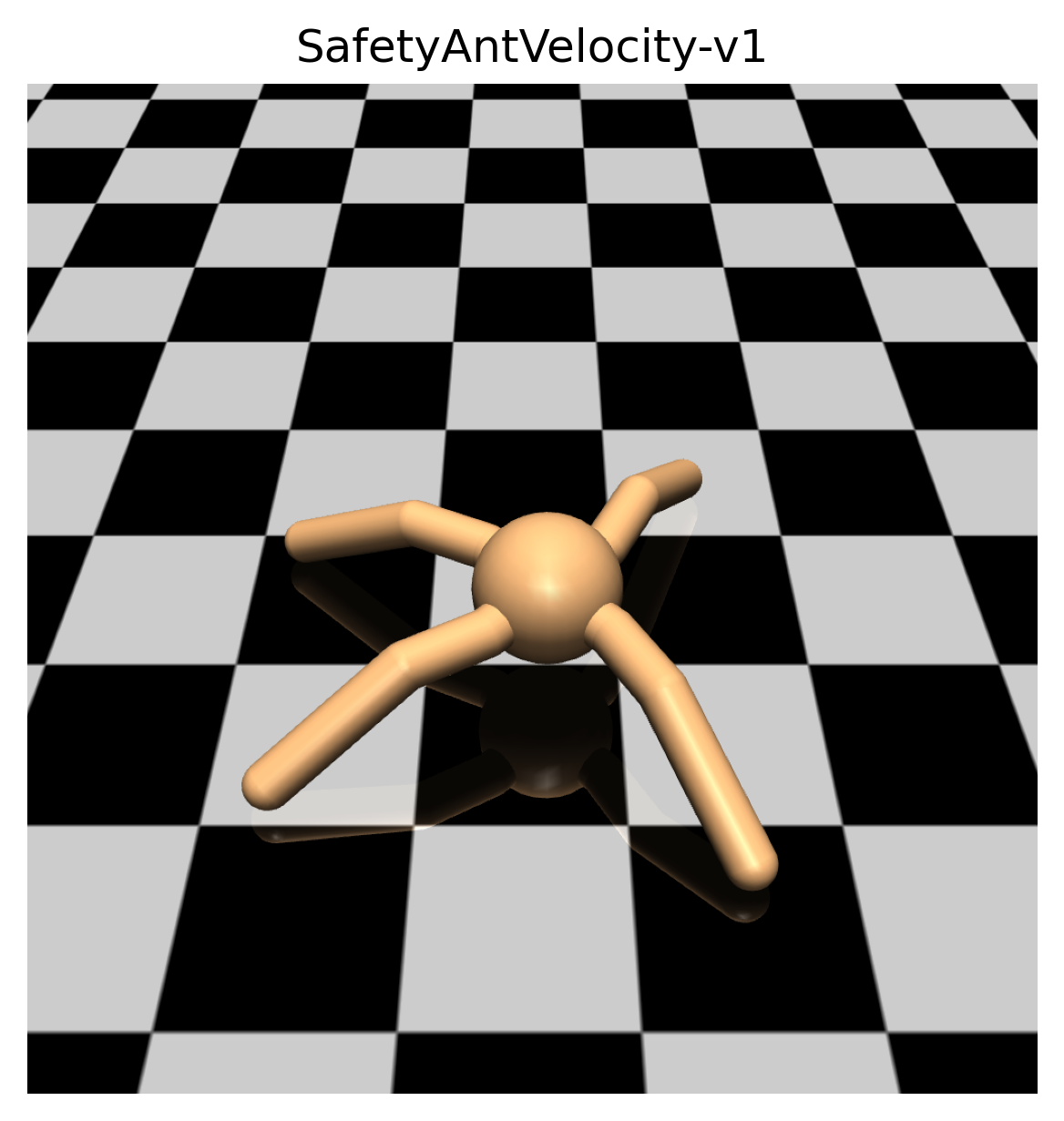}
    \caption{}
    \label{fig:task_ant}
  \end{subfigure}

  \caption{Evaluated tasks. (\ref{fig:task_circle})-(\ref{fig:task_push}) Safe navigation tasks. (\ref{fig:task_hopper})-(\ref{fig:task_ant}) Safe velocity tasks. All tasks are shown with increasing complexity from left to right.}
  \label{fig:tasks}
\end{figure}

For both the safe navigation and safe velocity tasks, we carefully picked the two subsets each consisting of four tasks based on difference in objective complexity and constraint complexity. Fig. \ref{fig:task_circle}-\ref{fig:task_push} and \ref{fig:task_hopper}-\ref{fig:task_ant} show a snapshot of each task with increasing complexity from left to right, for safe navigation and safe velocity, respectively.

\section{Additional results}\label{sec:supp: additional results}

\begin{figure}[H]
  \centering

  \hspace{-3em}
  \begin{subfigure}{0.33\textwidth}
    \centering
    \includegraphics[width=1.1\linewidth]{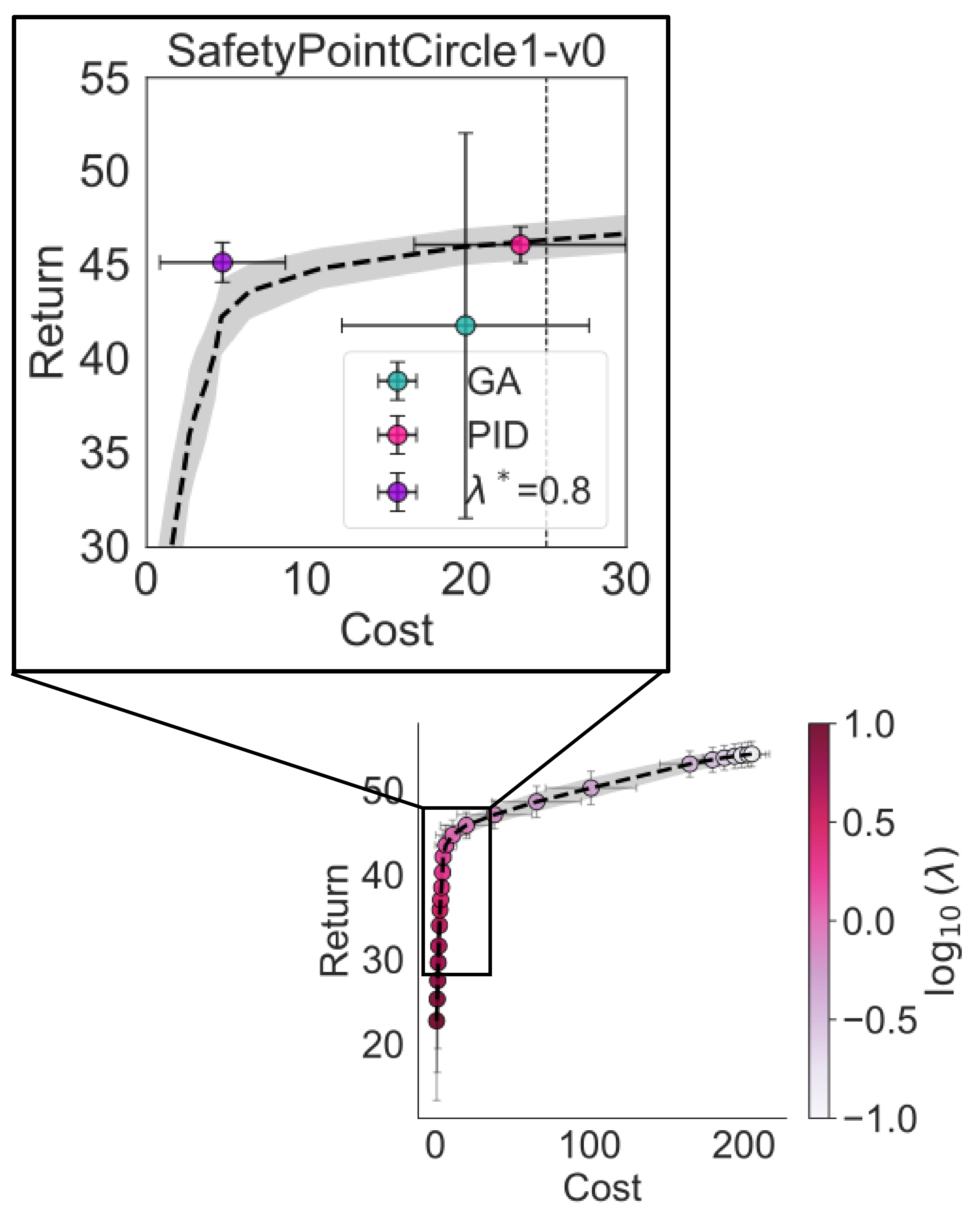}
    \caption{}
    \label{fig:pareto_rsi_circle_zoomed_combined}
  \end{subfigure}\hfill
  \begin{subfigure}{0.33\textwidth}
    \centering
    \includegraphics[width=1.1\linewidth]{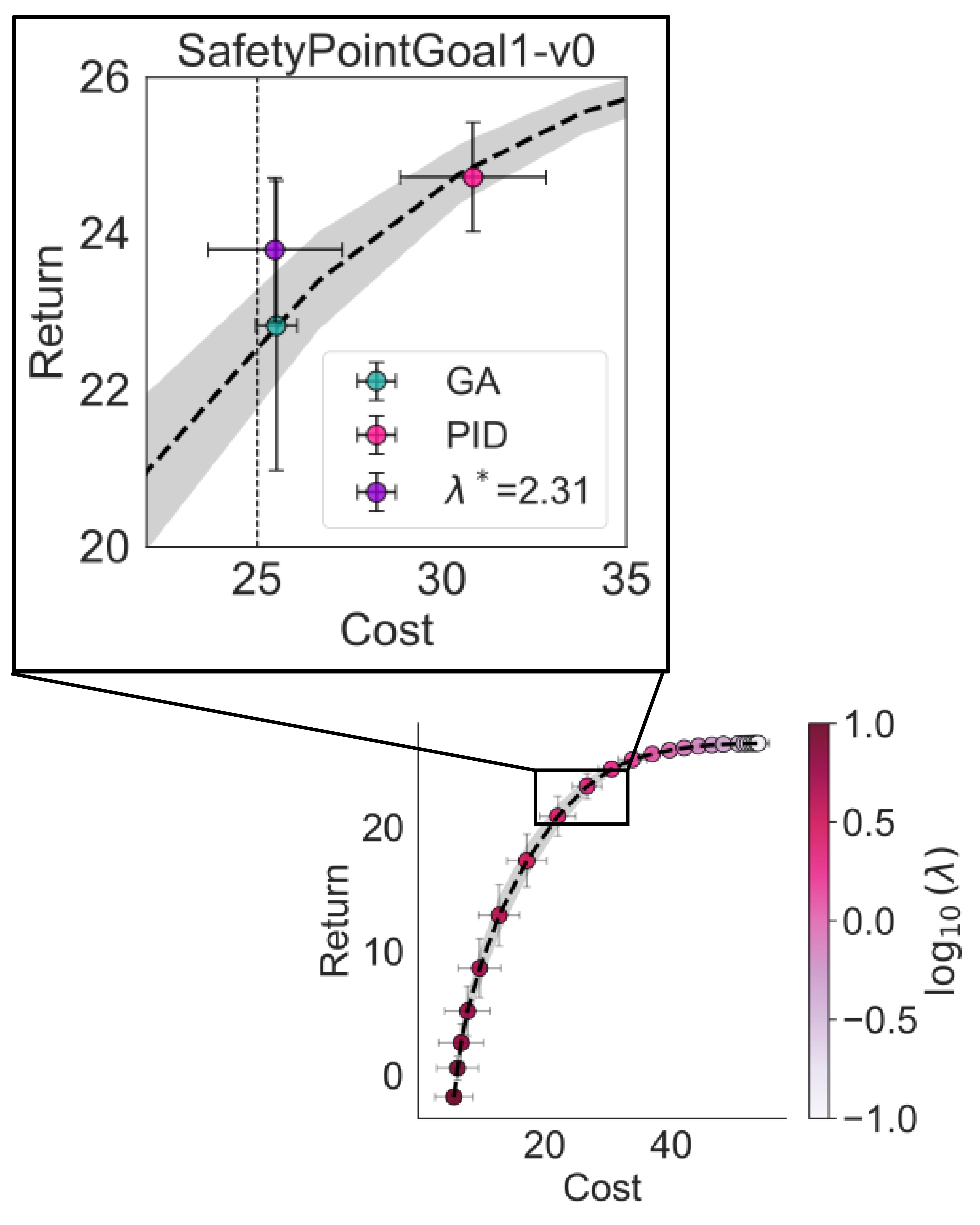}
    \caption{}
    \label{fig:pareto_rsi_goal_zoomed_combined}
  \end{subfigure}\hfill
  \begin{subfigure}{0.33\textwidth}
    \centering
    \includegraphics[width=1.1\linewidth]{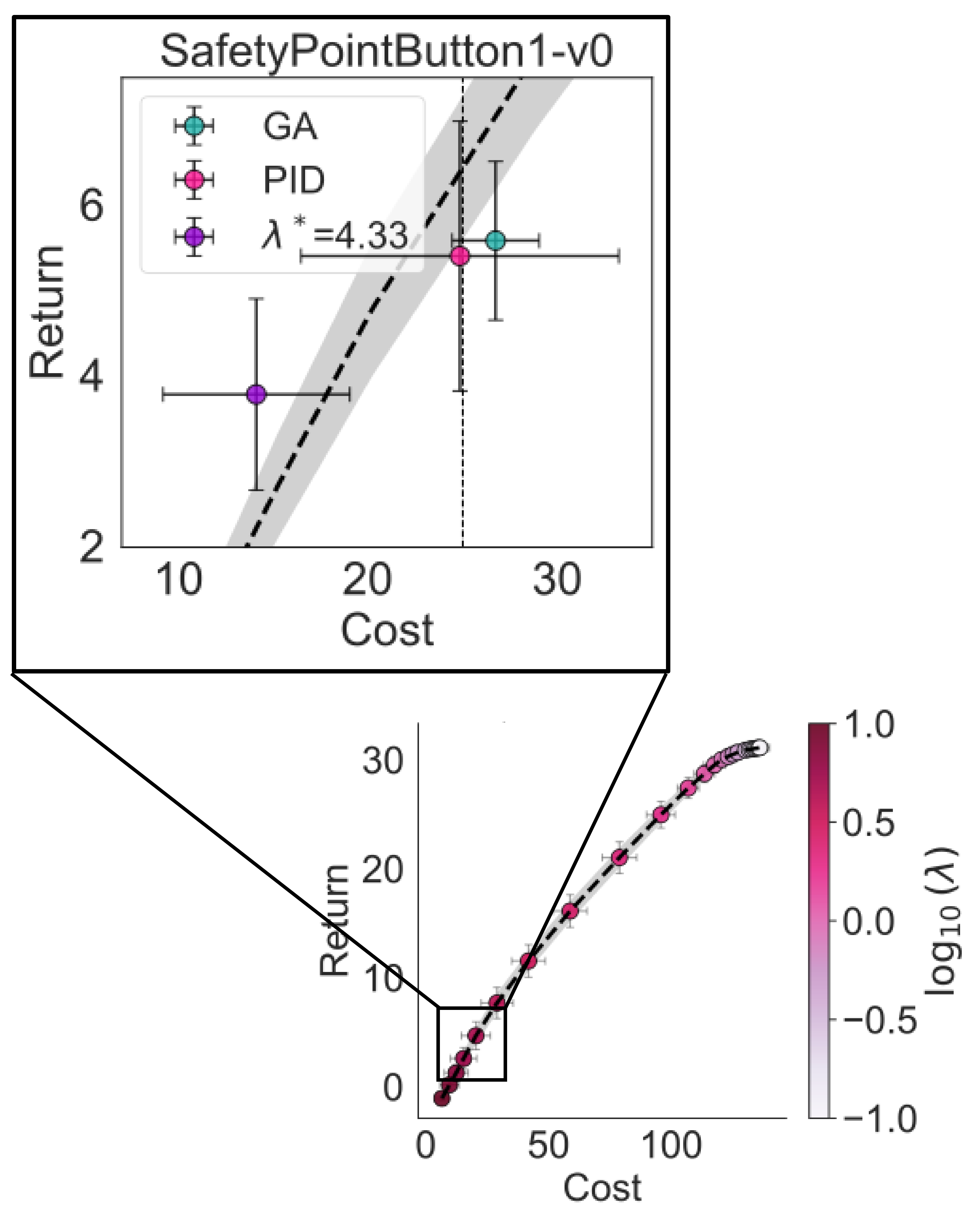}
    \caption{}
    \label{fig:pareto_rsi_button_zoomed_combined}
  \end{subfigure}
  \vspace{0.5em}
  \hspace{-3em}
  \begin{subfigure}{0.33\textwidth}
    \centering
    \includegraphics[width=1.1\linewidth]{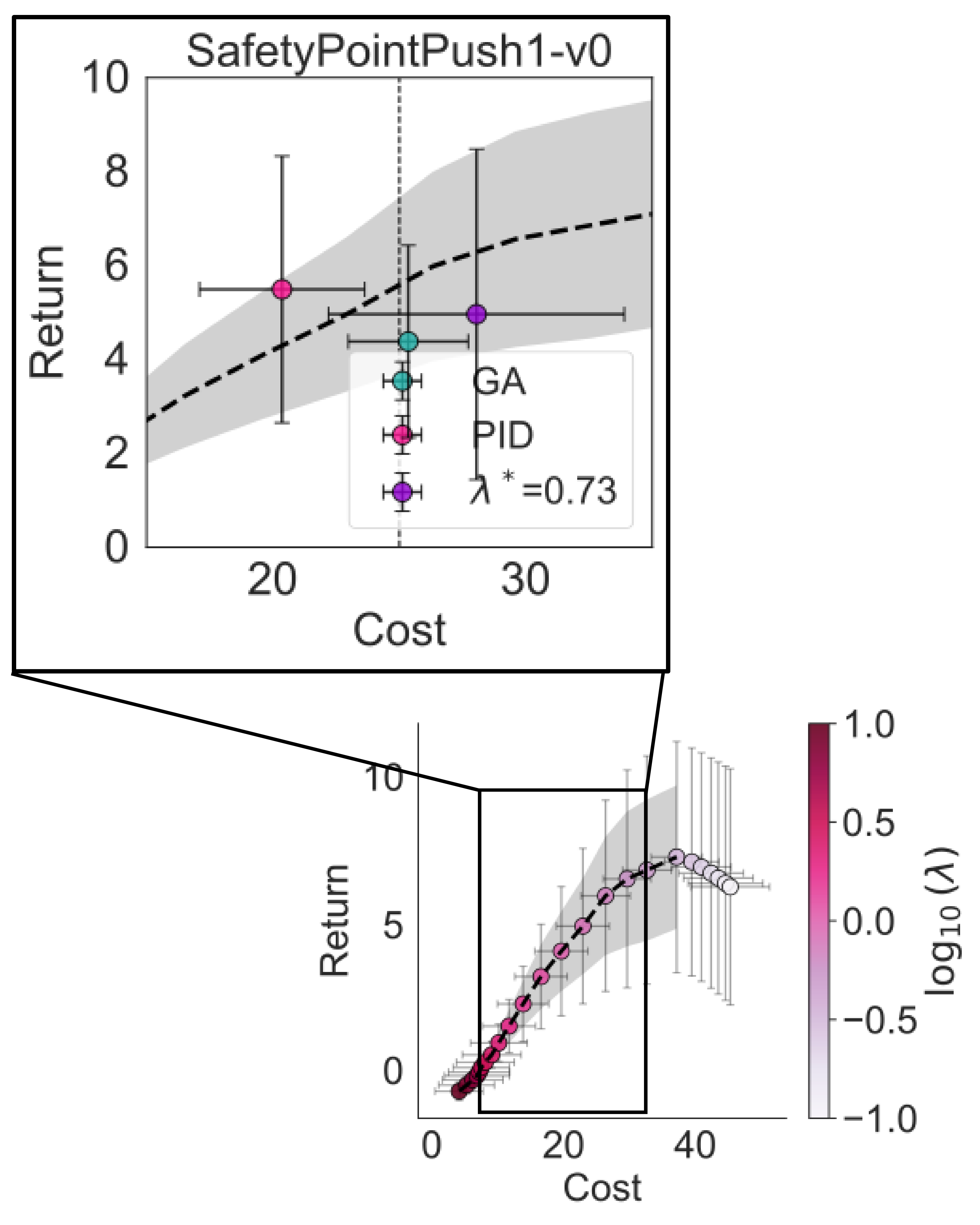}
    \caption{}
    \label{fig:pareto_rsi_push_zoomed_combined}
  \end{subfigure}\hfill
  \begin{subfigure}{0.33\textwidth}
    \centering
    \includegraphics[width=1.1\linewidth]{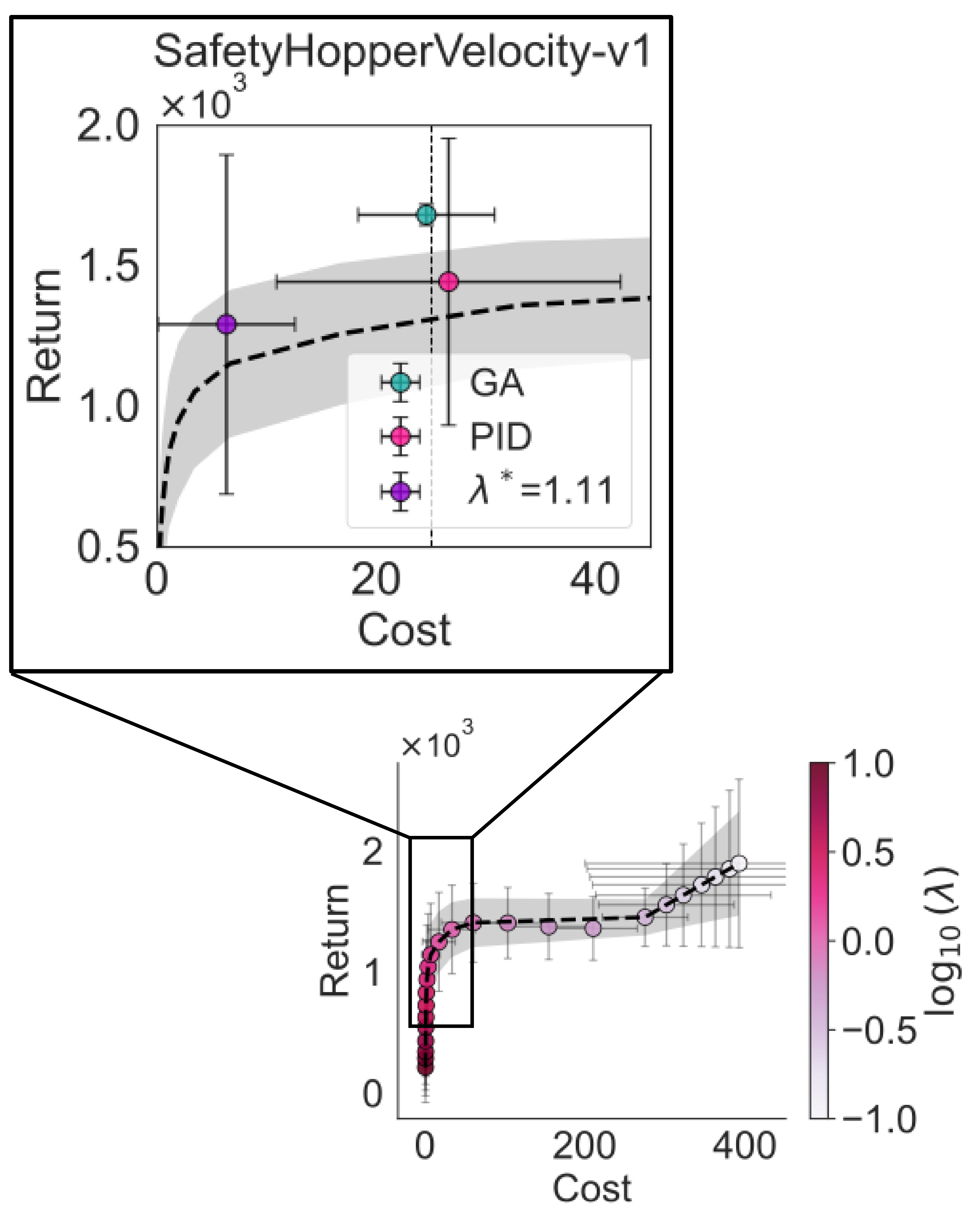}
    \caption{}
    \label{fig:pareto_rsi_hopper_zoomed_combined}
  \end{subfigure}\hfill
  \begin{subfigure}{0.33\textwidth}
    \centering
    \includegraphics[width=1.1\linewidth]{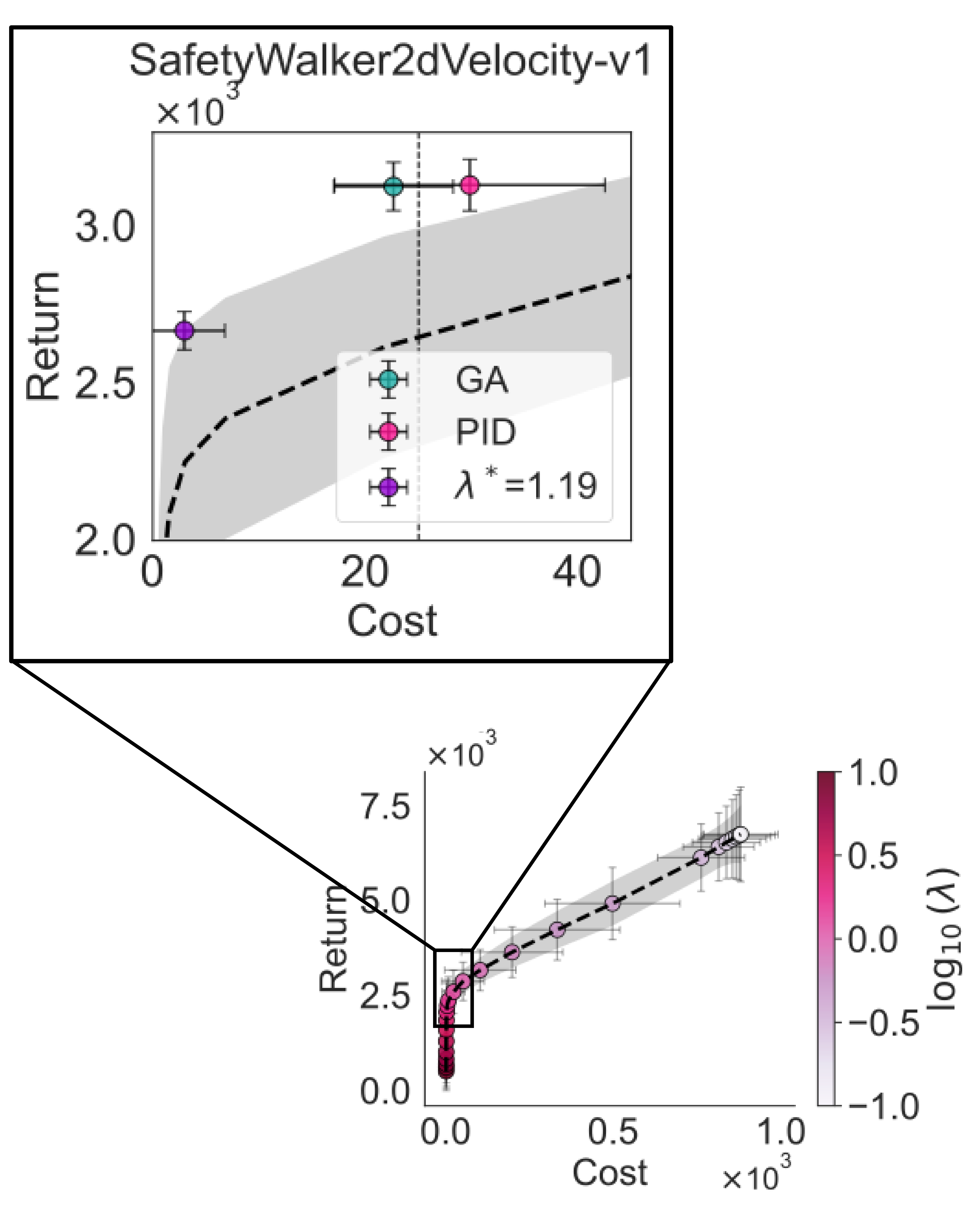}
    \caption{}
    \label{fig:pareto_rsi_walker2d_zoomed_combined}
  \end{subfigure}
  \vspace{0.5em}
  \hspace{-3em}
  \begin{subfigure}{0.33\textwidth}
    \centering
    \includegraphics[width=1.1\linewidth]{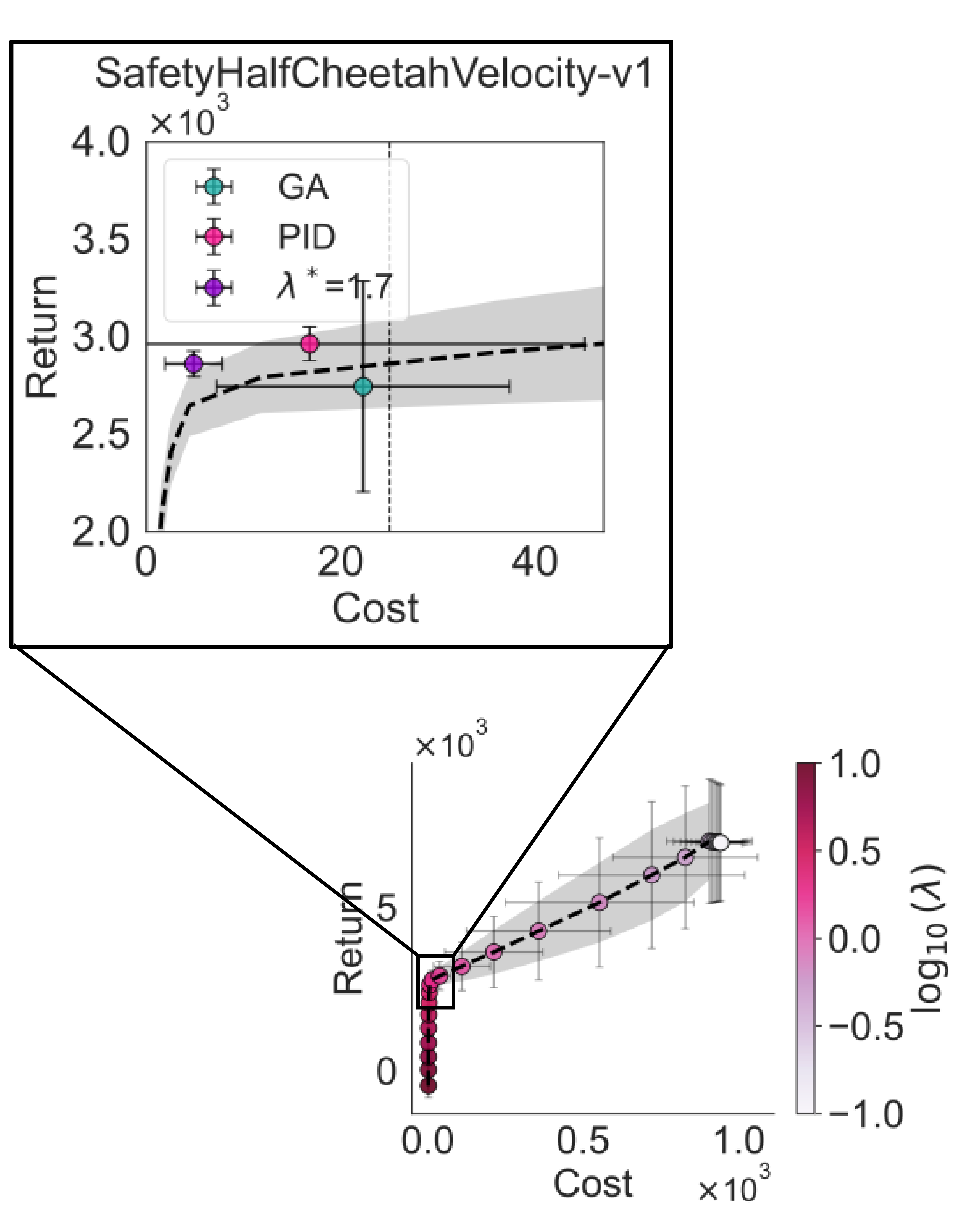}
    \caption{}
    \label{fig:pareto_rsi_halfcheetah_zoomed_combined}
  \end{subfigure}
  \begin{subfigure}{0.33\textwidth}
    \centering
    \includegraphics[width=1.1\linewidth]{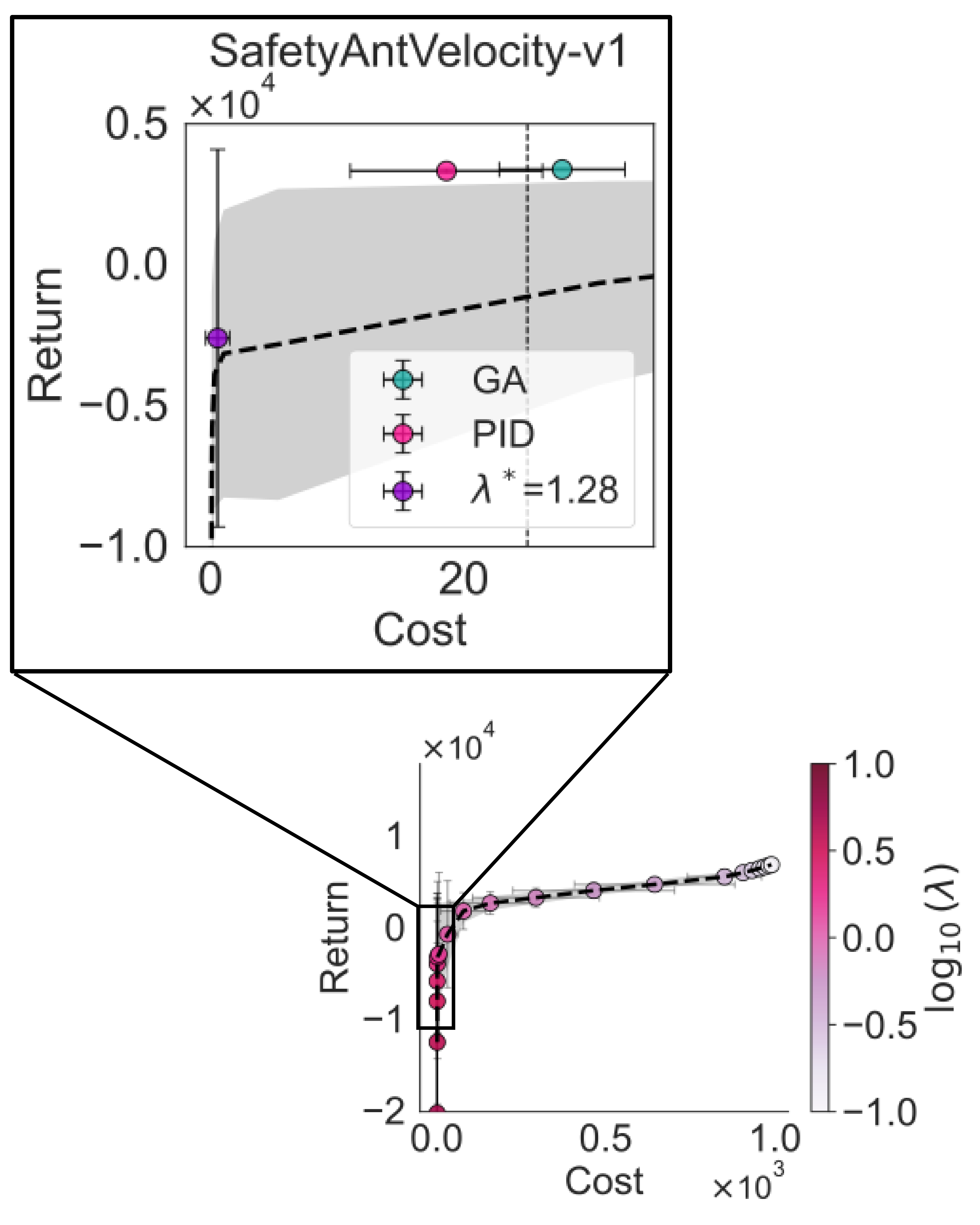}
    \caption{}
    \label{fig:pareto_rsi_ant_zoomed_combined}
  \end{subfigure}

  \caption{Smoothed empirical Pareto Frontiers of the return versus the cost as a function of $\lambda$, zoomed in at the cost limit of 25.0 (indicated by the dashed vertical lines), with the 95\% confidence intervals of the Pareto frontier in the return- and cost-dimension indicated by the grey region. The performance of a fixed $\lambda^*$ and the GA- and PID-update methods trained on the cost limit of 25.0 are shown with their standard deviations over the last 5\% of training timesteps. All results are averaged over 10 seeds.}
  \label{fig:pareto_curves_rsi_zoomed_combined}
\end{figure}

\begin{figure}[H]
  \centering

  \hspace{-3em}
  \begin{subfigure}{0.33\textwidth}
    \centering
    \includegraphics[width=1.1\linewidth]{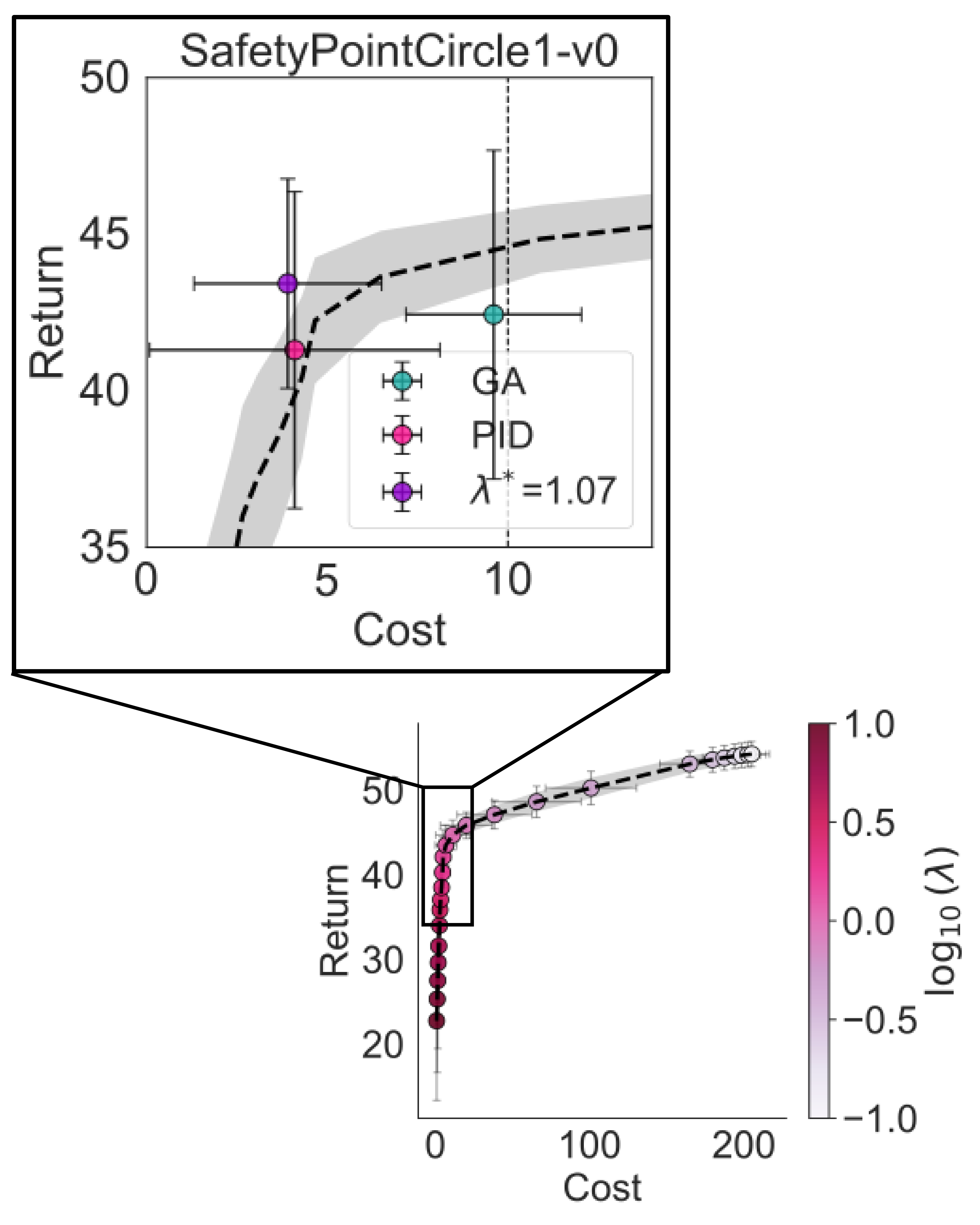}
    \caption{}
    \label{fig:pareto_rsi_circle_zoomed_combined_10}
  \end{subfigure}\hfill
  \begin{subfigure}{0.33\textwidth}
    \centering
    \includegraphics[width=1.1\linewidth]{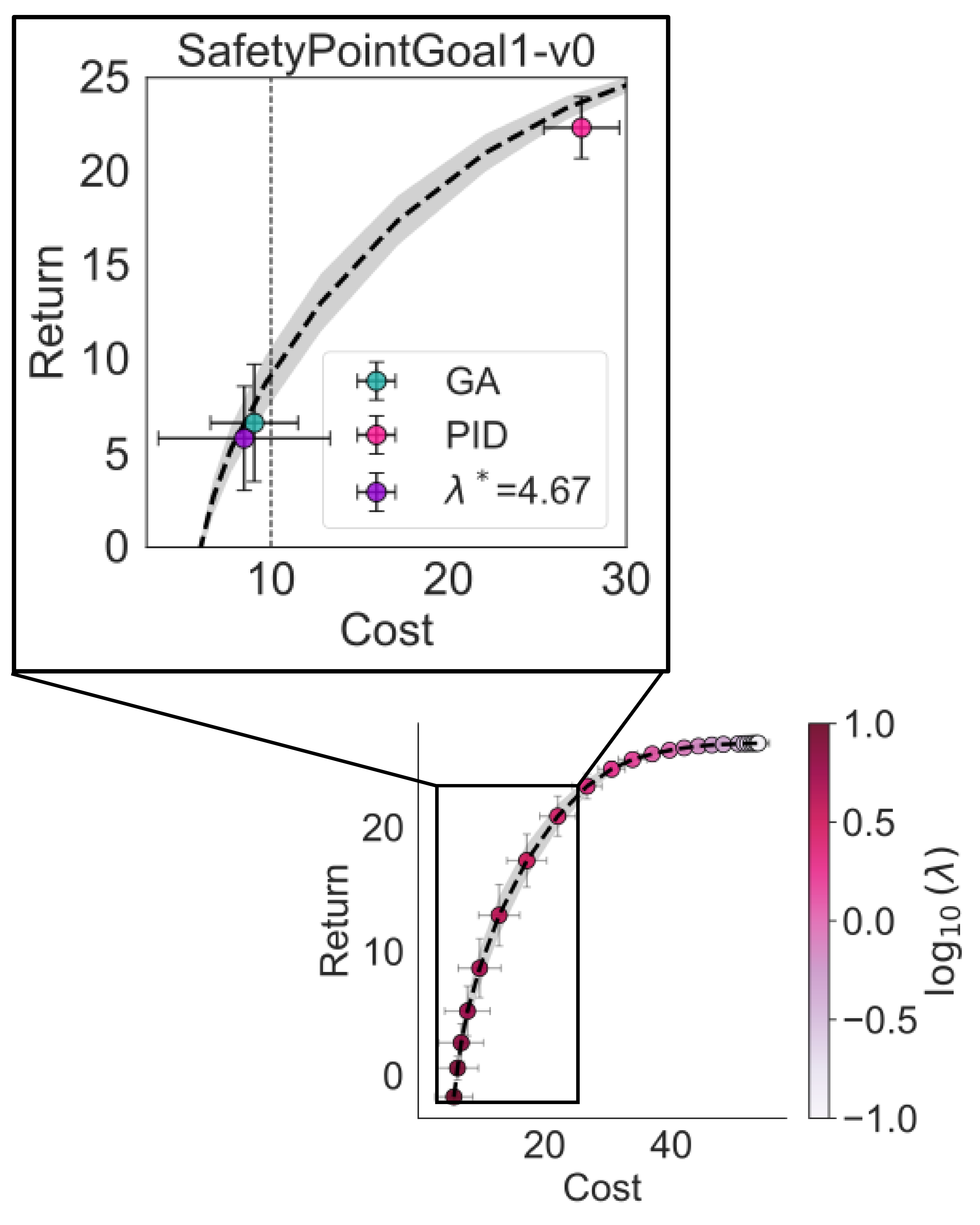}
    \caption{}
    \label{fig:pareto_rsi_goal_zoomed_combined_10}
  \end{subfigure}\hfill
  \begin{subfigure}{0.33\textwidth}
    \centering
    \includegraphics[width=1.1\linewidth]{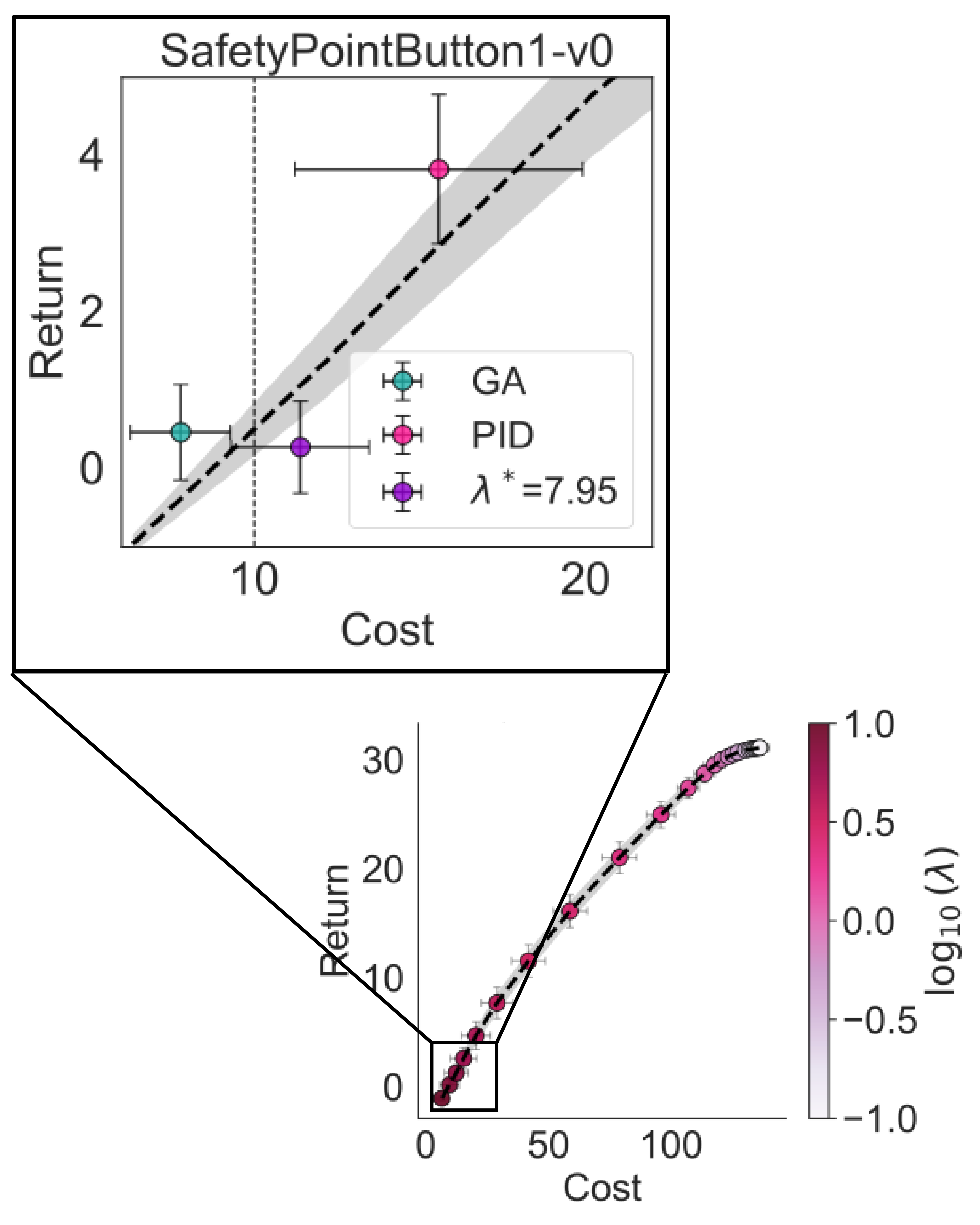}
    \caption{}
    \label{fig:pareto_rsi_button_zoomed_combined_10}
  \end{subfigure}
  \vspace{0.5em}
  \hspace{-3em}
  \begin{subfigure}{0.33\textwidth}
    \centering
    \includegraphics[width=1.1\linewidth]{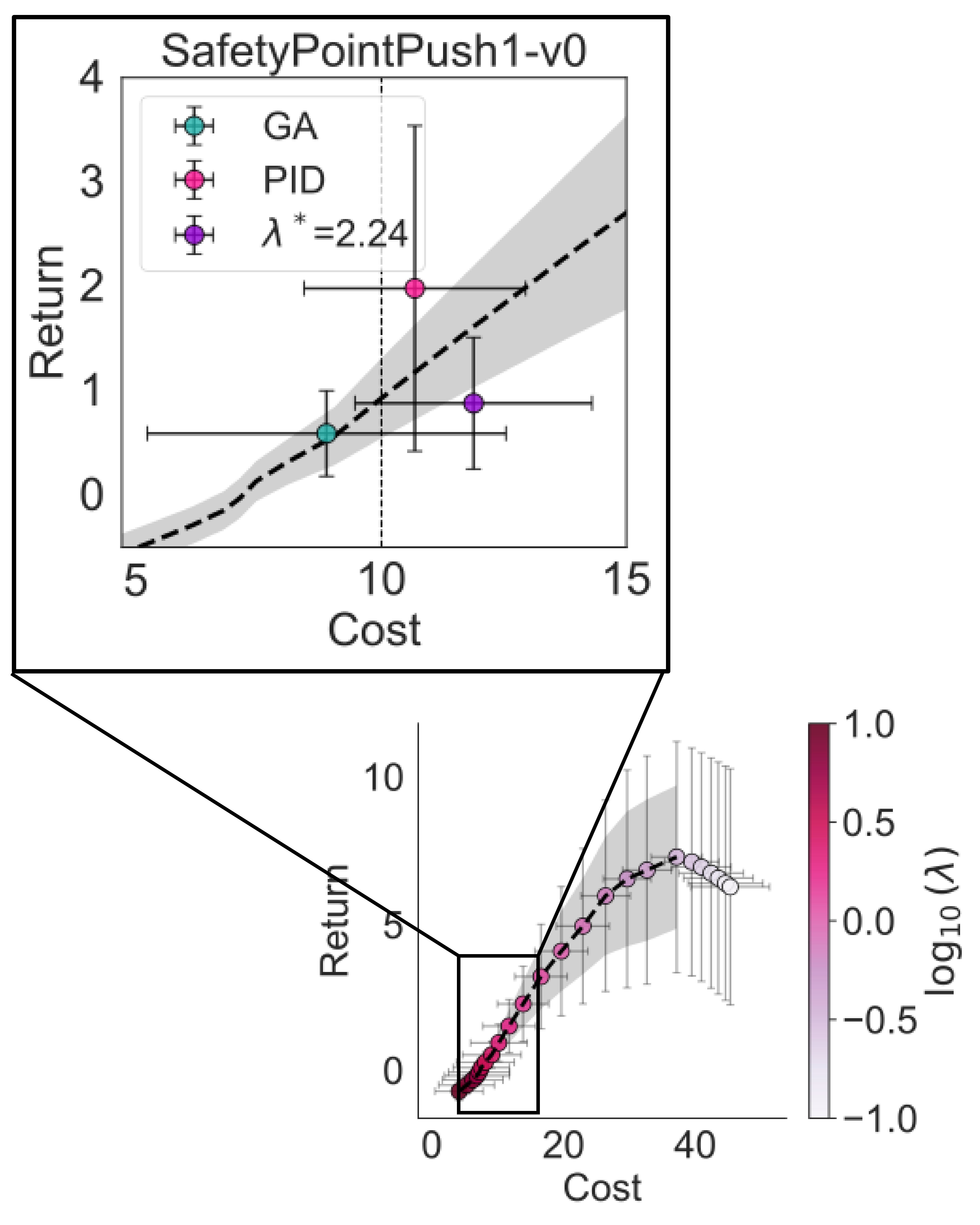}
    \caption{}
    \label{fig:pareto_rsi_push_zoomed_combined_10}
  \end{subfigure}\hfill
  \begin{subfigure}{0.33\textwidth}
    \centering
    \includegraphics[width=1.1\linewidth]{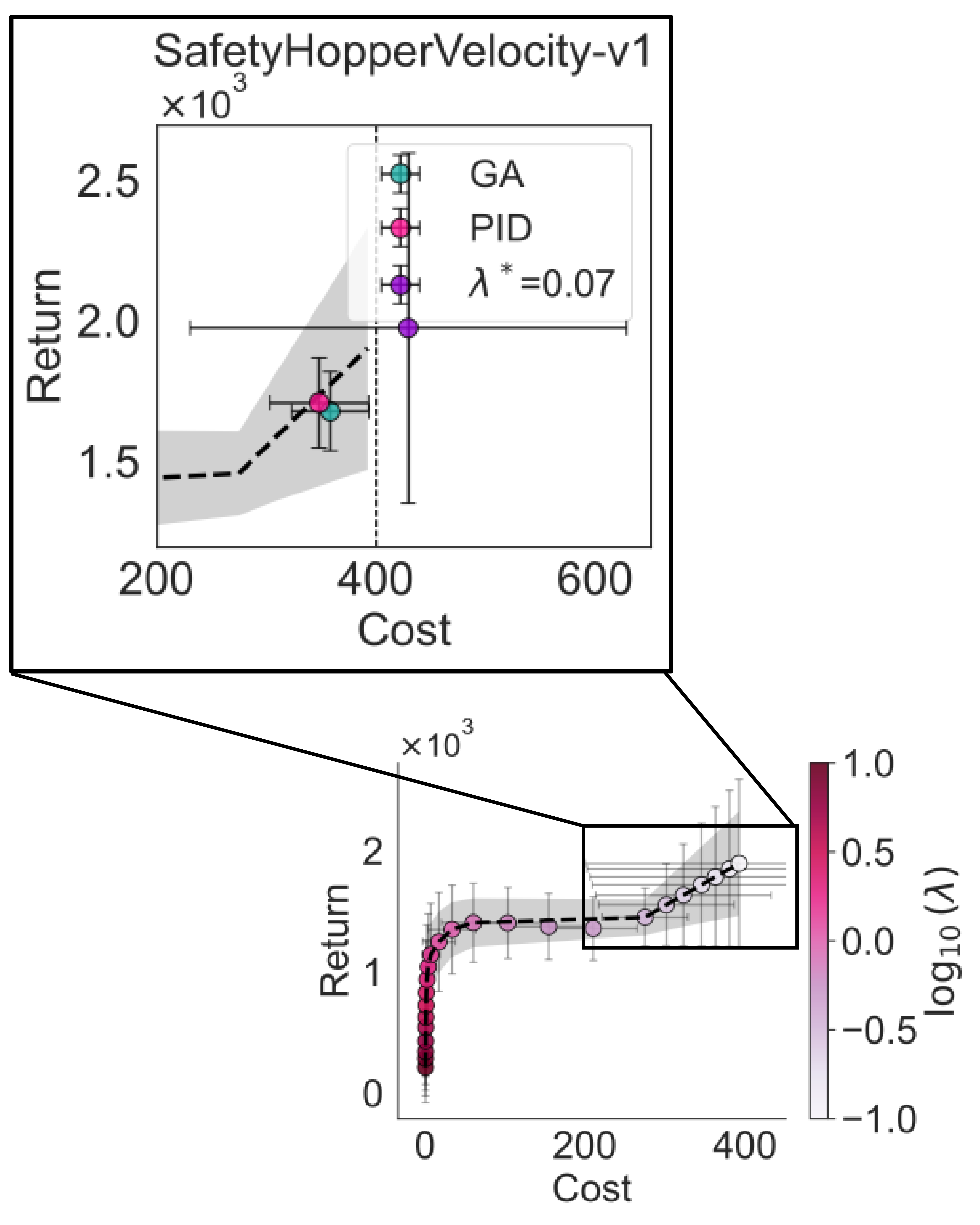}
    \caption{}
    \label{fig:pareto_rsi_hopper_zoomed_combined_400}
  \end{subfigure}\hfill
  \begin{subfigure}{0.33\textwidth}
    \centering
    \includegraphics[width=1.1\linewidth]{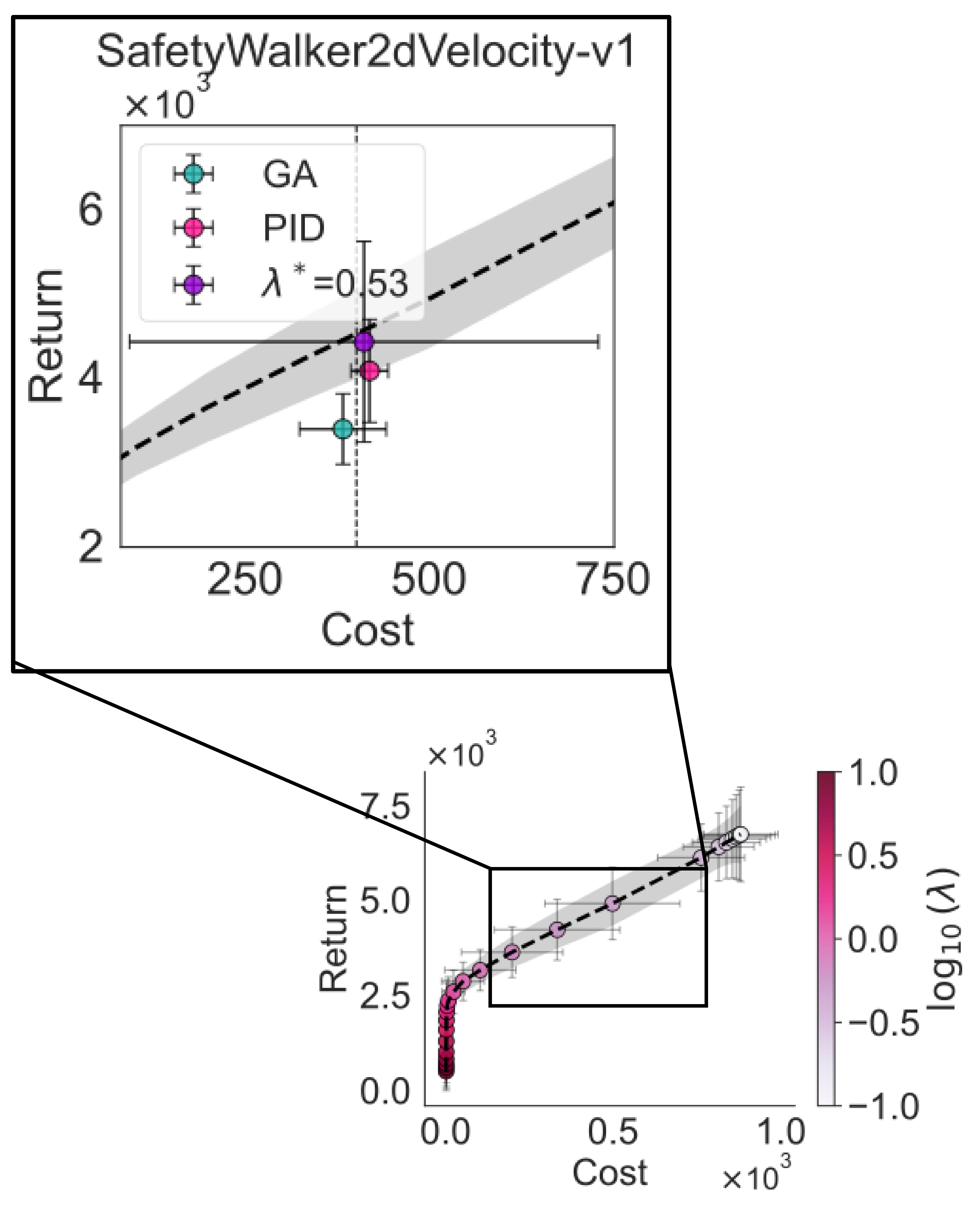}
    \caption{}
    \label{fig:pareto_rsi_walker2d_zoomed_combined_400}
  \end{subfigure}
  \vspace{0.5em}
  \hspace{-3em}
  \begin{subfigure}{0.33\textwidth}
    \centering
    \includegraphics[width=1.1\linewidth]{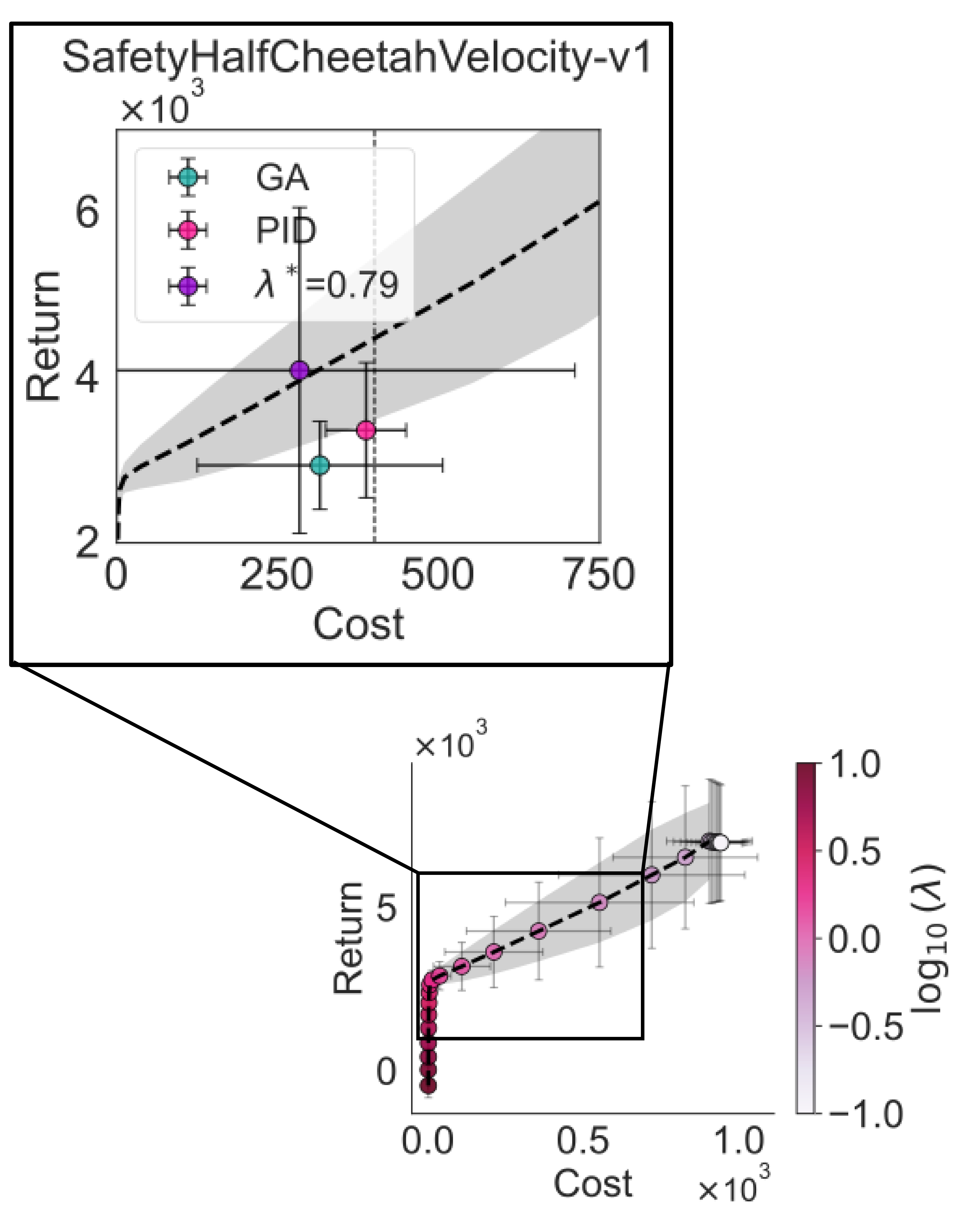}
    \caption{}
    \label{fig:pareto_rsi_halfcheetah_zoomed_combined_400}
  \end{subfigure}
  \begin{subfigure}{0.33\textwidth}
    \centering
    \includegraphics[width=1.1\linewidth]{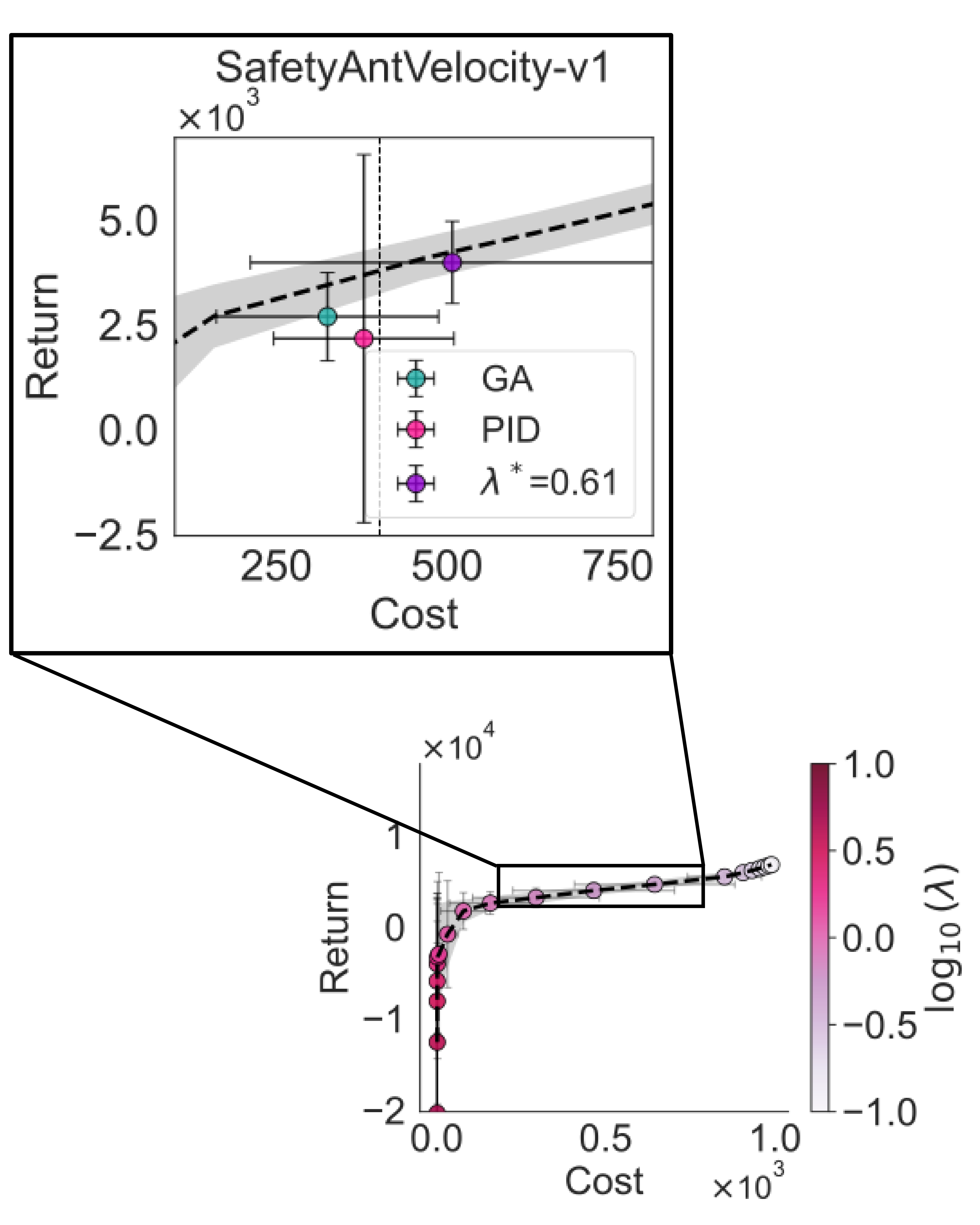}
    \caption{}
    \label{fig:pareto_rsi_ant_zoomed_combined_400}
  \end{subfigure}

  \caption{Smoothed empirical Pareto Frontiers of the return versus the cost as a function of $\lambda$, zoomed in at the cost limit of 10.0 for the navigation tasks, and a cost limit of 400.0 for the velocity tasks (indicated by the dashed vertical lines). The 95\% confidence intervals of the Pareto frontier in the return- and cost-dimension is indicated by the grey region. The performance of a fixed $\lambda^*$ and the GA- and PID-update methods trained on their respective cost limits are shown with their standard deviations over the last 5\% of training timesteps. All results are averaged over 10 seeds.}
  \label{fig:pareto_curves_rsi_zoomed_combined_10_400}
\end{figure}

The results from Table \ref{tab:comparison_table} in the main paper are projected onto the empirical Pareto frontiers and are shown in Figures \ref{fig:pareto_curves_rsi_zoomed_combined} and \ref{fig:pareto_curves_rsi_zoomed_combined_10_400}. Small deviations of the fixed $\lambda^*$ from the frontier arise due to smoothing effects. The results differ substantially across tasks and across the evaluated cost limits. We remark that the results shown in Figure \ref{fig:pareto_rsi_button_zoomed_combined} exhibit anomalous behaviour, with the datapoint corresponding to the fixed $\lambda^*$ deviating substantially from the cost limit of 25.0. Based on the shape of the Pareto frontier, the optimal return–cost trade-off for this limit is expected to lie around a cost of 25, as the frontier is steep in this region. However, the datapoint is projected at a suboptimal trade-off location. We attribute this deviation to smoothing effects in the empirical Pareto frontier estimation, which appear to be more pronounced for this task than for the others.

\subsection{Training curves} \label{sec: supp: training_curves}

\begin{figure}[H]
  \centering

  \begin{subfigure}{0.33\textwidth}
    \centering
    \includegraphics[width=1\linewidth]{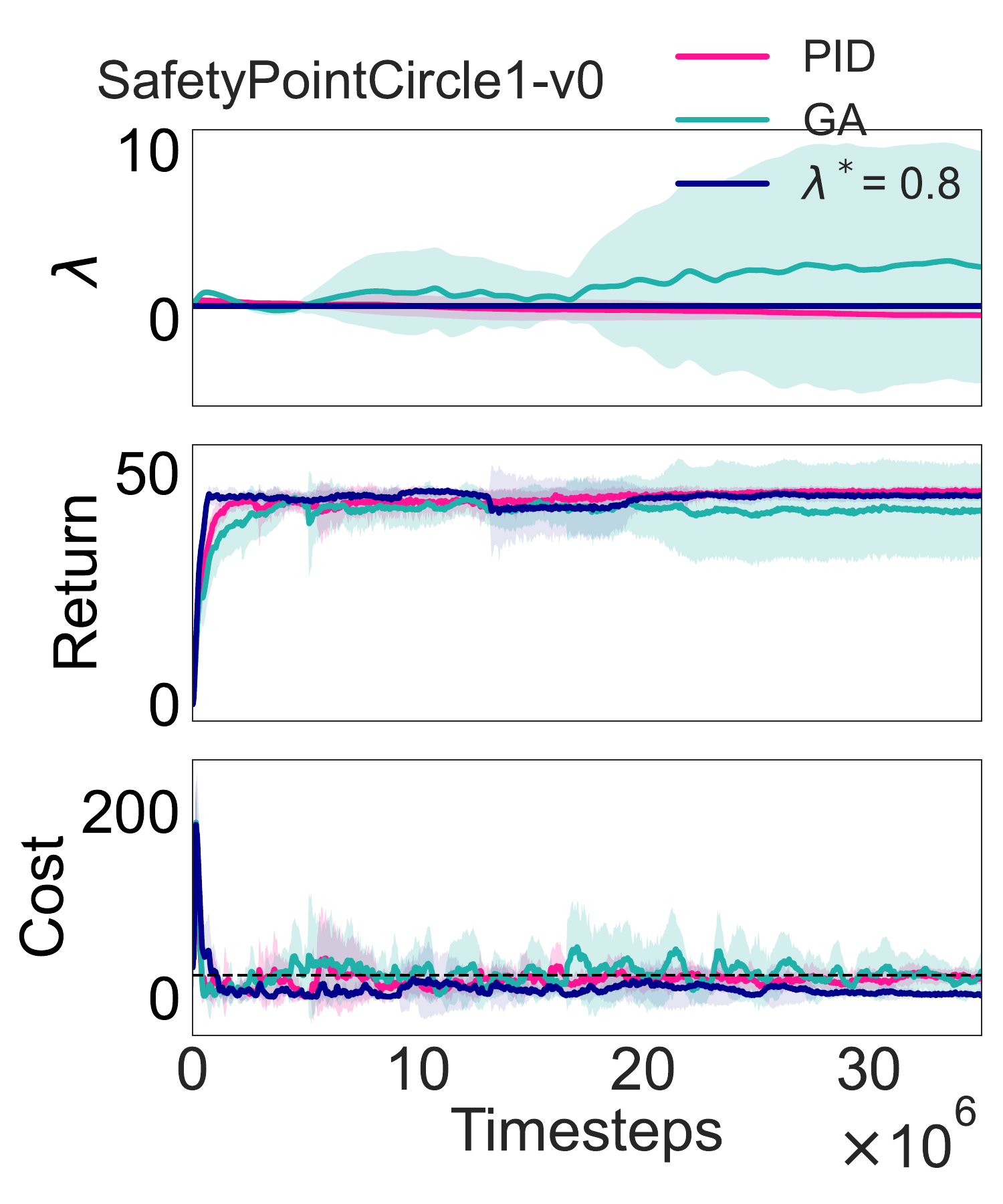}
    \caption{}
    \label{fig:comparison_circle}
  \end{subfigure}\hfill
  \begin{subfigure}{0.33\textwidth}
    \centering
    \includegraphics[width=1\linewidth]{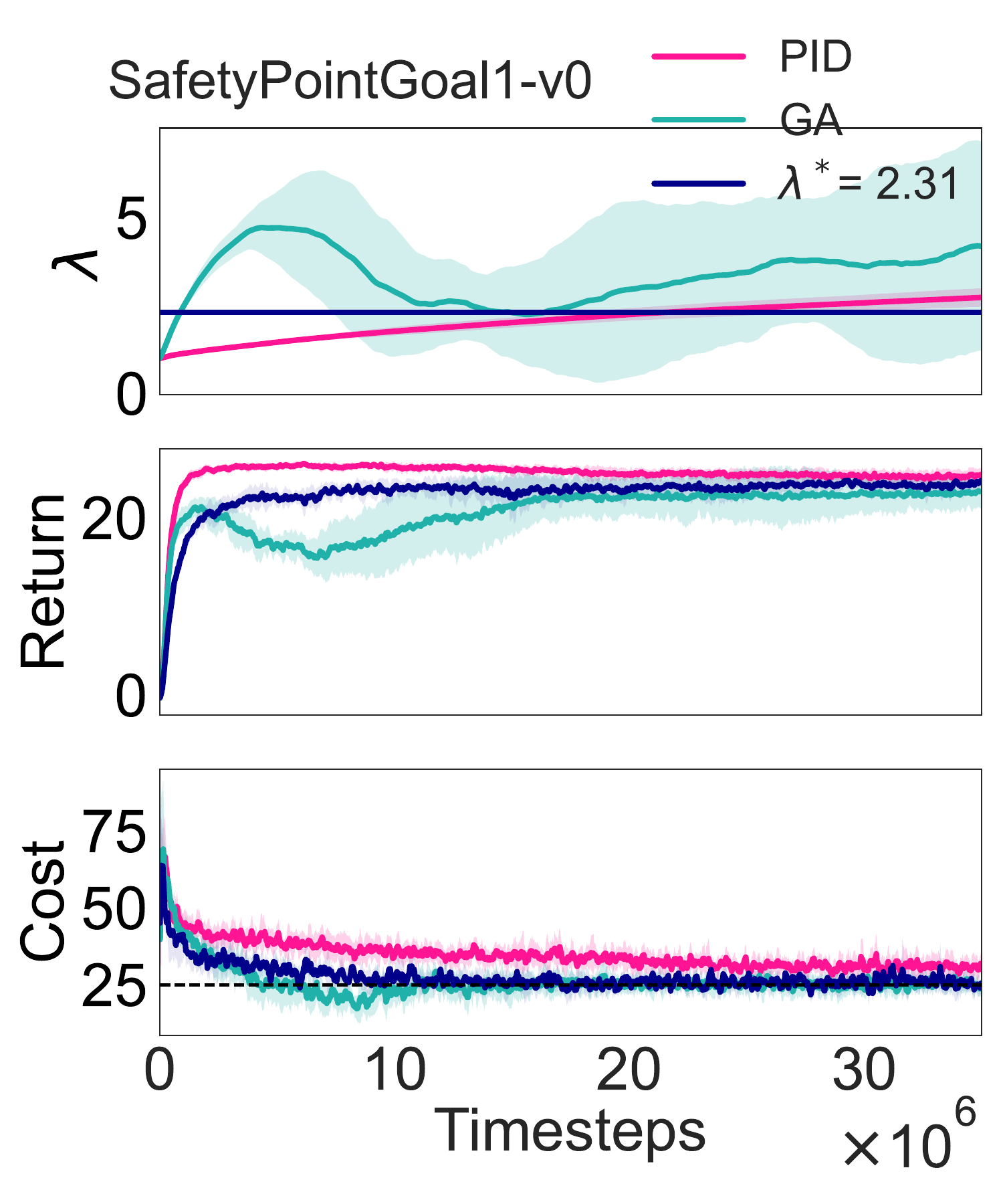}
    \caption{}
    \label{fig:comparison_goal}
  \end{subfigure}\hfill
  \begin{subfigure}{0.33\textwidth}
    \centering
    \includegraphics[width=1\linewidth]{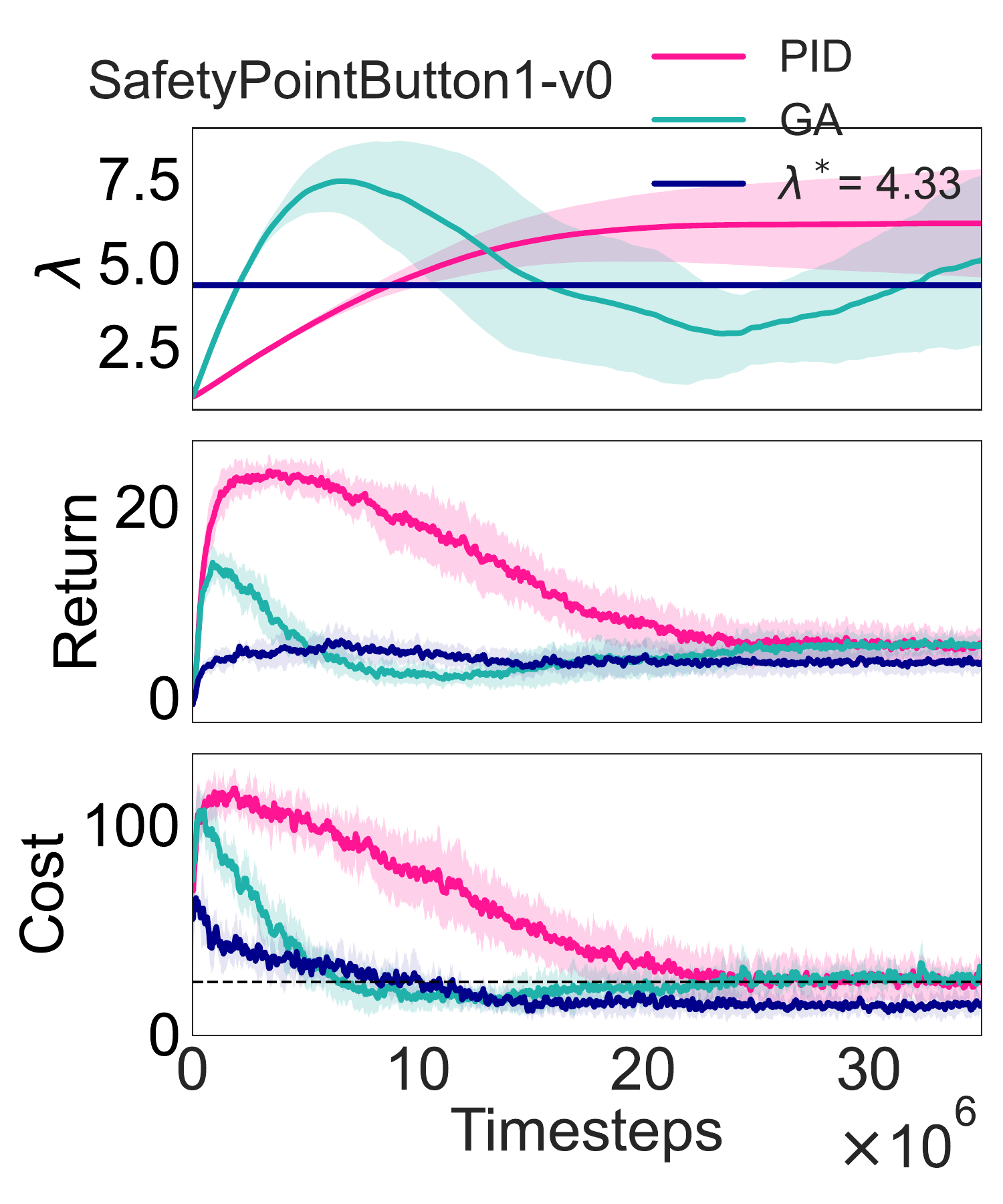}
    \caption{}
    \label{fig:comparison_button}
  \end{subfigure}
  \vspace{0.5em}
  \begin{subfigure}{0.33\textwidth}
    \centering
    \includegraphics[width=1\linewidth]{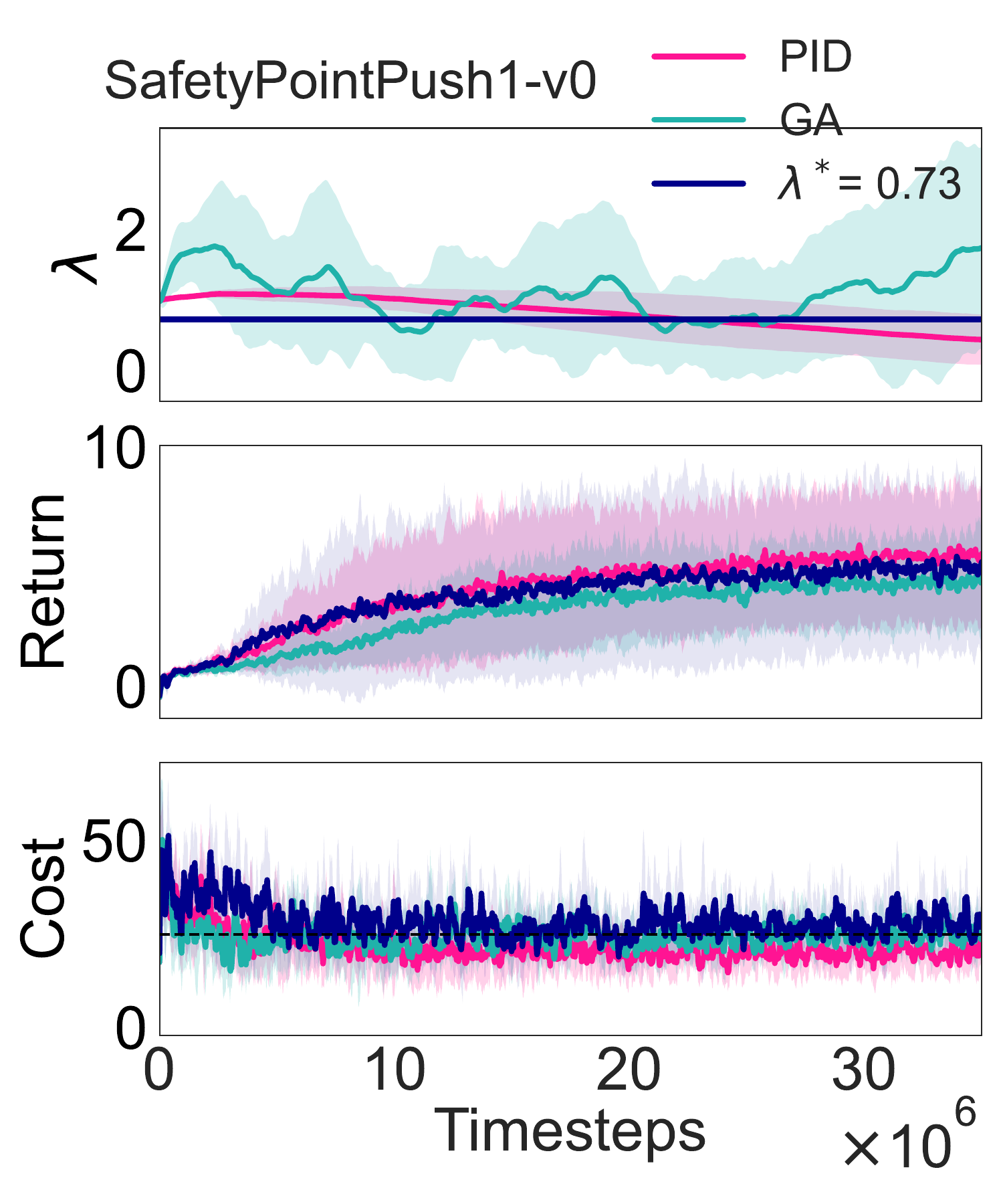}
    \caption{}
    \label{fig:comparison_push}
  \end{subfigure}\hfill
  \begin{subfigure}{0.33\textwidth}
    \centering
    \includegraphics[width=1\linewidth]{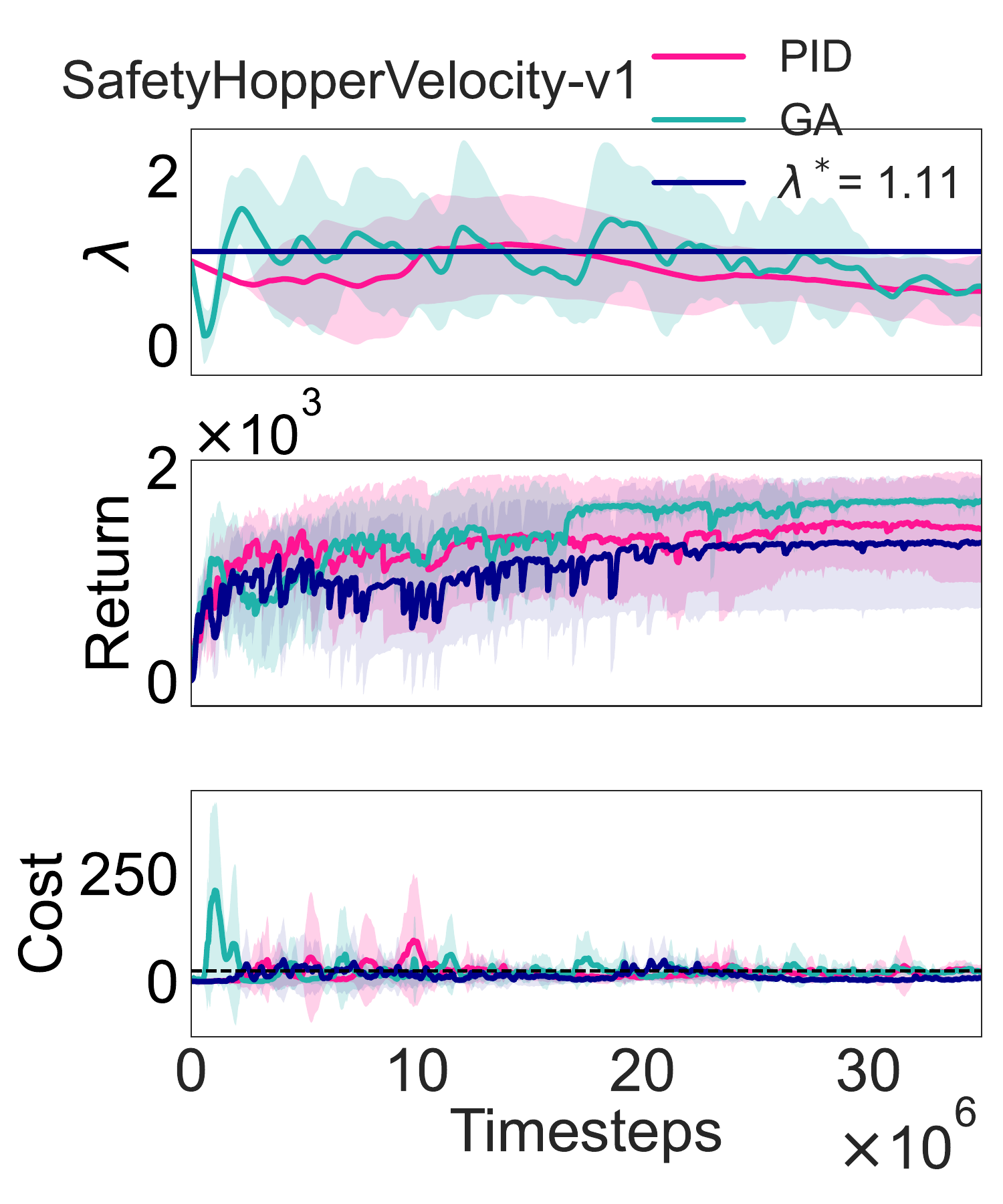}
    \caption{}
    \label{fig:comparison_hopper}
  \end{subfigure}\hfill
  \begin{subfigure}{0.33\textwidth}
    \centering
    \includegraphics[width=1\linewidth]{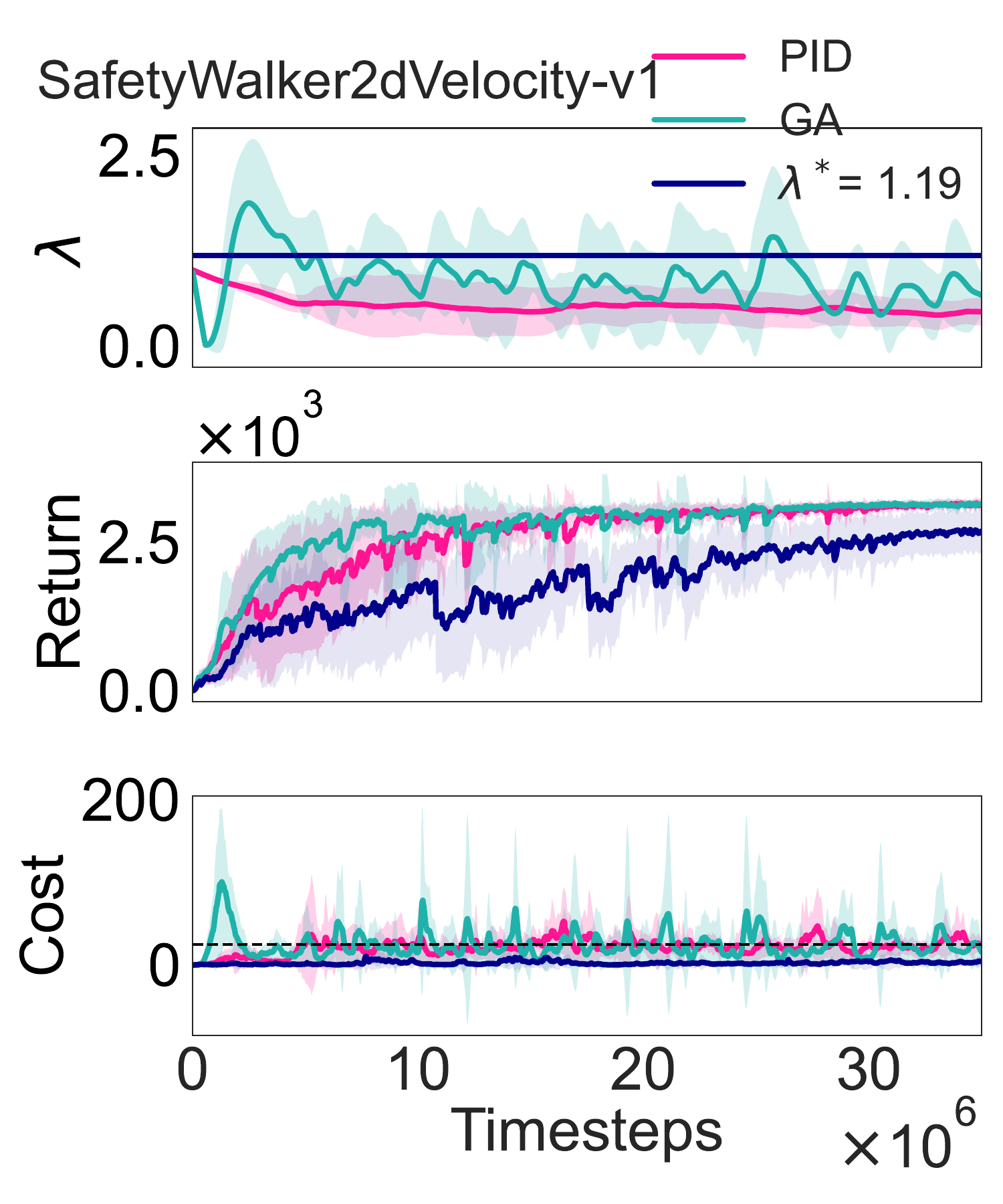}
    \caption{}
    \label{fig:comparison_walker2d}
  \end{subfigure}
  \vspace{0.5em}
  \begin{subfigure}{0.33\textwidth}
    \centering
    \includegraphics[width=1\linewidth]{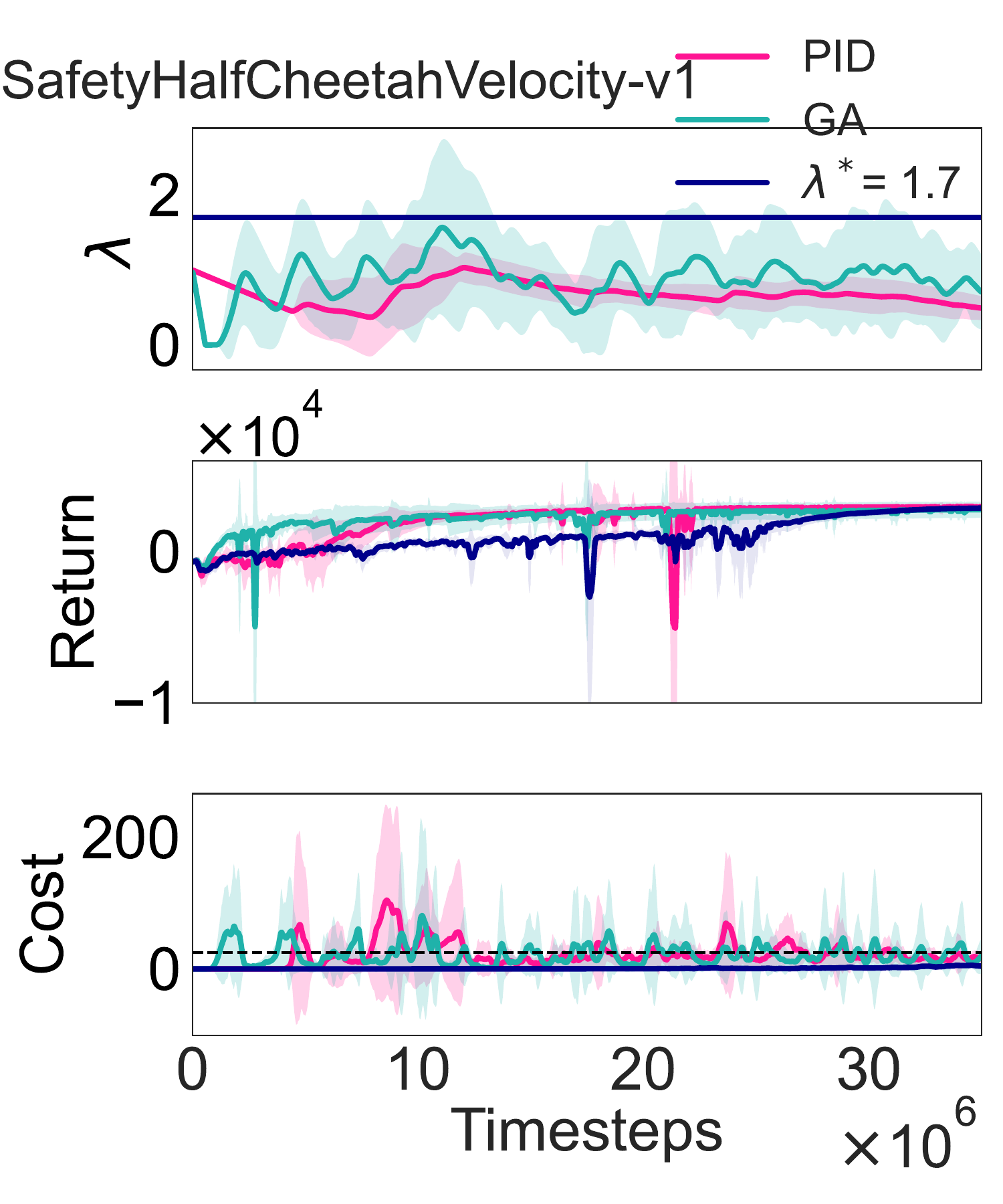}
    \caption{}
    \label{fig:comparison_halfcheetah}
  \end{subfigure}
  \hspace{2em}
  \begin{subfigure}{0.33\textwidth}
    \centering
    \includegraphics[width=1\linewidth]{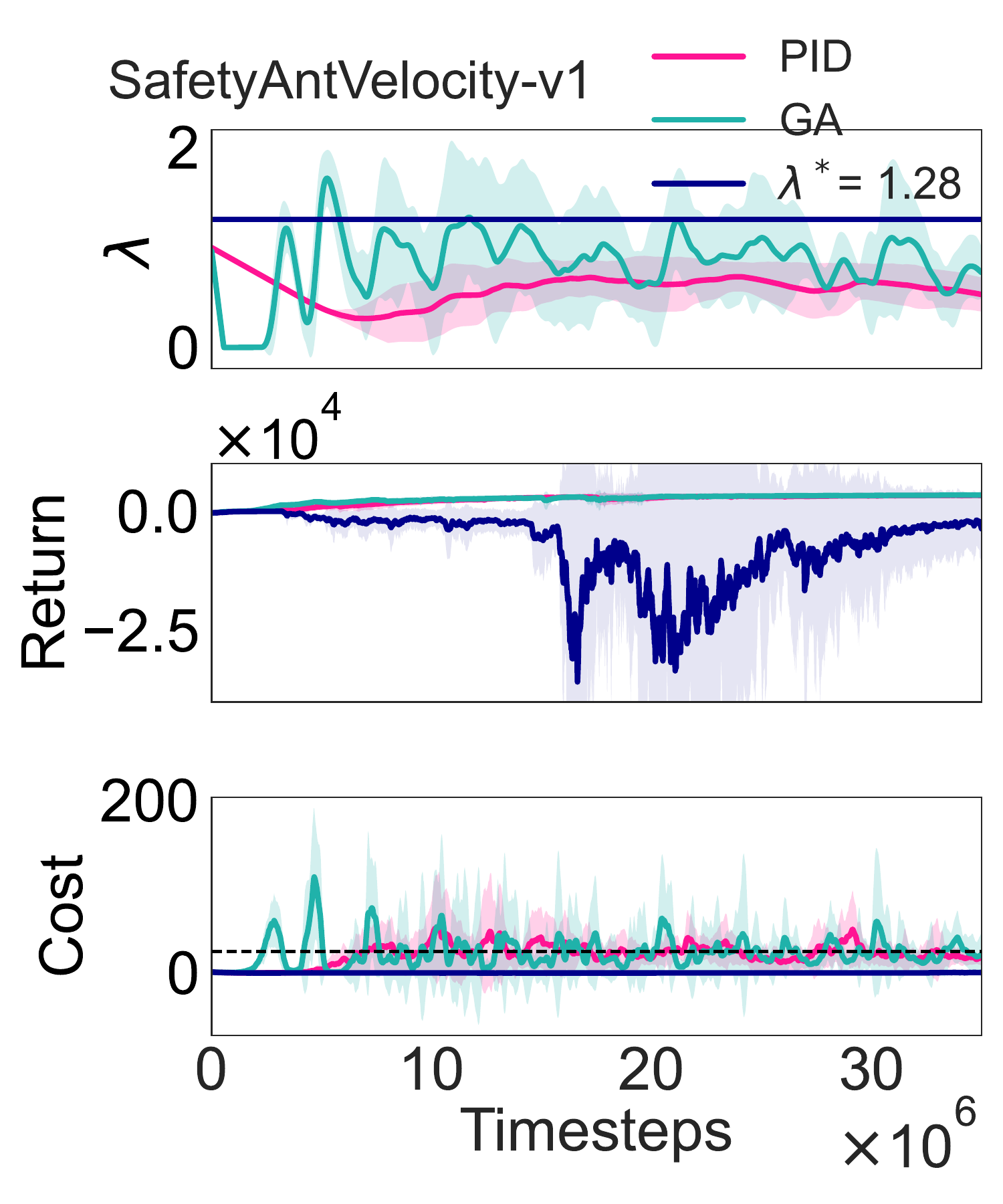}
    \caption{}
    \label{fig:comparison_ant}
  \end{subfigure}

  \caption{Training trajectories of $\lambda$ (top), return (middle), and cost (bottom) for GA- and PID-updated $\lambda$ and the fixed optimal multiplier $\lambda^*$ for a cost limit of 25.0. All results are averaged over 10 seeds. The shaded regions indicate the standard deviation over seeds. $K_P=10^{-4}, K_I=10^{-4}, K_D=0.0, \lambda_0=1.0$.}
  \label{fig:comparison_rsi}
\end{figure}

\begin{figure}[H]
  \centering

  \begin{subfigure}{0.33\textwidth}
    \centering
    \includegraphics[width=1\linewidth]{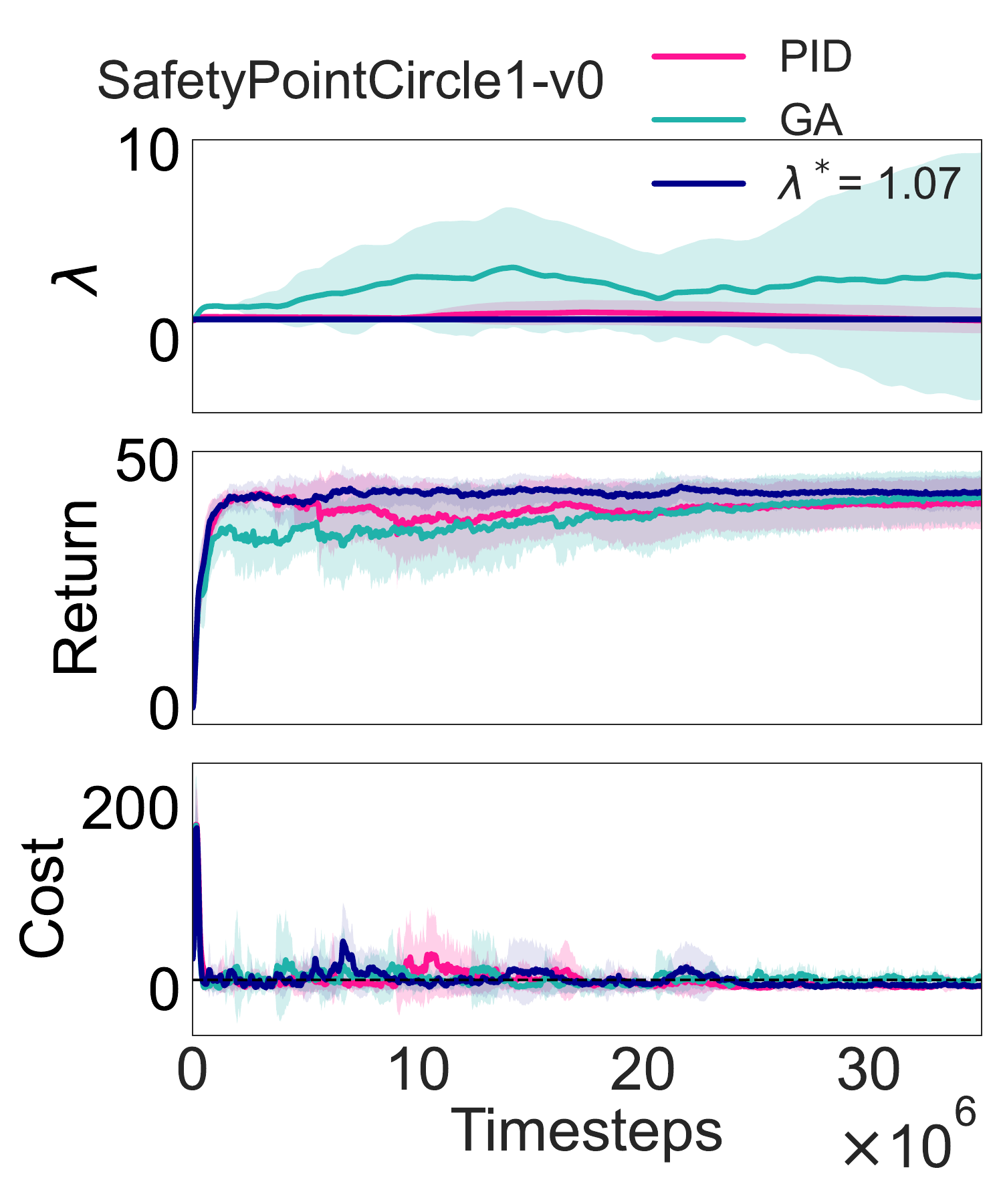}
    \caption{}
    \label{fig:comparison_circle10}
  \end{subfigure}\hfill
  \begin{subfigure}{0.33\textwidth}
    \centering
    \includegraphics[width=1\linewidth]{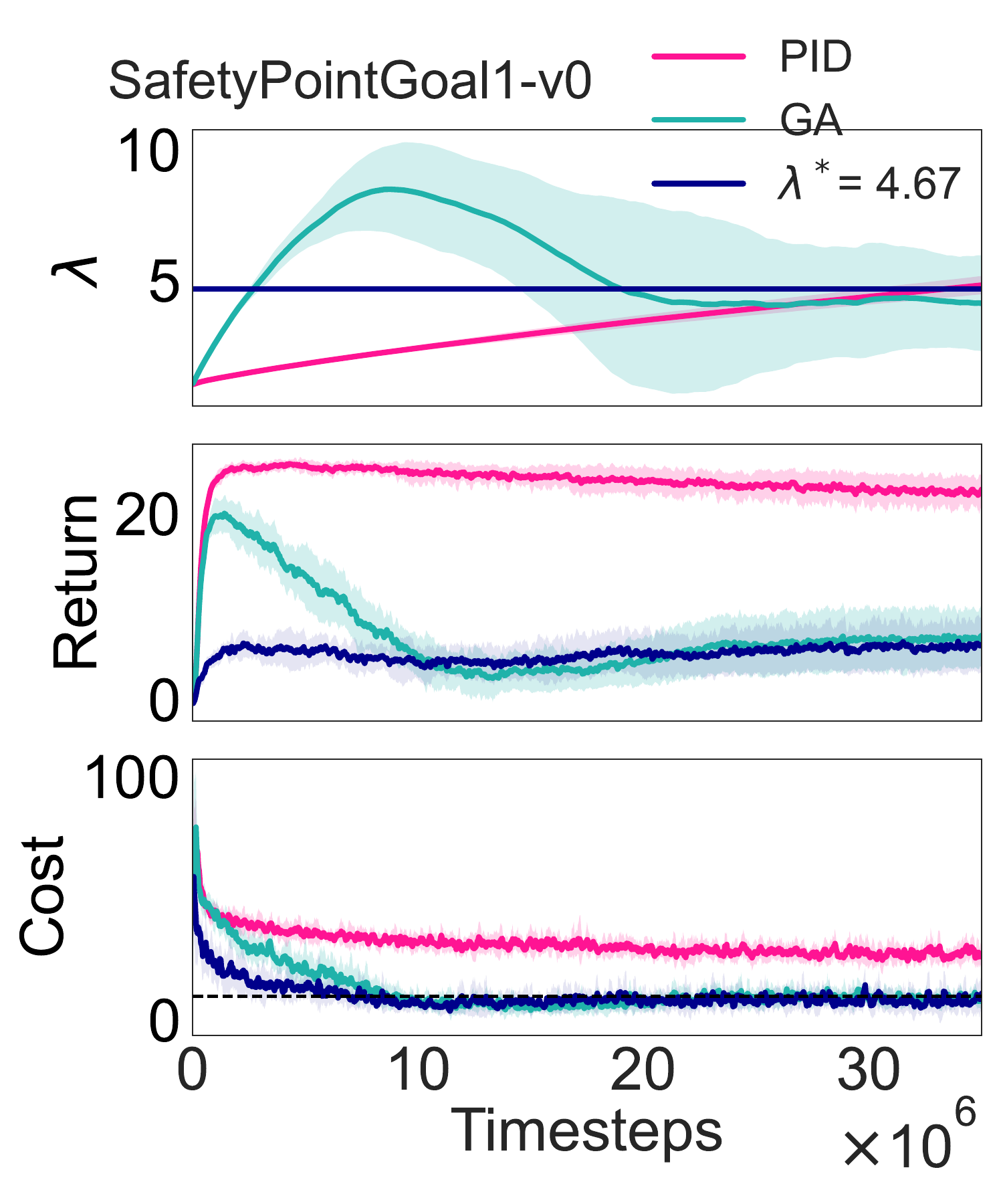}
    \caption{}
    \label{fig:comparison_goal10}
  \end{subfigure}\hfill
  \begin{subfigure}{0.33\textwidth}
    \centering
    \includegraphics[width=1\linewidth]{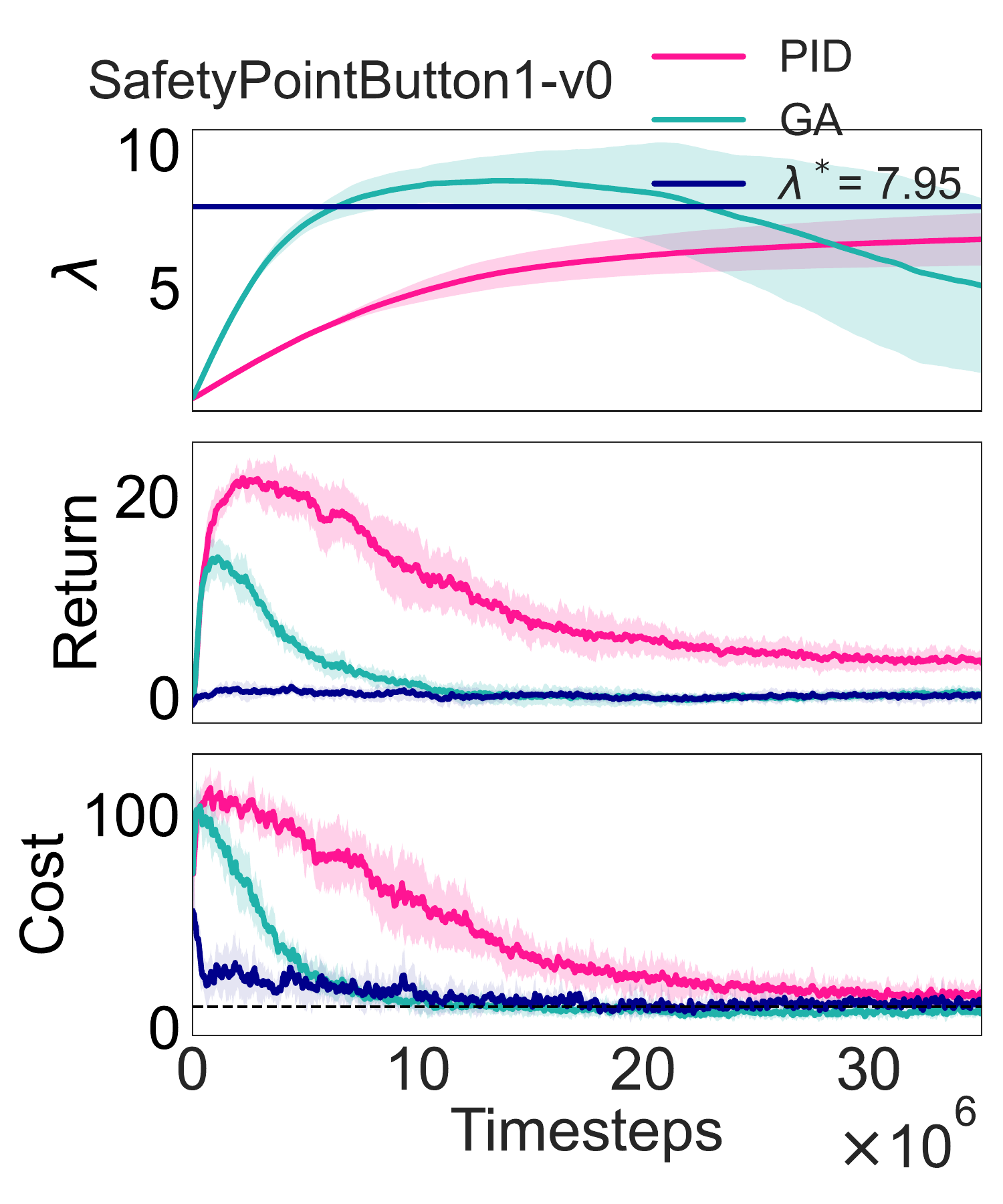}
    \caption{}
    \label{fig:comparison_button10}
  \end{subfigure}
  \vspace{0.5em}
  \begin{subfigure}{0.33\textwidth}
    \centering
    \includegraphics[width=1\linewidth]{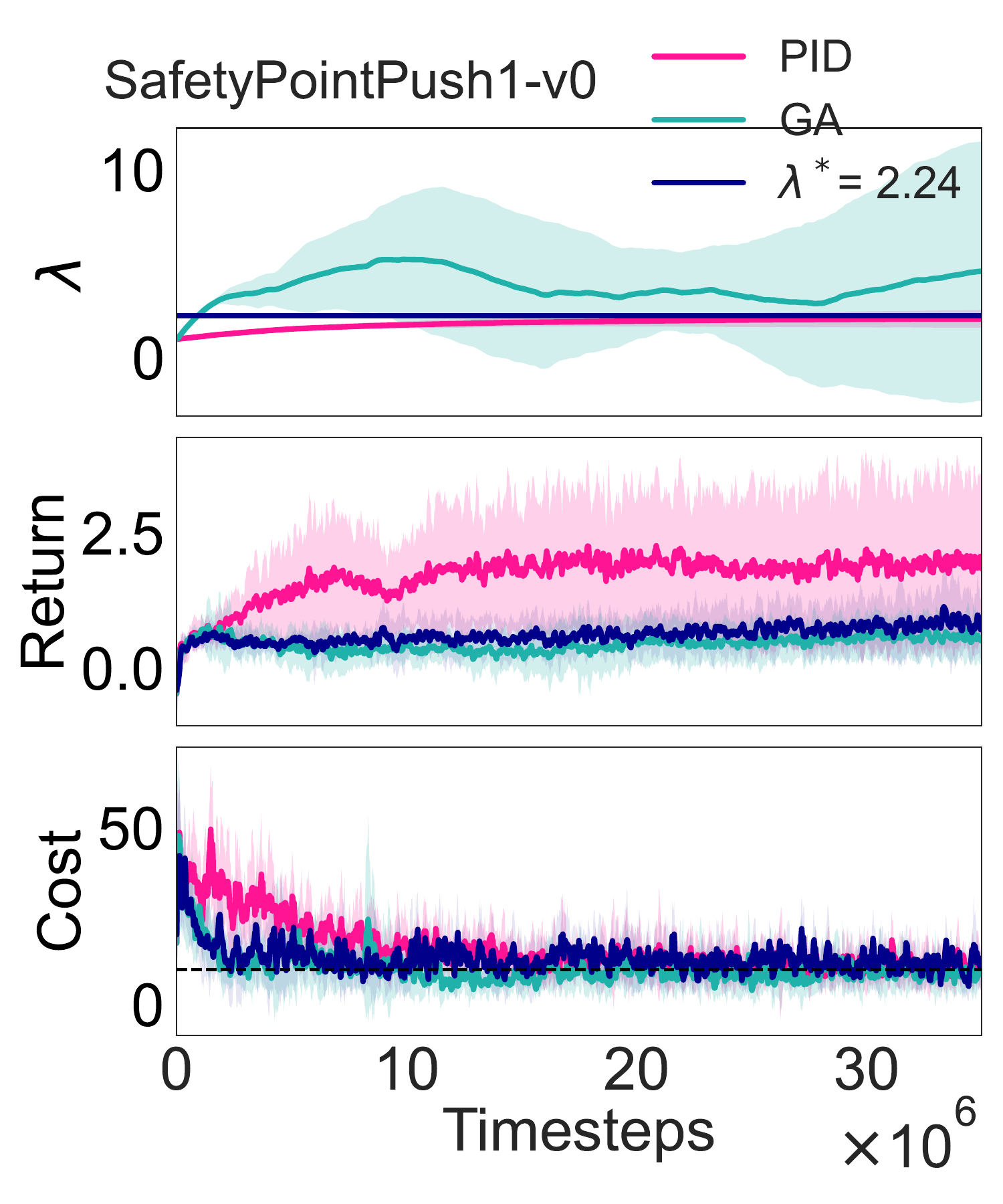}
    \caption{}
    \label{fig:comparison_push10}
  \end{subfigure}\hfill
  \begin{subfigure}{0.33\textwidth}
    \centering
    \includegraphics[width=1\linewidth]{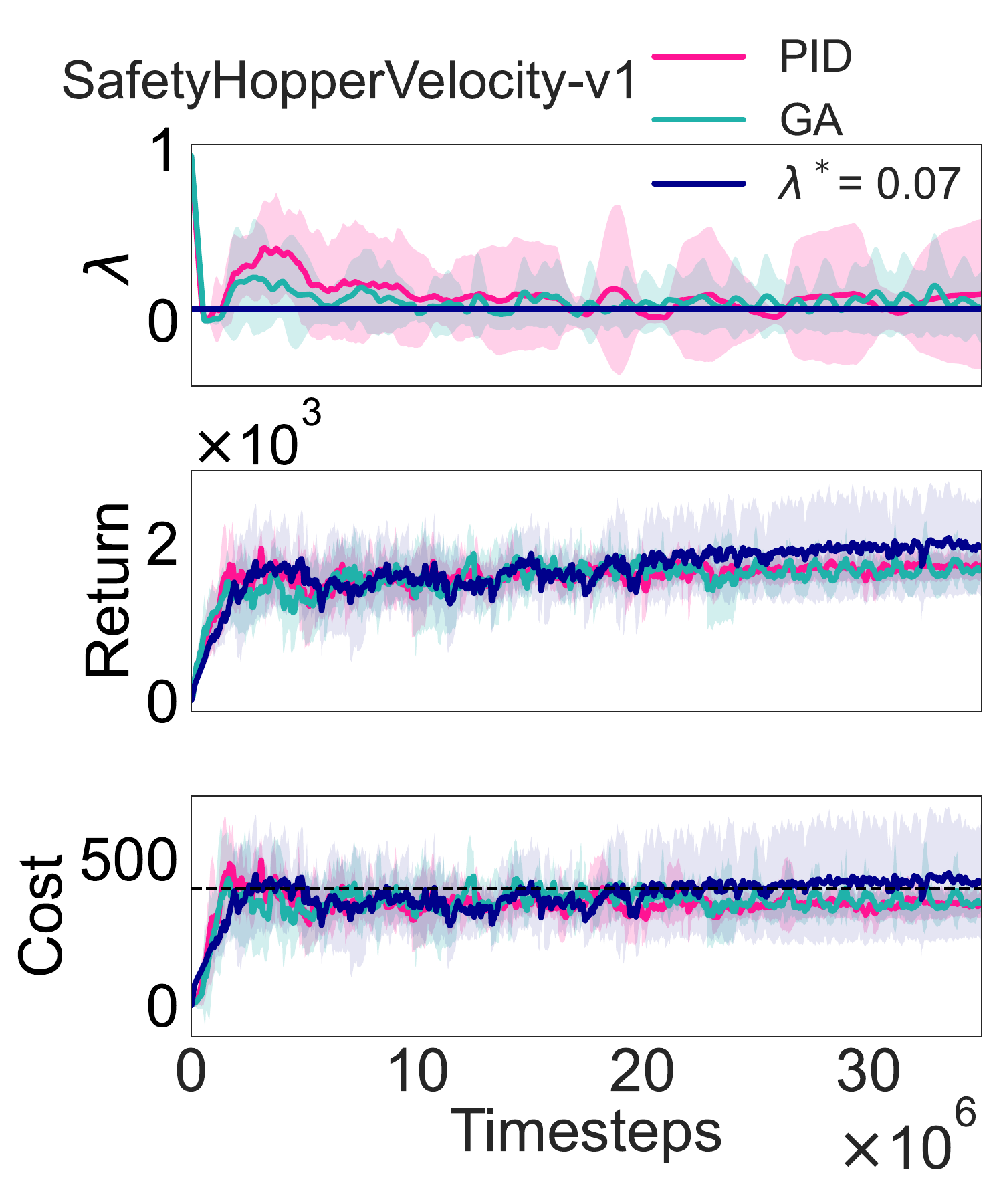}
    \caption{}
    \label{fig:comparison_hopper400}
  \end{subfigure}\hfill
  \begin{subfigure}{0.33\textwidth}
    \centering
    \includegraphics[width=1\linewidth]{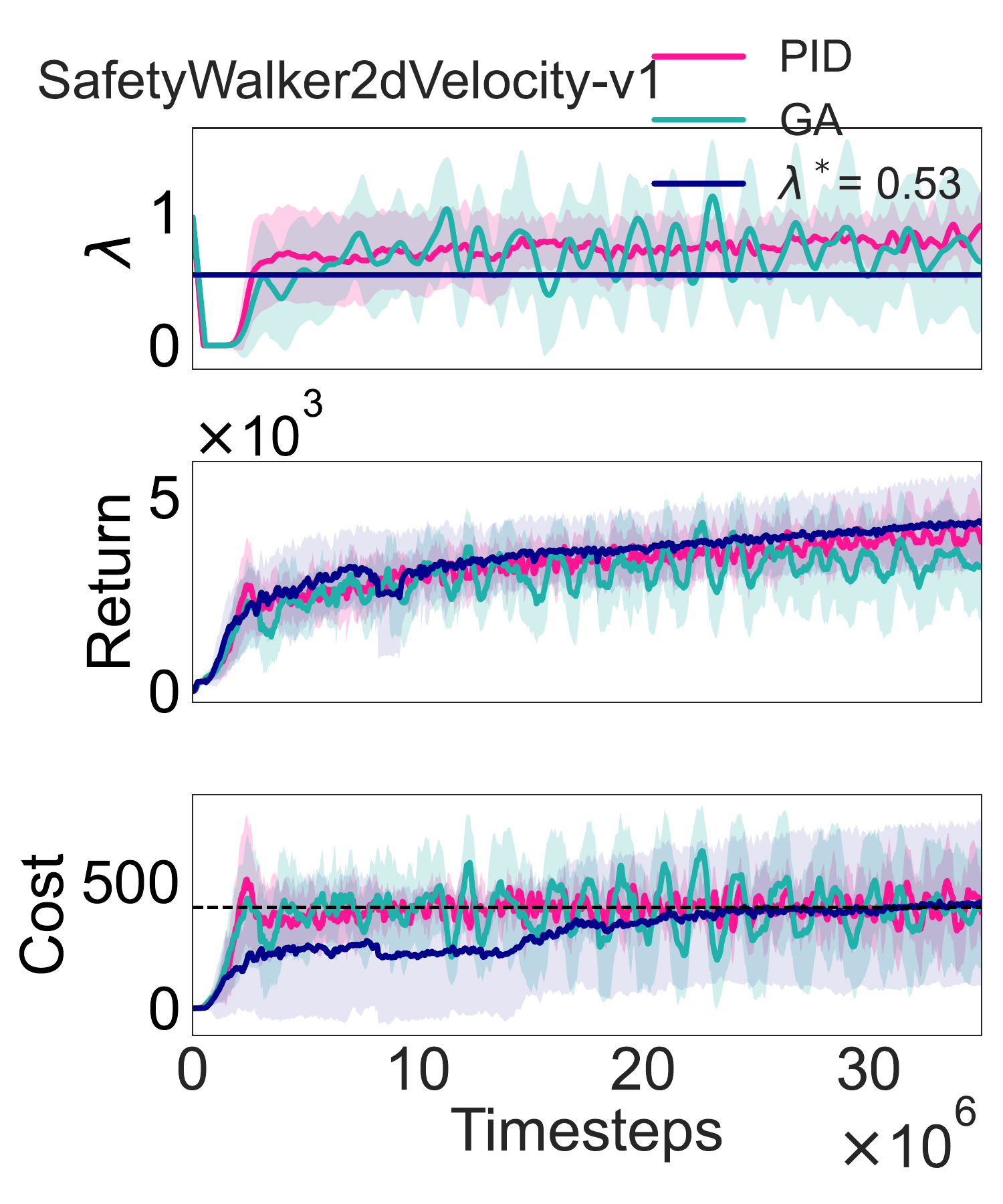}
    \caption{}
    \label{fig:comparison_walker2d400}
  \end{subfigure}
  \vspace{0.5em}
  \begin{subfigure}{0.33\textwidth}
    \centering
    \includegraphics[width=1\linewidth]{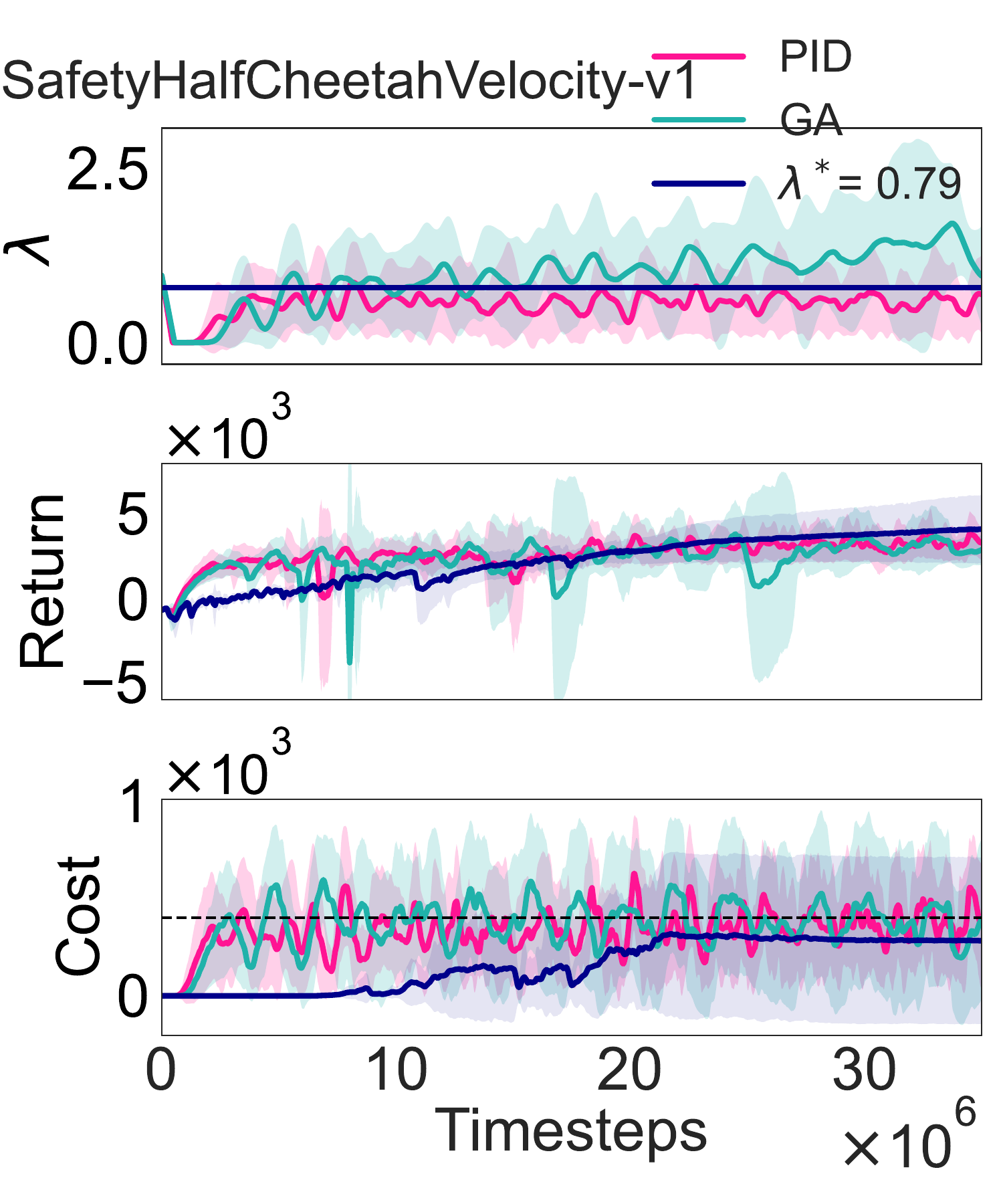}
    \caption{}
    \label{fig:comparison_halfcheetah400}
  \end{subfigure}
  \hspace{2em}
  \begin{subfigure}{0.33\textwidth}
    \centering
    \includegraphics[width=1\linewidth]{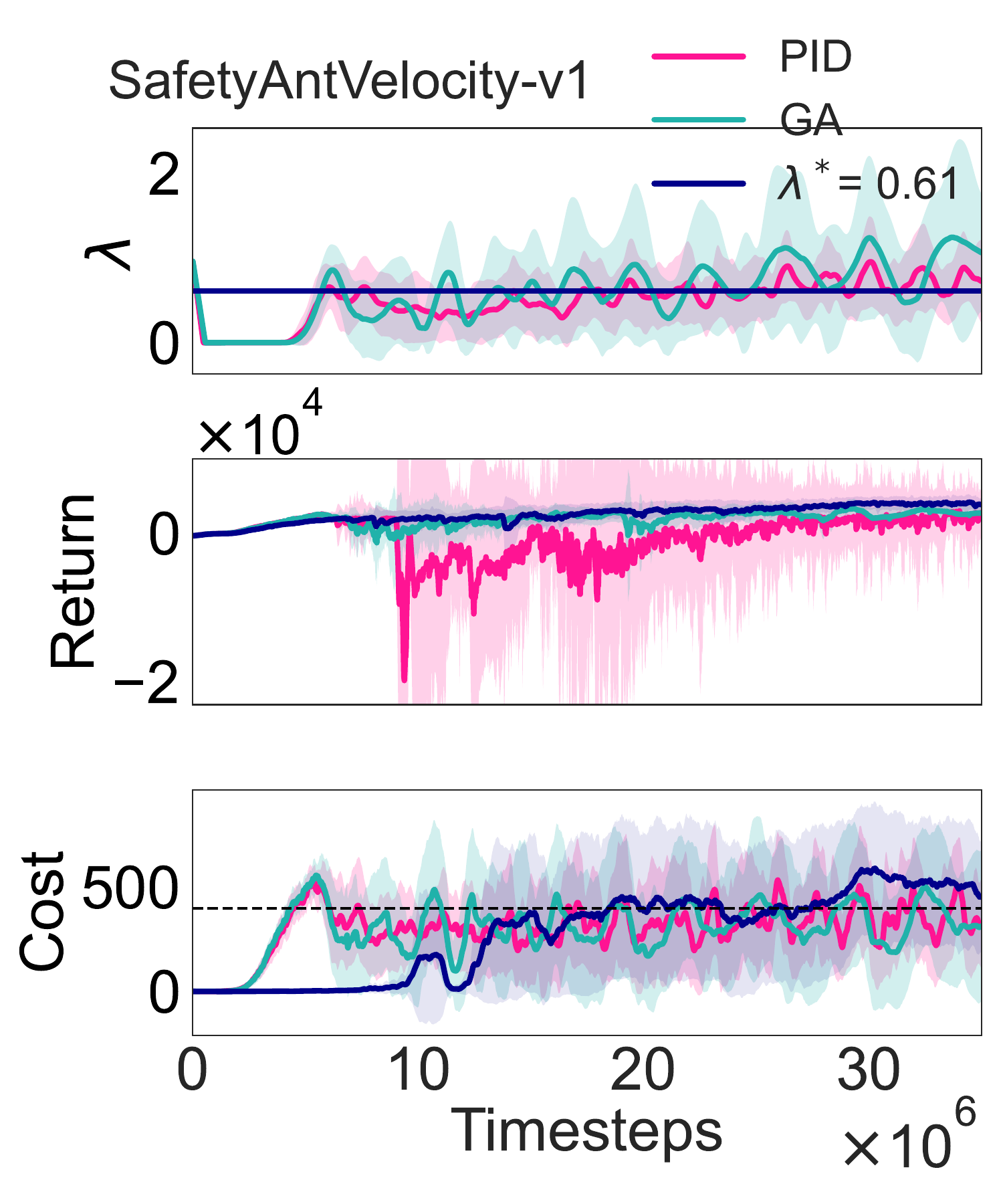}
    \caption{}
    \label{fig:comparison_ant400}
  \end{subfigure}

  \caption{Training trajectories of $\lambda$ (top), return (middle), and cost (bottom) for GA- and PID-updated $\lambda$ and the fixed optimal multiplier $\lambda^*$ for a cost limit of 10.0 for the navigation tasks, and a cost limit of 400.0 for the velocity tasks. All results are averaged over 10 seeds. The shaded regions indicate the standard deviation over seeds.  $K_P=10^{-4}, K_I=10^{-4}, K_D=0.0, \lambda_0=1.0$.}
  \label{fig:comparison_rsi_10_400}
\end{figure}

Figures \ref{fig:comparison_rsi} and \ref{fig:comparison_rsi_10_400} show the training curves of the results presented in Table \ref{tab:comparison_table} in the main paper. Training curves of individual seeds can be provided upon reasonable request.

\subsection{Preliminary results on off-policy methods} \label{sec: supp: saclagrsi}

Empirical results may vary under alternative algorithms. We therefore show preliminary results on off-policy methods, for which we evaluated the Lagrangian version of SAC on SafetyPointCircle1-v0 \citep{haarnoja2018soft}. 

\begin{figure}[H]
  \centering

  \begin{subfigure}{0.4\textwidth}
    \centering
    \includegraphics[width=1\linewidth]{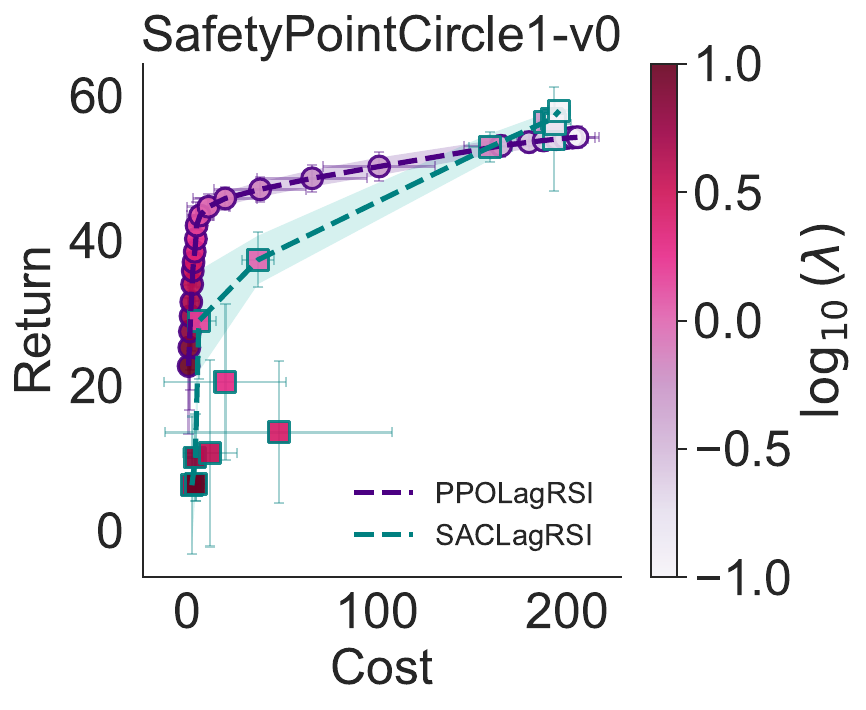}
    \caption{}
    \label{fig:saclagrsi_circle1_pareto_frontier}
  \end{subfigure}\hspace{4em}
  \begin{subfigure}{0.4\textwidth}
    \centering
    \includegraphics[width=1\linewidth]{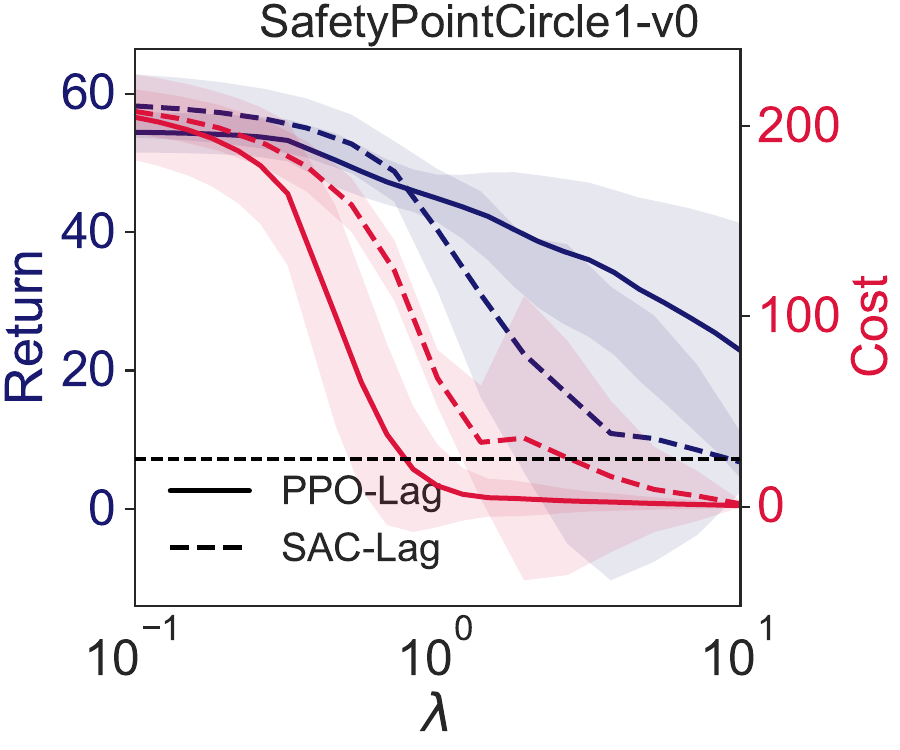}
    \caption{}
    \label{fig:saclagrsi_circle1_lambda_profile}
  \end{subfigure}

  \caption{(\ref{fig:saclagrsi_circle1_pareto_frontier}) Smoothed empirical PF of return versus cost as a function of $\lambda$, averaged over 5 seeds for SACLag and averaged over 10 seeds for PPOLag. Error bars denote the $1\sigma$ across seeds, and the shaded region shows the 95\% confidence interval of the mean empirical PF. The local slope of each curve captures the sensitivity of the return with respect to changes in the allowed cost. (\ref{fig:saclagrsi_circle1_lambda_profile}) Smoothened $\lambda$-profiles averaged over 5 seeds for SACLag and over 10 seeds for PPOLag, with the shaded regions indicating the 95\% confidence intervals of the return and cost, for all evaluated tasks. A cost limit of 25.0 is indicated by the dashed lines.}
  \label{fig:saclagrsi_lambda_profile_pareto_curve_circle1}
\end{figure}

We train 15 policies with fixed Lagrange multipliers by sweeping $\lambda \in \{10^{\ell_i}\}_{i=1}^{15}$, where $\ell_i$ are evenly spaced in $(-1,1)$. For each $\lambda$, we compute the average return and cost over the final 5\% of training. As in the PPO experiments, we apply the policy gradient scaling factor from Eq. \ref{eq:rsi_policy_gradient} (Section \ref{sec: supp: details on experimental setup}). For SAC, $J^R(\pi_{\theta_k})$ corresponds to the standard maximum-entropy actor objective \citep{haarnoja2018soft}. Hyperparameters for SAC-Lag are listed in Table \ref{tab:hparams_saclagrsi}. Results are shown as an empirical PF and $\lambda$-profile and compared with PPO-Lag in Figure \ref{fig:saclagrsi_lambda_profile_pareto_curve_circle1}.

It is important to note that the preliminary SAC-Lag results in Figure \ref{fig:saclagrsi_lambda_profile_pareto_curve_circle1}, based on 5 seeds, are less statistically significant than the PPO-Lag results averaged over 10 seeds. Nevertheless, the empirical PF indicates that the constraint geometry of SAC-Lag policies differs from PPO-Lag. This suggests the need for further evaluation of Lagrangian-based off-policy algorithms. Figure \ref{fig:saclagrsi_circle1_pareto_frontier} shows that PPO-Lag achieves a more favorable return–cost trade-off for this task.

\begin{table}[H]
\centering
\footnotesize
\caption{Hyperparameter settings for SAC-Lag.}
\label{tab:hparams_saclagrsi}

\begin{tabular}{@{}ll @{\hspace{1.2cm}} ll@{}}
\toprule
\multicolumn{2}{c}{\textbf{General training settings}} 
& \multicolumn{2}{c}{\textbf{Other settings}} \\
\cmidrule(r){1-2} \cmidrule(l){3-4}

Steps / epoch & 2{,}000 
& Replay buffer size & $10^6$ \\

Update cycle & 20 
& Discount $\gamma$ & 0.99 \\

Update iterations / cycle & 1 
& Polyak averaging & 0.005 \\

Batch size & 256 
& Start learning steps & 10{,}000 \\

Policy delay & 2 
& Entropy temperature $\alpha$ & $10^{-5}$ \\

Actor network & [512, 512], ELU 
& Critic network & [512, 512], ELU \\

Actor LR & $5\times10^{-6}$ 
& Critic LR & $10^{-3}$ \\

\midrule
\multicolumn{4}{c}{\textbf{Reward-scale invariance}} \\
\midrule

$\Delta\beta$ EMA $\alpha$ & 0.9 
& $\beta$ grad batch & 128 \\

$\beta$ grad frequency & 50 
& $\beta$ grad $\epsilon$ & $10^{-8}$ \\

\bottomrule
\end{tabular}
\end{table}

\end{document}